\documentclass{article}

    \usepackage[final]{icbinb2021}

\usepackage[utf8]{inputenc} 
\usepackage[T1]{fontenc}    
\usepackage{hyperref}       
\usepackage{url}            
\usepackage{booktabs}       
\usepackage{amsfonts}       
\usepackage{nicefrac}       
\usepackage{microtype}      
\usepackage[dvipsnames]{xcolor}         
\usepackage{amsmath}
\usepackage[noabbrev]{cleveref}
\usepackage{graphicx}
\usepackage{subfigure}
\usepackage{amssymb}

\usepackage{wrapfig}
\usepackage{lipsum}
\usepackage{placeins}

\usepackage{tikz}
\usetikzlibrary{arrows, arrows.meta, bayesnet, chains, positioning, decorations.pathreplacing, fit, calc, decorations.pathmorphing}

\usepackage{layouts}  
\tikzset{scaled/.style={scale=#1}}
\tikzset{scaled/.default=1.}

\definecolor{depth1}{HTML}{ff7f0e}
\definecolor{depth2}{HTML}{2ca02c}
\definecolor{depth3}{HTML}{d62728}
\definecolor{depth4}{HTML}{9467bd}
\definecolor{depth5}{HTML}{17becf}
\tikzset{nicecolor/.style={draw=#1!50!black, fill=#1!40!white}}
    
\tikzset{block/.style={rectangle, text width=2ex, align=center, minimum height=2ex, nicecolor=#1, font=\scshape, rounded corners}}
\tikzset{block/.default=red}

\tikzset{opnode/.style={circle, draw, minimum size=1ex, text width=1ex}}
\tikzset{plus/.style ={opnode, font={$+$}}}

\tikzset{line/.style={line width=.15ex, draw=black}}
\tikzset{arrow/.style={-{Latex[length=1.3ex,width=0.8ex]},line}}
\tikzset{arrowsm/.style={-{Latex[length=1.ex,width=0.66ex]},line}}
\tikzset{snake/.style={decoration={snake, pre length=0.01mm, segment length=2mm, amplitude=0.2mm, post length=1mm}, decorate}}



\title{Addressing Bias in Active Learning  with Depth Uncertainty Networks... or Not}

\author{%
  Chelsea Murray \hspace{2.5mm} 
  James U. Allingham \hspace{2.5mm}
  Javier Antor\'an \hspace{2.5mm}
  Jos\'e Miguel Hern\'andez-Lobato \\
  Department of Engineering \\
  University of Cambridge \\
  \texttt{\{clm88, jua23, ja666, jmh233\}@cam.ac.uk}
}

\begin{document}

\maketitle

\begin{abstract}

\citet{farquhar_statistical_2020} show that correcting for active learning bias with underparameterised models leads to improved downstream performance. For overparameterised models such as NNs, however, correction leads either to decreased or unchanged performance.
They suggest that this is due to an ``overfitting bias'' which offsets the active learning bias.  
We show that depth uncertainty networks operate in a low overfitting regime, much like underparameterised models. They should therefore see an increase in performance with bias correction. Surprisingly, they do not.
We propose that this negative result, as well as the results \citet{farquhar_statistical_2020}, can be explained via the lens of the bias-variance decomposition of generalisation error.
\end{abstract}

\section{Introduction}\label{sec:intro}


Active learning improves data efficiency by identifying which examples maximise the expected gain in model performance \citep{cohn1995active, settles2009active}. An acknowledged but relatively poorly understood problem with this approach, however, is that actively selecting informative training points introduces bias in inference, as the training data no longer follow the population distribution \citep{mackay1992information}. Given that many statistical results rely on data points being identically and independently distributed (i.i.d.) samples from the population distribution, applying standard estimators to actively sampled datasets results in optimising for the wrong objective \citep{farquhar_statistical_2020}. 

Recently, \citet{farquhar_statistical_2020} introduced an estimator, $\tilde{R}_{\text{LURE}}$, that corrects for the bias induced by active sampling by re-weighting the contribution of individual data points to the estimated population risk. \citet{farquhar_statistical_2020} find that, while underparameterised models such as linear regression benefit from the elimination of active learning bias, the performance of overparameterised models such as neural networks is either unaffected or negatively impacted by the unbiased risk estimator. This result is explained as active learning bias acting as a regulariser and helping to prevent flexible models from overfitting on small datasets.

Depth uncertainty networks (DUNs) are a class of Bayesian neural network (BNN) in which a prior is placed over the depth of the network, allowing us to favour simpler (i.e., shallower) functions. This enables DUNs to adapt their complexity to the varying size of the training dataset as additional labels are acquired during active learning. Because of this, we posit that DUNs will suffer less from overfitting in the early stages of training than other BNNs. This would imply that using the unbiased risk estimator of \citet{farquhar_statistical_2020} to eliminate active sampling bias should improve the performance of DUNs. Our main contributions are:



\begin{itemize}
    \item Empirically validating the hypothesis that DUNs present less overfitting bias than other BNNs, specifically Monte Carlo dropout (MCDO), during active learning;
    \item Investigating the application of the $\tilde{R}_{\text{LURE}}$ estimator to eliminate active learning bias from DUNs. Unexpectedly, despite DUNs having larger active learning bias than overfitting bias, we find that $\tilde{R}_{\text{LURE}}$ does not result in improved downstream performance;
    \item Providing hypotheses for why we do not observe an impact on DUNs' downstream performance when applying the unbiased risk estimator.
\end{itemize}

\section{Background}\label{sec:background}

\subsection{Bias in active learning}\label{sec:bias}

We use notation similar to \citet{farquhar_statistical_2020}. Given a loss function $\mathcal{L}( \mathbf{y}, f_{\theta}(\mathbf{x}))$, we aim to find $\theta$ that minimises the population risk over $p_{\text{data}}(\mathbf{y}, \mathbf{x})$: ${\mathbb{E}_{\mathbf{x},  \mathbf{y} \sim p_{\text {data }}}\left[\mathcal{L}\left(\mathbf{y}, f_{\theta}(\mathbf{x})\right)\right]}$. In practice, we only have access to $N$ samples from $p_{\text{data}}$. These yield the empirical risk, ${r=\frac{1}{N} \sum_{n=1}^{N} \mathcal{L}\left(\mathbf{y}_{n}, f_{\theta}\left(\mathbf{x}_{n}\right)\right)}$. $r$ is an unbiased estimator of the population risk when the data are drawn i.i.d. from $p_{\text{data}}$. If these samples have been used to train $\theta$, our estimators become biased. We refer to estimators biased by computing with the training data with capital letters (i.e., $R$). In the active learning setting, our model is optimised using a subset of $M$ actively sampled data points:
\begin{equation}
    \tilde{R}=\frac{1}{M} \sum_{m=1}^{M} \mathcal{L}\left(\mathbf{y}_{i_m}, f_{\theta}\left(\mathbf{x}_{i_m}\right)\right), \quad i_m \sim \alpha\left(i_{1: m-1}, \mathcal{D}_{\text {pool }}\right) . \label{eq:r_tilde}
\end{equation} 
The proposal distribution $\alpha\left(i_{m} ; i_{1: m-1}, \mathcal{D}_{\text {pool }}\right)$ over the pool of unlabelled data $\mathcal{D}_{\text{pool}}$ represents the probability of each index being sampled next, given that we have already acquired $\mathcal{D}_{\text{train}} = \{\mathbf{x}_i\}_{1:m-1}$. As a result of active sampling, $\tilde{R}$ is not an unbiased estimator of $R$. \citet{farquhar_statistical_2020} propose $\tilde{R}_{\text{LURE}}$ (``levelled unbiased risk estimator''), which removes active learning bias:
\begin{equation}
    \begin{aligned}
    \tilde{R}_{\text{LURE}} \!\equiv\! \frac{1}{M} \sum_{m=1}^{M} v_{m} \mathcal{L}_{i_{m}} ;
    \quad v_{m} \!\equiv 1+\frac{N-M}{N-m}\left(\frac{1}{(N-m+1)\ \alpha\left(i_{m} ; i_{1: m-1}, \mathcal{D}_{\text{pool}}\right)}-1\right), \label{eq:r_lure} 
    \end{aligned}
\end{equation}
where $\mathcal{L}_{i_{m}} \equiv \mathcal{L}\left(\mathbf{y}_{i_{m}}, f_{\theta}\left(\mathbf{x}_{i_{m}}\right)\right)$.  Intuitively, the estimator works by re-weighting each example's contribution to the total loss by its inverse acquisition probability, such that if unusual examples have particularly high acquisition probabilities, they contribute proportionally less to the total risk. A detailed derivation of this estimator is given by \citet{farquhar_statistical_2020}.

Surprisingly, \citet{farquhar_statistical_2020} find that using $\tilde{R}_{\text{LURE}}$ during training can negatively impact the performance of flexible models. They explain this by noting that the overfitting bias ($r{-}R$) typical in such models has the opposite sign to the bias induced by active learning. That is, active learning encourages the selection of ``difficult'' to predict data points, increasing the overall risk, whereas overfitting results in the total risk being underestimated. Where overfitting is substantial, it is (partially) offset by active learning bias, such that correcting for active learning bias does not improve model performance.

\subsection{Depth uncertainty networks for Active Learning}\label{sec:duns}

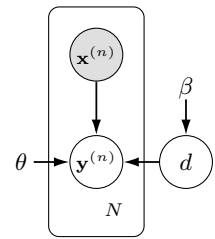
\begin{wrapfigure}{r}{0.2\textwidth}
\vspace{-0.1in}
    \centering
    \begin{subfigure}
    \centering
    \begin{tikzpicture}[every node/.style={scale=1}]
      \node[obs] (xn) {\scriptsize $\mathbf{x}^{(n)}$};
      \node[latent, below=4.5ex of xn] (yn) {\scriptsize $\mathbf{y}^{(n)}$};
      \node[const, left=3ex of yn] (theta) {\footnotesize $\theta \ $};
      \node[latent, right=3ex of yn] (d) {\footnotesize $d$};
      \node[const, above=3ex of d] (beta) {\footnotesize $\beta$};
      
      \edge[arrow] {xn} {yn} ; %
      \edge[arrow] {theta} {yn} ;
      \edge[arrow] {d} {yn} ;
      \edge[arrow] {beta} {d} ;
      
      \plate[inner sep=1.7ex] {} {(yn)(xn)} {\scriptsize $N$} ;
    \end{tikzpicture}
\end{subfigure}
\caption{DUN graphical model.}
\label{fig:dun_only_graph_model}
\vspace{-0.1in}
\end{wrapfigure}


DUNs, depicted in \cref{fig:dun_only_graph_model}, are a BNN variant in which the depth of the network $d \in [1,2,\!...,\!D]$ is treated as a random variable. Model weights $\theta$ are kept deterministic, simplifying optimisation. We place a categorical prior over depth $p(d)\,{=}\,\mathrm{Cat}(d| \{\beta_{i}\}_{i=0}^{D})$, resulting in a categorical posterior which can be marginalised in closed form. This procedure can be seen as Bayesian model averaging over an ensemble of subnetworks of increasing depth.
Thus, for DUNs, the predictive distribution and Bayesian active learning by disagreement (BALD) acquisition objective \citep{houlsby2011bayesian} are tractable and cheap to compute.
See \cref{app:duns} or \citep{antoran2020depth} for a detailed description.

In DUNs, predictions from different depths of the network correspond to different kinds of functions---shallow networks induce simple functions, while deeper networks induce more complex functions. 
We choose a prior that favours shallow models, limiting the flexibility of DUNs when few datapoints are observed. As the actively acquired set increases, the model likelihood will start to dominate the prior, leading the posterior over depth to favour deeper models and thus more flexible functions. We show this in \Cref{fig:res_reg_posts}. 
We hypothesize that the capability to automatically adapt model flexibility as more data are observed will reduce overfitting in the small data regimes typical of active learning. Therefore DUNs, unlike standard BNNs, may in fact benefit from the use active learning bias corrective weights. It is worth noting that automatically adapting model complexity to the observed data is also a promised feature of traditional BNNs. However, unlike in DUNs where inferences are exact, weight space models require crude approximations which fall short in this regard \citep{JMLR:v20:19-236}.




\section{Results}\label{sec:results}

First, using the analysis of \citet{farquhar_statistical_2020}, we show that the weighted estimator removes active learning bias for DUNs. We then show that DUNs overfit less than MCDO. Finally, we examine how training with $\tilde{R}_{\text{LURE}}$ impacts downstream performance for DUNs. Experiments are performed on nine UCI regression datasets \citep{hernandez2015probabilistic}; results for the Boston, Concrete and Energy datasets are presented in the following sections, with results for the rest provided in \cref{app:add_results}. The experimental setup is described in \cref{app:exp_setup}. All experiments are repeated 40 times, with the mean and standard deviations over the repetitions reported in our figures.

\subsection{Quantifying active learning bias}\label{sec:res_bias_alb}

\begin{figure}[ht!] 
\vspace{-1mm}
\centering
    \begin{subfigure}
        \centering
        \includegraphics[width=0.335\linewidth]{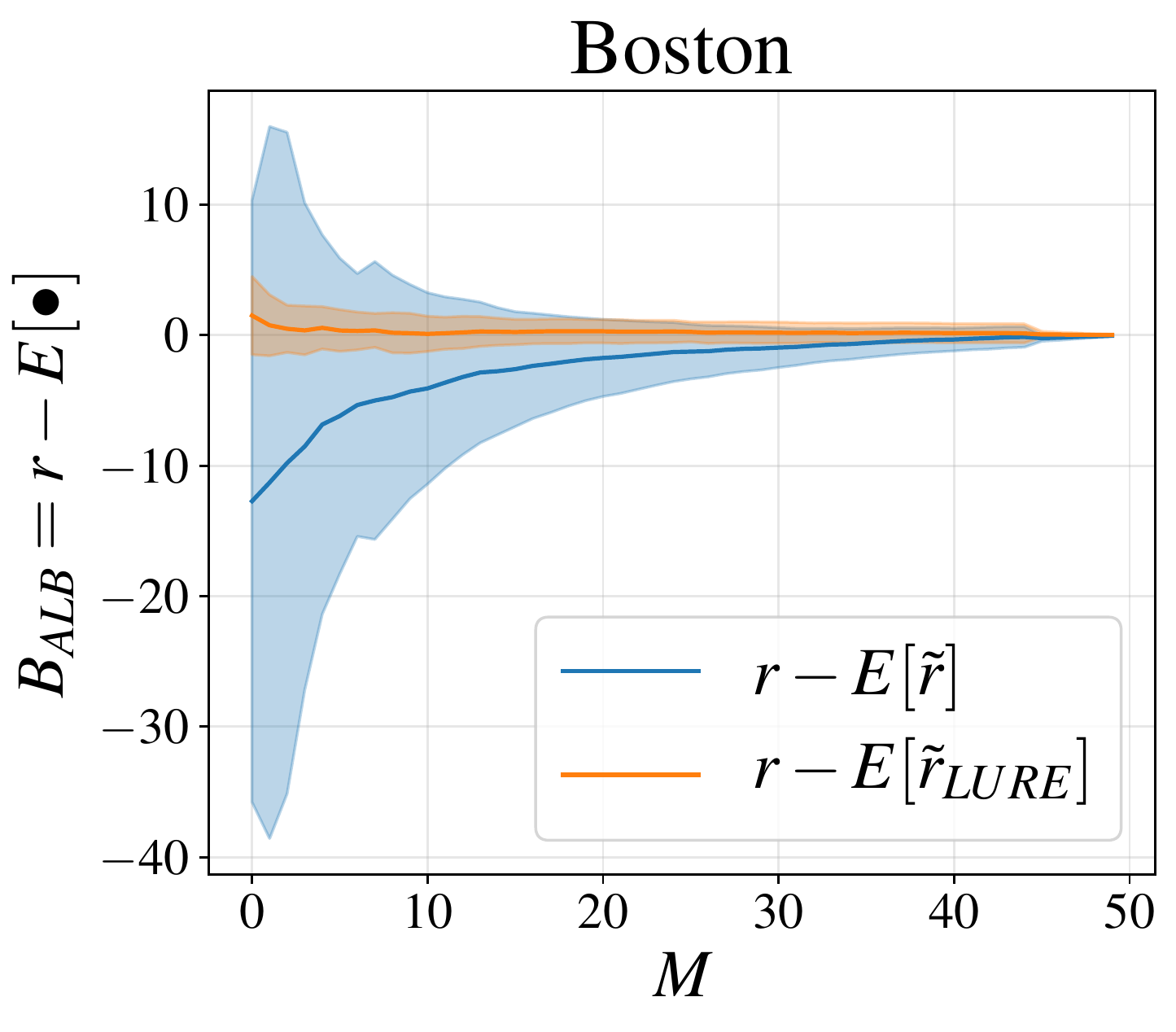}
    \end{subfigure}
    \begin{subfigure}
        \centering
        \includegraphics[width=0.31\linewidth]{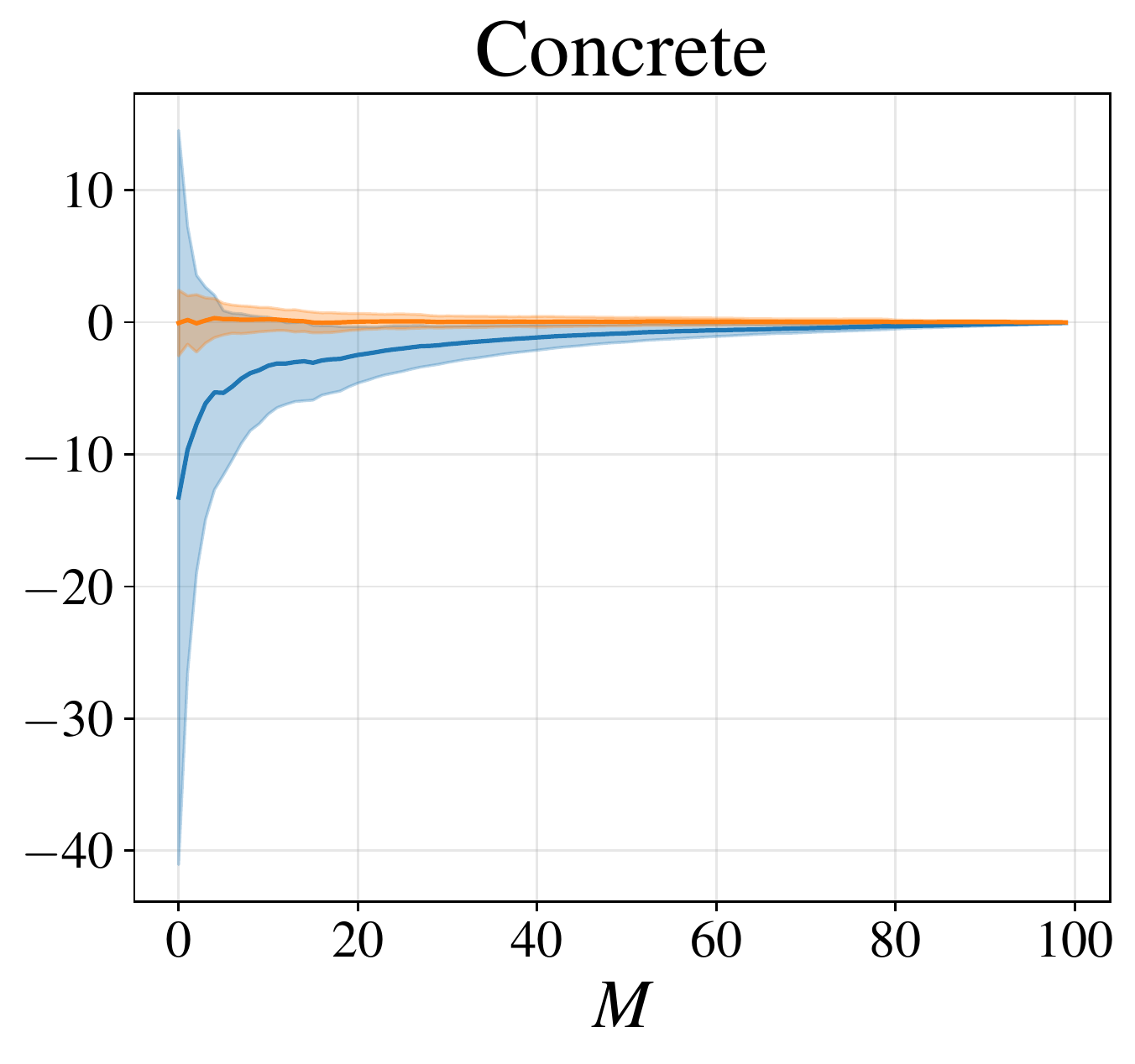}
    \end{subfigure} 
    \begin{subfigure}
        \centering
        \includegraphics[width=0.315\linewidth]{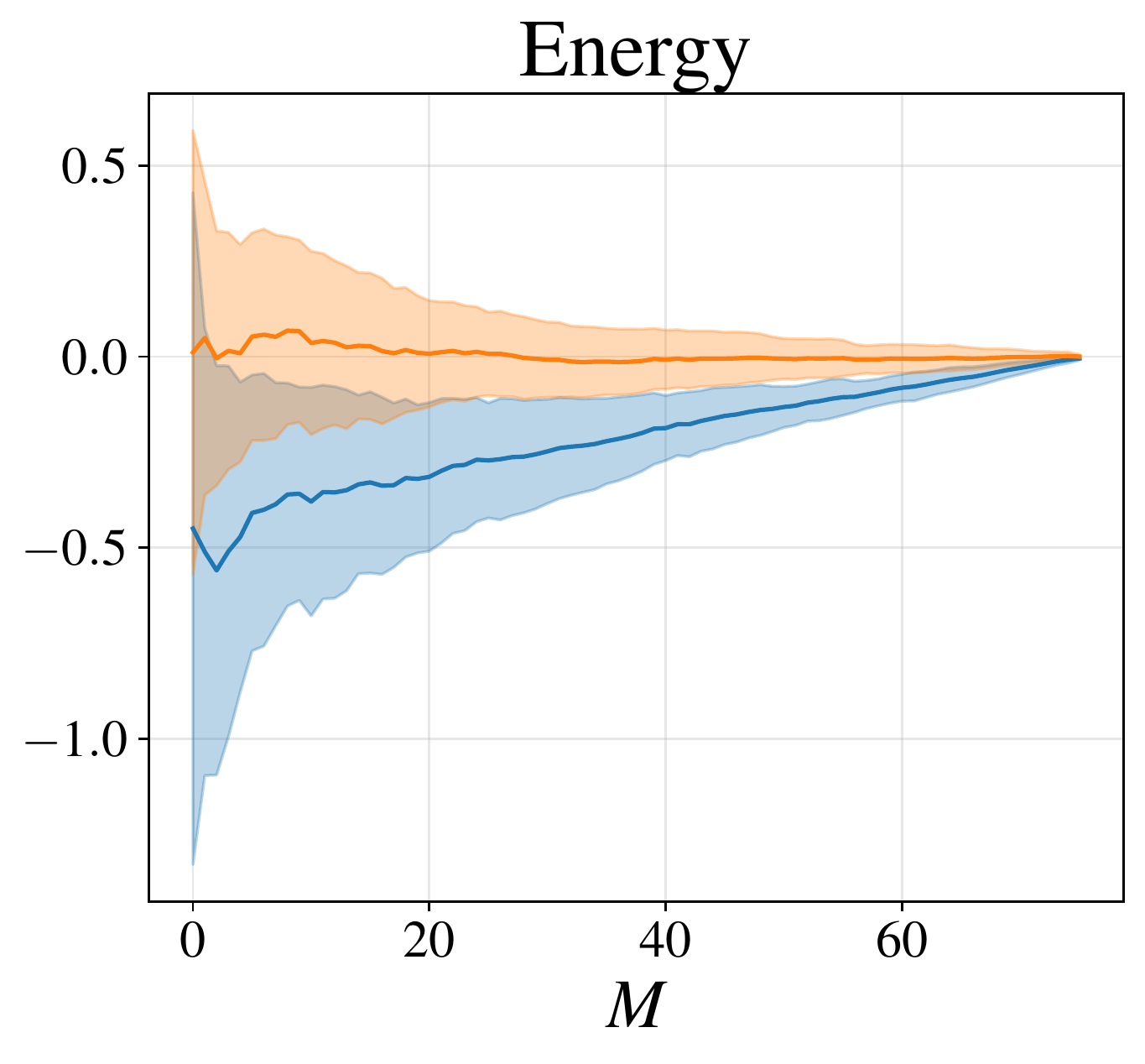} 
    \end{subfigure}
    \vspace{-5mm}
    \caption{Active learning bias in risk estimation with DUNs, evaluated using $\tilde{r}$ (blue) and $\tilde{r}_{\text{LURE}}$ (orange). The unbiased risk estimator $\tilde{r}_{\text{LURE}}$ eliminates active learning bias, while the standard estimator $\tilde{r}$ tends to overestimate risk due to $\alpha$ prioritising more difficult to fit points.}
    \label{fig:res_alb}
    \vspace{-1mm}
\end{figure}

We estimate the extent of active learning bias when using $\tilde{r}$ and verify the unbiasedness of $\tilde{r}_{\text{LURE}}$. A DUN is trained on $1,000$ randomly selected data points from $\mathcal{D}_{\text{pool}}$ using the standard negative log-likelihood (NLL) loss. We then estimate the model's risk using both $\tilde{r}$ and $\tilde{r}_{\text{LURE}}$, drawing $M$ samples from an unobserved test set $\mathcal{D}_{\text{test}}$. The active learning bias is estimated as
\begin{equation}\label{eq:balb}
    B_{\text{ALB}}( \bullet ) = r - E_{x_{1:M}\sim \alpha(x_{1:M}; \mathcal{D}_{\text{pool}})}[\bullet] ,
\end{equation}
where $\bullet$ is either $\tilde{r}$ or $\tilde{r}_{\text{LURE}}$ and $r$ is computed on the whole $\mathcal{D}_{\text{test}}$. \Cref{fig:res_alb} shows that $\tilde{r}_{\text{LURE}}$ is unbiased, whereas the bias in $\tilde{r}$ approaches zero only as $M \to N$.


\subsection{Overfitting bias}\label{sec:res_bias_ofb}

To test the hypothesis that DUNs' prior over depth reduces overfitting in low data regimes, we follow the procedure of \citet{farquhar_statistical_2020} to isolate overfitting bias. We train a model with $M$ actively sampled points using $\tilde{R}_{\text{LURE}}$ as the training objective. After acquiring each new point, we evaluate
\begin{equation}\label{eq:bofb}
    B_{\text{OFB}} = r - \tilde{R}_{\text{LURE}} .
\end{equation}
The difference between the risk evaluated on the test set ($r$) and on the train set once the active learning bias is removed ($\tilde{R}_{\text{LURE}}$) is taken as a measure of overfitting ($B_{\text{OFB}}$).

\begin{figure}[h!] 
\centering
    \begin{subfigure}
        \centering
        \includegraphics[width=0.335\linewidth]{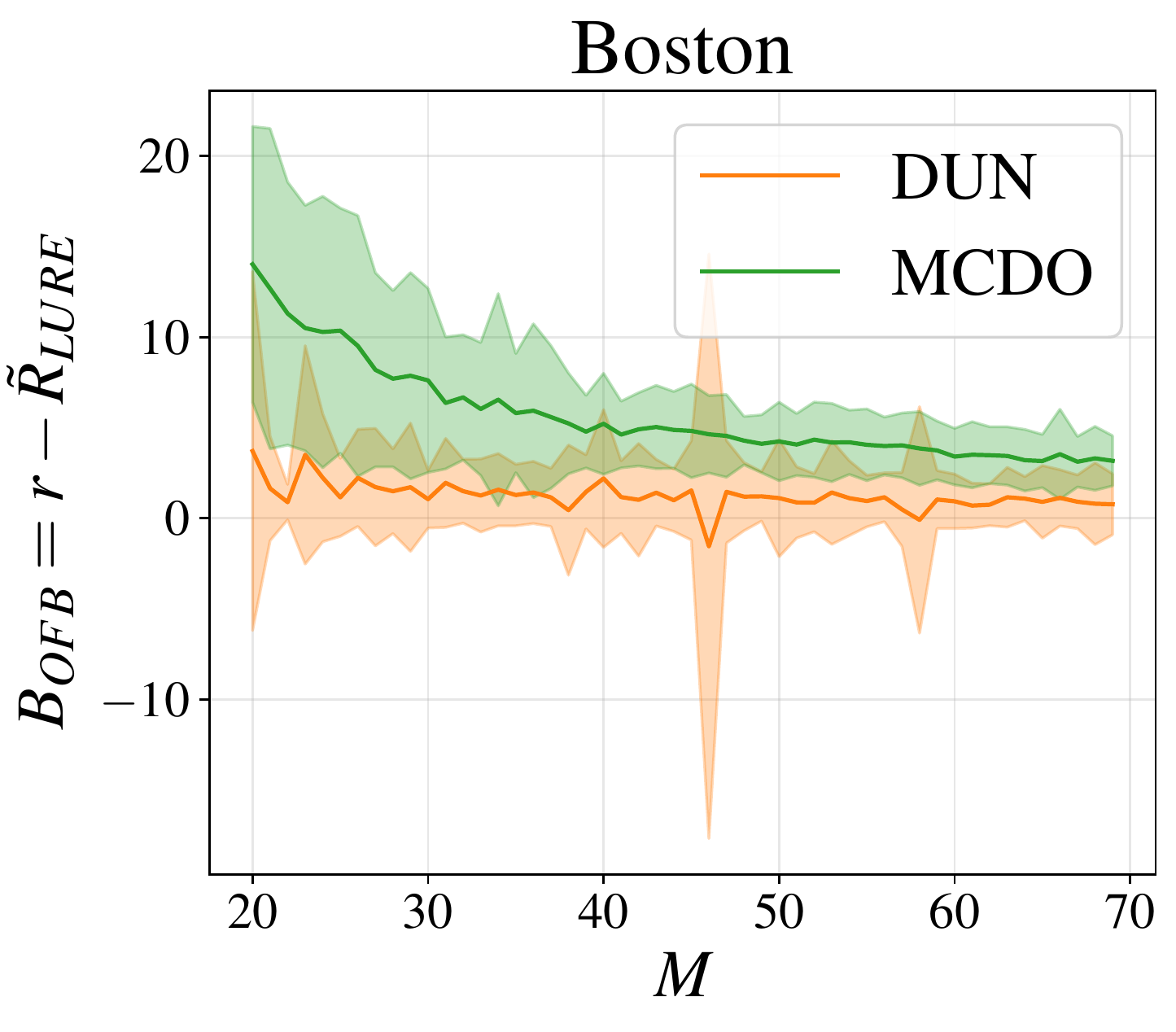}
    \end{subfigure}
    \begin{subfigure}
        \centering
        \includegraphics[width=0.315\linewidth]{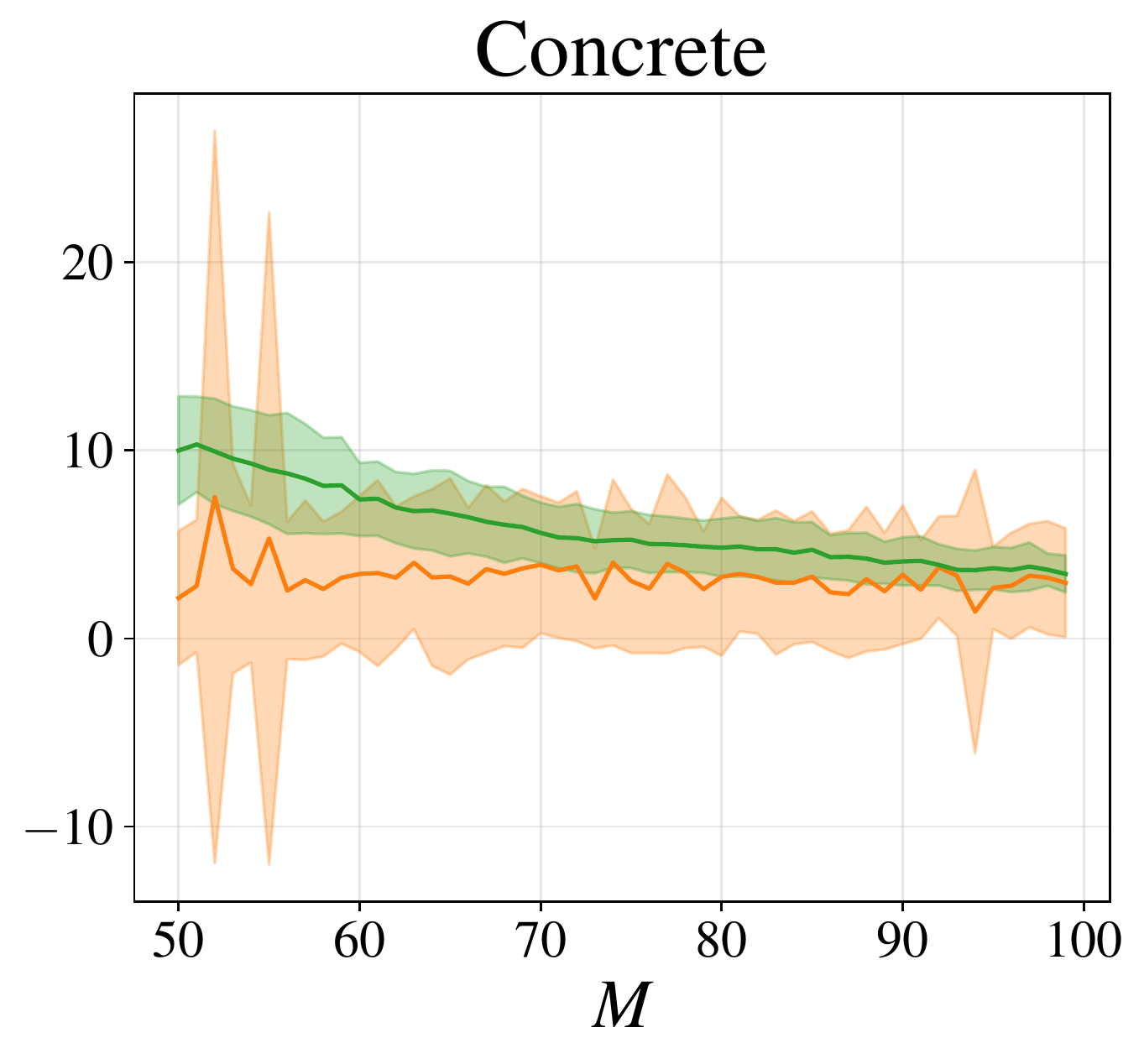}
    \end{subfigure} 
    \begin{subfigure}
        \centering
        \includegraphics[width=0.31\linewidth]{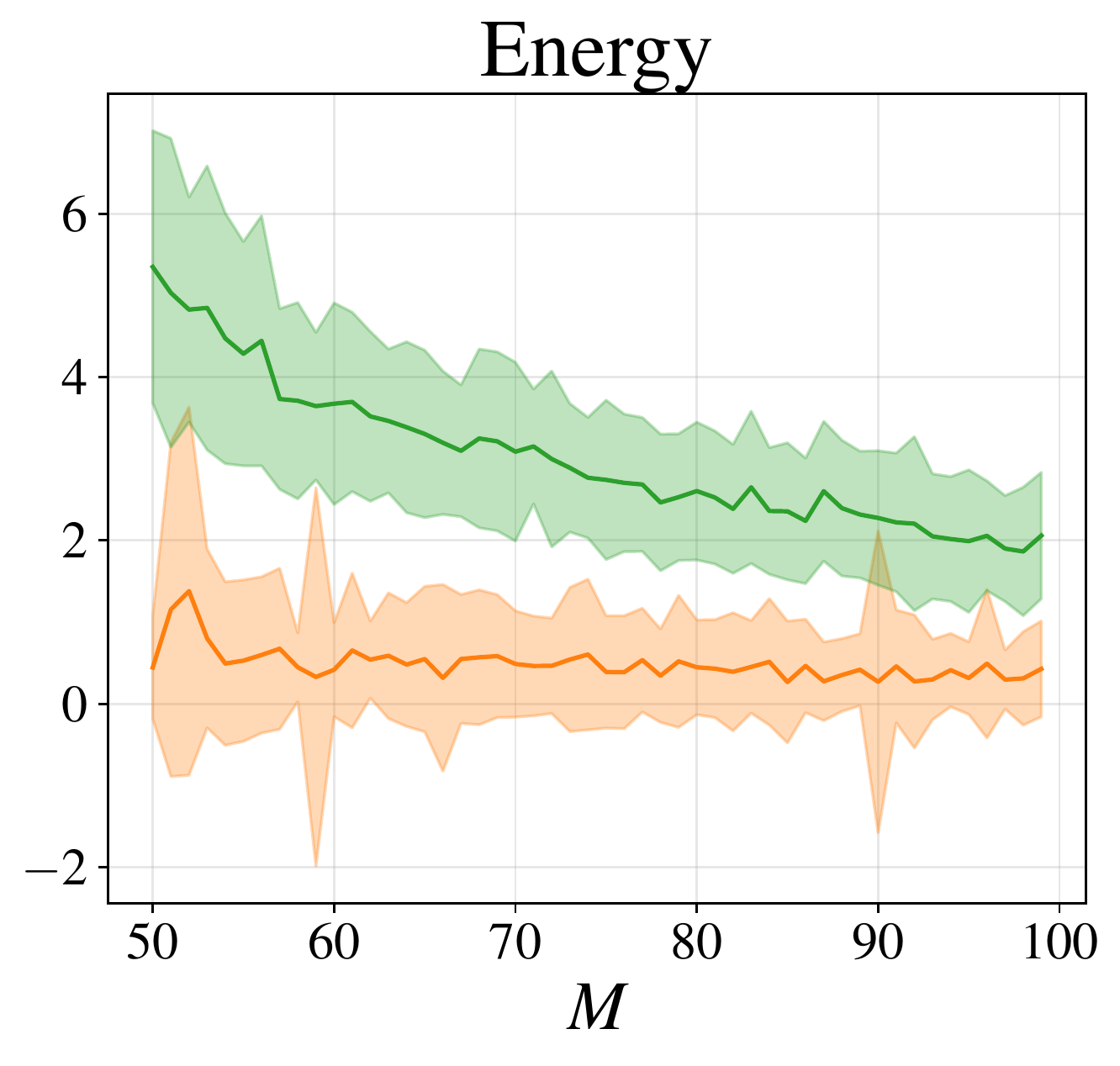} 
    \end{subfigure}
    \vspace{-5mm}
    \caption{Evolution of overfitting bias for DUNs (orange) and MCDO (green) trained with $\tilde{R}_{\text{LURE}}$ as the amount of acquired data $M$ grows. Overfitting bias is smaller in DUNs than for MCDO.}
    \label{fig:res_ofb_DUNvMCDO}
\end{figure}

\Cref{fig:res_ofb_DUNvMCDO} shows that the magnitude of $B_{\text{OFB}}$ for DUNs is small relative to BNNs trained with MCDO \citep{gal2016dropout}, one of the methods used in \citet{farquhar_statistical_2020}. This suggests that DUNs are, indeed,  robust to overfitting on small datasets. Overfitting bias in DUNs is also larger in magnitude than active learning bias for the datasets shown (comparing \cref{fig:res_ofb_DUNvMCDO} to \cref{fig:res_alb}). This is not the case for MCDO (comparing \cref{fig:res_ofb_DUNvMCDO} to \cref{fig:app_res_alb_mcdo}).  

\subsection{Training with unbiased risk estimators}\label{sec:res_bias_fig3}

\begin{figure}[h!] 
\centering
    \begin{subfigure}
        \centering
        \includegraphics[width=0.328\linewidth]{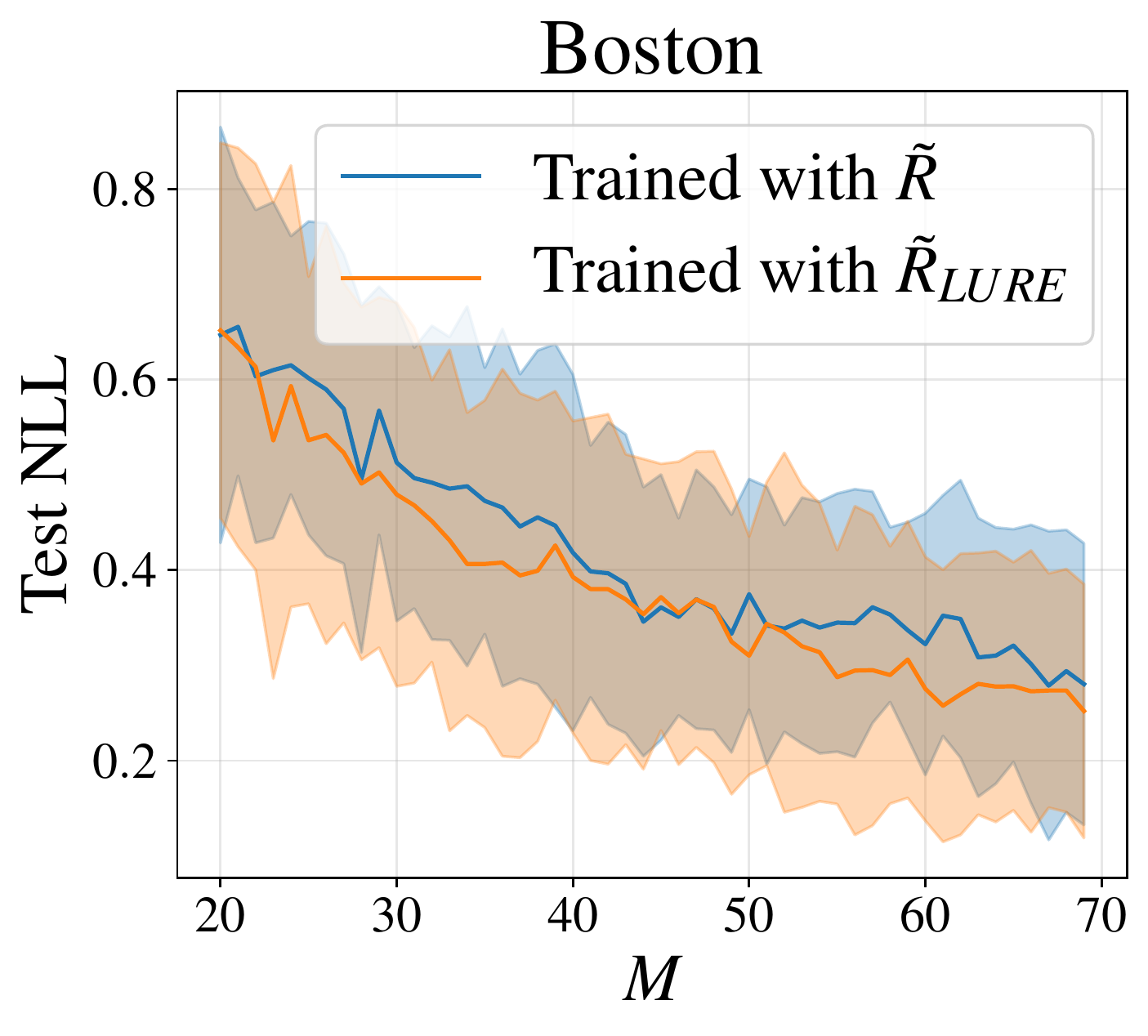}
    \end{subfigure}
    \begin{subfigure}
        \centering
        \includegraphics[width=0.312\linewidth]{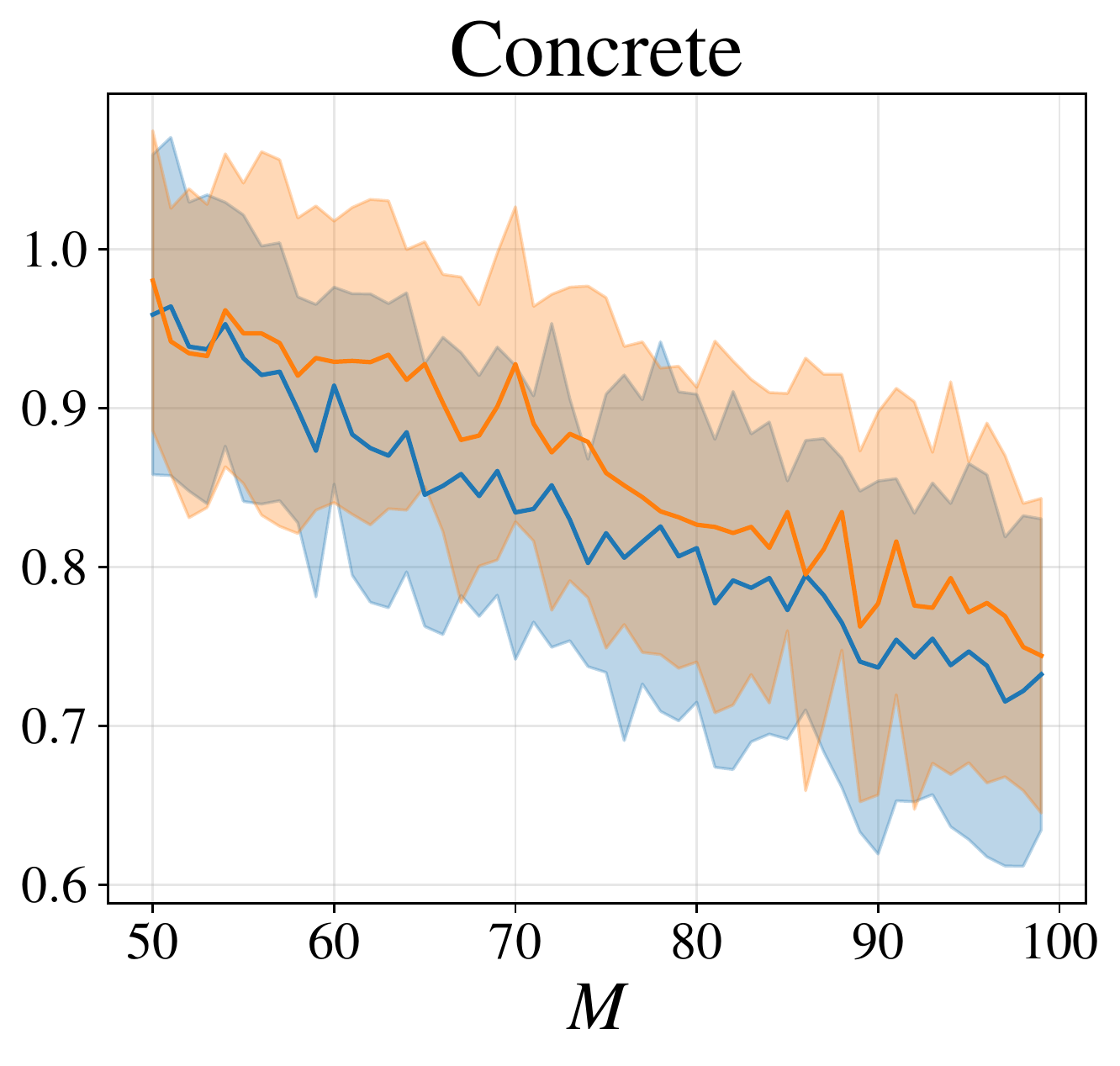}
    \end{subfigure} 
    \begin{subfigure}
        \centering
        \includegraphics[width=0.322\linewidth]{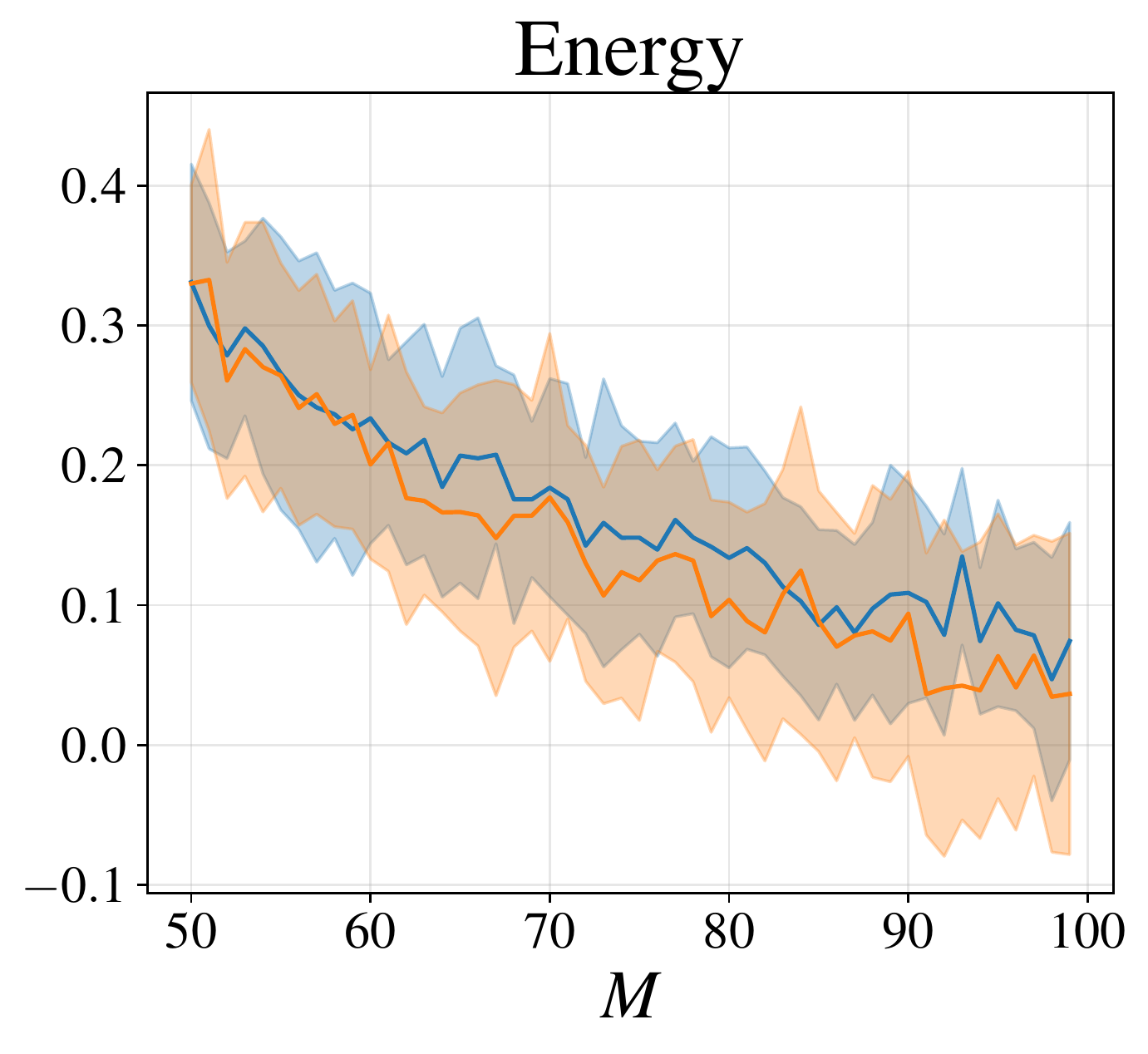} 
    \end{subfigure}
    \vspace{-5mm}
    \caption{Test NLL as more training points are acquired for DUNs trained with $\tilde{R}$ (blue) and $\tilde{R}_{\text{LURE}}$ (orange). Active learning bias correction has minimal impact on downstream performance.}
    \label{fig:res_fig3_uniform_nll}
\end{figure}

According to \citet{farquhar_statistical_2020}, removing the active learning bias in cases where ${\vert B_{\text{ALB}}\vert \gg \vert B_{\text{OFB}}\vert}$ should positively affect performance. Indeed, they observe this for linear models. Thus, having confirmed that the LURE weights eliminate active learning bias in DUNs, and that DUNs are in the $\vert{B_{\text{ALB}}\vert \gg \vert B_{\text{OFB}}}\vert$ regime, at least for certain datasets, we evaluate whether using the unbiased estimator during training improves downstream performance. We compare the NLL obtained by evaluating our models on $\mathcal{D}_{\text{test}}$ when using $\tilde{R}$ and $\tilde{R}_{\text{LURE}}$ as the training objectives. Surprisingly,  \cref{fig:res_fig3_uniform_nll} shows that using $\tilde{R}_{\text{LURE}}$ does not affect NLL performance at all. 

\section{Discussion}\label{sec:discussion}

In \cref{fig:res_fig3_uniform_nll}, we measure our models' generalisation error, given by $\mathtt{bias}^2 + \mathtt{variance}$~\citep{bishop2006pattern}. Considering that neural networks are very flexible, we expect model bias to be negligible, and active learning bias to dominate. When we eliminate the latter with importance weights, as in \cref{eq:r_lure}, we pay the price of additional variance \citep{mcbook}.
We therefore hypothesise that our results can be explained by gains due to bias reduction being lost due to increased variance. This increase in variance will depend on the choice of sampling distribution $\alpha$.
``Overfitting bias'', as presented in \cref{eq:bofb} and \citep{farquhar_statistical_2020} is computed using the train set and thus is not directly comparable with active learning bias in \cref{eq:balb}. 
Furthermore, the effect of ``overfitting bias'' is captured by model variance. The error induced by model variance is always additive with any statistical biases that may exist, precluding any cancellation effects.
This hypothesis also explains why \citet{farquhar_statistical_2020} find $\tilde{R}_{\text{LURE}}$ to be successful for linear models: since inflexible models are well-specified by small amounts of data, applying $\tilde{R}_{\text{LURE}}$ barely increases variance.
We also highlight that in the experimental design used by \citet{farquhar_statistical_2020} to obtain their linear model results, the temperature of the proposal distribution ($T=10,000$) is much larger than that used here ($T=10$). Higher temperatures approach a deterministic proposal, interfering with the assumption that $\alpha(\cdot)$ has full support.

\clearpage

\section*{Acknowledgments}
JA acknowledges support from Microsoft Research, through its PhD Scholarship Programme, and from the EPSRC.
JUA acknowledges funding from the EPSRC and the Michael E. Fisher Studentship in Machine Learning.
CM acknowledges assistance from Sebastian Farquhar in clarifying the implementation of the active learning bias experiments.
This work has been performed using resources provided by the Cambridge Tier-2 system operated by the University of Cambridge Research Computing Service (http://www.hpc.cam.ac.uk) funded by EPSRC Tier-2 capital grant EP/P020259/1.

\bibliographystyle{plainnat}
\bibliography{references}

\begin{thebibliography}{14}
\providecommand{\natexlab}[1]{#1}
\providecommand{\url}[1]{\texttt{#1}}
\expandafter\ifx\csname urlstyle\endcsname\relax
  \providecommand{\doi}[1]{doi: #1}\else
  \providecommand{\doi}{doi: \begingroup \urlstyle{rm}\Url}\fi

\bibitem[Antor{\'a}n et~al.(2020)Antor{\'a}n, Allingham, and
  Hern{\'a}ndez-Lobato]{antoran2020depth}
Javier Antor{\'a}n, James Allingham, and Jos{\'e}~Miguel Hern{\'a}ndez-Lobato.
\newblock Depth uncertainty in neural networks.
\newblock \emph{Advances in Neural Information Processing Systems}, 33, 2020.

\bibitem[Bishop(2007)]{bishop2006pattern}
Christopher~M. Bishop.
\newblock \emph{Pattern Recognition and Machine Learning, 5th Edition}.
\newblock Information science and statistics. Springer, 2007.

\bibitem[Cohn et~al.(1995)Cohn, Ghahramani, and Jordan]{cohn1995active}
David~A Cohn, Zoubin Ghahramani, and Michael~I Jordan.
\newblock Active learning with statistical models.
\newblock Technical report, MASSACHUSETTS INST OF TECH CAMBRIDGE ARTIFICIAL
  INTELLIGENCE LAB, 1995.

\bibitem[D’Amour et~al.(2020)D’Amour, Heller, Moldovan, Adlam, Alipanahi,
  Beutel, Chen, Deaton, Eisenstein, Hoffman, et~al.]{d2020underspecification}
A~D’Amour, K~Heller, D~Moldovan, B~Adlam, B~Alipanahi, A~Beutel, C~Chen,
  J~Deaton, J~Eisenstein, MD~Hoffman, et~al.
\newblock Underspecification presents challenges for credibility in modern
  machine learning. arxiv 2020.
\newblock \emph{arXiv preprint arXiv:2011.03395}, 2020.

\bibitem[Farquhar et~al.(2021)Farquhar, Gal, and
  Rainforth]{farquhar_statistical_2020}
Sebastian Farquhar, Yarin Gal, and Tom Rainforth.
\newblock On statistical bias in active learning: How and when to fix it.
\newblock \emph{International Conference on Learning Representations}, 2021.

\bibitem[Gal and Ghahramani(2016)]{gal2016dropout}
Yarin Gal and Zoubin Ghahramani.
\newblock Dropout as a {B}ayesian approximation: Representing model uncertainty
  in deep learning.
\newblock In \emph{International Conference on Machine Learning}, pages
  1050--1059. PMLR, 2016.

\bibitem[Ghosh et~al.(2019)Ghosh, Yao, and Doshi-Velez]{JMLR:v20:19-236}
Soumya Ghosh, Jiayu Yao, and Finale Doshi-Velez.
\newblock Model selection in bayesian neural networks via horseshoe priors.
\newblock \emph{Journal of Machine Learning Research}, 20\penalty0
  (182):\penalty0 1--46, 2019.
\newblock URL \url{http://jmlr.org/papers/v20/19-236.html}.

\bibitem[Hern{\'a}ndez-Lobato and Adams(2015)]{hernandez2015probabilistic}
Jos{\'e}~Miguel Hern{\'a}ndez-Lobato and Ryan Adams.
\newblock Probabilistic backpropagation for scalable learning of {B}ayesian
  neural networks.
\newblock In \emph{International Conference on Machine Learning}, pages
  1861--1869. PMLR, 2015.

\bibitem[Houlsby et~al.(2011)Houlsby, Huszar, Ghahramani, and
  Lengyel]{houlsby2011bayesian}
Neil Houlsby, Ferenc Huszar, Zoubin Ghahramani, and M{\'{a}}t{\'{e}} Lengyel.
\newblock Bayesian active learning for classification and preference learning.
\newblock \emph{CoRR}, abs/1112.5745, 2011.
\newblock URL \url{http://arxiv.org/abs/1112.5745}.

\bibitem[Ioffe and Szegedy(2015)]{ioffe2015batch}
S.~Ioffe and C.~Szegedy.
\newblock Batch normalization: Accelerating deep network training by reducing
  internal covariate shift.
\newblock \emph{arXiv preprint arXiv:1502.03167}, 2015.

\bibitem[MacKay(1992)]{mackay1992information}
David~JC MacKay.
\newblock Information-based objective functions for active data selection.
\newblock \emph{Neural computation}, 4\penalty0 (4):\penalty0 590--604, 1992.

\bibitem[Owen(2013)]{mcbook}
Art~B. Owen.
\newblock \emph{Monte Carlo theory, methods and examples}.
\newblock 2013.

\bibitem[Settles(2010)]{settles2009active}
Burr Settles.
\newblock Active learning literature survey.
\newblock \emph{Machine Learning}, 15\penalty0 (2):\penalty0 201--221, 2010.

\bibitem[Zeiler and Fergus(2014)]{zeiler2014visualizing}
Matthew~D. Zeiler and Rob Fergus.
\newblock Visualizing and understanding convolutional networks.
\newblock In David~J. Fleet, Tom{\'{a}}s Pajdla, Bernt Schiele, and Tinne
  Tuytelaars, editors, \emph{Computer Vision - {ECCV} 2014 - 13th European
  Conference, Zurich, Switzerland, September 6-12, 2014, Proceedings, Part
  {I}}, volume 8689 of \emph{Lecture Notes in Computer Science}, pages
  818--833. Springer, 2014.

\end{thebibliography}


\clearpage
\appendix
\renewcommand\thefigure{\thesection.\arabic{figure}}  
\renewcommand\thetable{\thesection.\arabic{table}}

\section{Depth uncertainty networks}\label{app:duns}

BNNs translate uncertainty in weight-space into predictive uncertainty by marginalising out the posterior over weights. The intractability of this posterior necessitates the use of approximate inference methods. DUNs, in contrast, leverage uncertainty about the appropriate \textit{depth} of the network in order to obtain predictive uncertainty estimates. There are two primary advantages of this approach: 1) the posterior over depth is tractable, mitigating the need for limiting approximations; and 2) due to the sequential structure of feed-forward neural networks, both inference and prediction can be performed with a single forward pass, making DUNs suitable for deployment in low-resource environments \citep{antoran2020depth}. 

\begin{figure}[h!]
\centering
    \begin{tikzpicture}[every node/.style={scale=0.85}]
    	\node[block=depth1] (RB0) {$f_{0}$};
    	\node[block=depth2, below right=.0ex and 4ex of RB0] (RB1) {$f_{1}$};
    	\node[block=depth3, below right=.0ex and 4ex of RB1] (RB2) {$f_{2}$};
    	\node[block=depth4, below right=.0ex and 4ex of RB2] (RB3) {$f_{3}$};
    	\node[block=depth5, below right=.0ex and 4ex of RB3] (RB4) {$f_{4}$};
    	\node[block=teal!80!white, below right=2ex and 7ex of RB4, text width=2.5ex] (RBD) {$f_{D}$};

    	\node[left=4ex of RB0] (X) {$\mathbf{x}$};

    	\draw[arrow] (X) -- (RB0);
    	\draw[arrow] (RB0) |- (RB1);
    	\draw[arrow] (RB1) |- (RB2);
    	\draw[arrow] (RB2) |- (RB3);
    	\draw[arrow] (RB3) |- (RB4);
    	\draw[line] (RB4) -- ([yshift=-.5ex] RB4.south);
    	\draw[line, dashed] ([yshift=-.5ex] RB4.south) |- ([xshift=-4.ex] RBD.west);
    	\draw[arrow] ([xshift=-4.ex] RBD.west) -- (RBD);
    
        \draw[arrow, snake] (RBD) -- ([xshift=4ex] RBD.east) node[coordinate] (end) {};
        \draw[arrow, snake] (RB0) -- (RB0 -| end);
        \draw[arrow, snake] (RB1) -- (RB1 -| end);
        \draw[arrow, snake] (RB2) -- (RB2 -| end);
        \draw[arrow, snake] (RB3) -- (RB3 -| end);
        \draw[arrow, snake] (RB4) -- (RB4 -| end);
        
        \draw [decorate,decoration={brace,amplitude=1ex, raise=.5ex},xshift=2ex, thick] (RB0 -| end.west) -- (end) node [block=blue!60!white, midway, xshift=6ex, text width=5ex, thin] (OB) {$f_{D+1}$};
        
    	\node[right=4ex of OB] (YD) {$\mathbf{\hat y}_{i}$};
    	\draw[arrow] (OB.east) -- (YD);
    \end{tikzpicture}
\caption[DUN computational model.]{DUN computational model \citep{antoran2020depth}. Each layer's activations are passed through the output block, producing per-depth predictions.}\label{fig:dun_comp_model}
\end{figure}
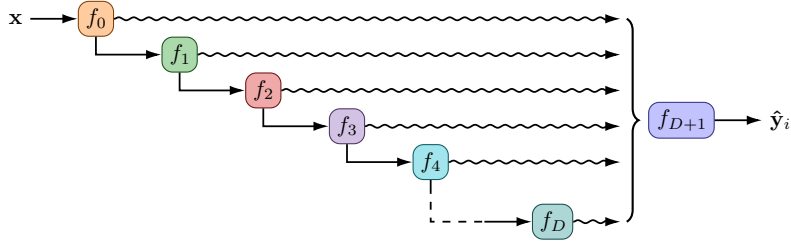

DUNs are composed of subnetworks of increasing depth, with each subnetwork contributing one prediction to the final model. The computational model for DUNs is shown in \cref{fig:dun_comp_model}. Inputs are passed through an input block, $f_0(\cdot)$, and then sequentially through each of $D$ intermediate blocks $\{f_i(\cdot)\}_{i=1}^{D}$, with the activations of the previous layer acting as the inputs of the current layer. The activations of each of the $D+1$ layers are passed through the output block $f_{D+1}(\cdot)$ to generate per-depth predictions. These are combined via marginalisation over the depth posterior, equivalent to forming an ensemble with networks of increasing depth. Variation between the predictions from each depth can be used to obtain predictive uncertainty estimates, in much the same way that variance in the predictions of different sampled weight configurations yield predictive uncertainties in BNNs.

\subsection{Uncertainty over depth}

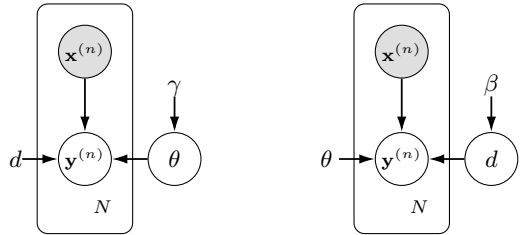
\begin{figure}[h]
\vspace{-0.1in}
\centering
\begin{subfigure}
    \centering
    \begin{tikzpicture}[every node/.style={scale=1}]
      \node[obs] (xn) {\scriptsize $\mathbf{x}^{(n)}$};
      \node[latent, below=4.5ex of xn] (yn) {\scriptsize $\mathbf{y}^{(n)}$};
      \node[latent, right=3ex of yn] (theta) {\footnotesize $\theta$};
      \node[const, above=3ex of theta] (gamma) {\footnotesize $\gamma$};
      \node[const, left=3ex of yn] (d) {\footnotesize $d$};
      
      \edge[arrow] {xn} {yn} ; %
      \edge[arrow] {theta} {yn} ;
      \edge[arrow] {gamma} {theta} ;
      \edge[arrow] {d} {yn}
      
      \plate[inner sep=1.7ex] {} {(yn)(xn)} {\scriptsize $N$} ;
    \end{tikzpicture}
\end{subfigure}\
\qquad \qquad
\begin{subfigure}
    \centering
    \begin{tikzpicture}[every node/.style={scale=1}]
      \node[obs] (xn) {\scriptsize $\mathbf{x}^{(n)}$};
      \node[latent, below=4.5ex of xn] (yn) {\scriptsize $\mathbf{y}^{(n)}$};
      \node[const, left=3ex of yn] (theta) {\footnotesize $\theta \ $};
      \node[latent, right=3ex of yn] (d) {\footnotesize $d$};
      \node[const, above=3ex of d] (beta) {\footnotesize $\beta$};
      
      \edge[arrow] {xn} {yn} ; %
      \edge[arrow] {theta} {yn} ;
      \edge[arrow] {d} {yn} ;
      \edge[arrow] {beta} {d} ;
      
      \plate[inner sep=1.7ex] {} {(yn)(xn)} {\scriptsize $N$} ;
    \end{tikzpicture}
\end{subfigure}
\caption[Graphical models of BNNs and DUNs.]{Left: graphical model of a BNN. Right: graphical model of a DUN.}
\label{fig:dun_graph_model}
\end{figure}

\Cref{fig:dun_graph_model} compares the graphical model representations of a BNN and a DUN. In BNNs, the weights $\theta$ are treated as random variables drawn from a distribution $p_{\gamma}(\theta)$ with hyperparameters $\gamma$. In DUNs, the depth of the network $d$ is assumed to be stochastic, while the weights are learned as hyperparameters. A categorical prior distribution is assumed for $d$, with hyperparameters $\beta$: $p_{\beta}(d)=\operatorname{Cat}(d \mid \{\beta_i \}_{i=0}^D)$. The model's marginal log likelihood (MLL) is given by
\begin{equation}
\log p(\mathcal{D}_{\text{train}} ; \theta)=\log \sum_{i=0}^{D}\left(p_{\beta}(d=i) \prod_{n=1}^{N} p\left(\mathbf{y}_n \mid \mathbf{x}_n, d=i ; \theta\right)\right), \label{eq:dun_mll}
\end{equation}
where the likelihood for each depth is parameterised by the corresponding subnetwork's output: $p(\mathbf{y} \mid \mathbf{x}, d=i ; \theta)=p\left(\mathbf{y} \mid f_{D+1}\left(\mathbf{a}_{i} ; \theta\right)\right)$, where  $\mathbf{a}_{i}=f_i(\mathbf{a}_{i-1})$. \\

The posterior, 
\begin{equation}
    p(d \mid \mathcal{D}_{\text{train}} ; \theta)=\frac{p(\mathcal{D}_{\text{train}} \mid d ; \theta) p_{\beta}(d)}{p(\mathcal{D}_{\text{train}} ; \theta)}, \label{eq:depth_post}
\end{equation}

is also a categorical distribution, whose probabilities reflect how well each depth subnetwork explains the data. DUNs leverage two properties of modern neural networks: first, that successive layers have been shown to extract features at increasing levels of abstraction \citep{zeiler2014visualizing}; and second, that networks are typically underspecified, meaning that several different models may explain the data well \citep{d2020underspecification}. The first property implies that initial layers in a network specialise on low-level feature extraction, yielding poor predictions from the shallower subnetworks in a DUN. This is handled by the depth posterior assigning low probabilities to earlier layers, in preference for later layers that specialise on prediction. The second property suggests that subnetworks at several depths are able to explain the data simultaneously and thus have non-zero posterior probabilities, yielding the diversity in predictions required to obtain useful estimates of model uncertainty. 

\subsection{Inference in DUNs}

\citet{antoran2020depth} find that directly optimising the MLL \cref{eq:dun_mll} with respect to $\theta$ results in convergence to a local optimum in which all but one layer is attributed a posterior probability of zero. That is, direct optimisation returns a deterministic network of arbitrary depth. The authors instead propose a stochastic gradient variational inference (VI) approach with the aim of separating optimisation of the weights $\theta$ from the posterior. A surrogate categorical distribution $q_{\phi}(d)$ is introduced as the variational posterior. The following evidence lower bound (ELBO) can be derived:
\begin{equation}
    \mathcal{L}(\phi, \theta)=\sum_{n=1}^{N} \mathbb{E}_{q_{\phi}(d)}\left[\log p\left(\mathbf{y}_n \mid \mathbf{x}_n, d ; \theta\right)\right]-\mathrm{KL}\left(q_{\phi}(d) \| p_{\beta}(d)\right) , \label{eq:dun_elbo}
\end{equation}
allowing $\theta$ and the variational parameters $\phi$ to be optimised simultaneously. \citet{antoran2020depth} show that under the VI approach, the variational parameters converge more slowly than the weights, enabling the weights to reach a setting at which a positive posterior probability is learnt for several depths.

It is important to note that \cref{eq:dun_elbo} can be computed exactly, and that optimising $\mathcal{L}(\phi, \theta)$ is equivalent to exact inference of the true posterior $p(d \mid \mathcal{D}_{\text{train}}; \theta)$ in the limit: since both $q$ and $p$ are categorical, \cref{eq:dun_elbo} is convex and tight at the optimum (i.e., $q_{\phi}(d)=p(d \mid \mathcal{D}_{\text{train}}; \theta)$). Since the expectation term can be computed exactly using the activations at each layer (recall $p(\mathbf{y} \mid \mathbf{x}, d=i ; \theta)=p\left(\mathbf{y} \mid f_{D+1}\left(\mathbf{a}_{i} ; \theta\right)\right)$), the ELBO can be evaluated exactly without Monte Carlo sampling.

\clearpage
\section{Experimental setup}\label{app:exp_setup}

\subsection{Data}\label{app:data}

We implement batch-based active learning, with batches of $20$ data points acquired in each query (with the exception of Yacht, which has a query size of 10 due to its smaller size). An initial training set representing a small proportion of the full data is selected uniformly at random in the first query. For all datasets, 80\% of the data are used for training, 10\% for validation and 10\% for testing. Details about dataset sizes, input dimensionality, and active learning specifications for each dataset are provided in \cref{tab:datasets}.

\setcounter{table}{0}    
\begin{table}[h]
    \centering
    \caption[Summary of datasets and active learning specifications.]{Summary of datasets and active learning specifications. 80\% of the data are used for training, 10\% for validation and 10\% for testing.}
    \resizebox{\textwidth}{!}{\begin{tabular}{l|ccccc}
    \toprule
         \textsc{Name} & \textsc{Size} & \textsc{Input Dim.} & \textsc{Init. train size} & \textsc{No. queries} & \textsc{Query size} \\
         \midrule
         Boston Housing & 506 & 13 & 20 & 17 & 20 \\
         Concrete Strength & 1,030 & 8 & 50 & 30 & 20 \\
         Energy Efficiency & 768 & 8 & 50 & 30 & 20 \\
         Kin8nm & 8,192 & 8 & 50 & 30 & 20 \\
         Naval Propulsion & 11,934 & 16 & 50 & 30 & 20 \\
         Power Plant & 9,568 & 4 & 50 & 30 & 20 \\ 
         Protein Structure & 45,730 & 9 & 50 & 30 & 20 \\
         Wine Quality Red & 1,599 & 11 & 50 & 30 & 20 \\
         Yacht Hydrodynamics & 308 & 6 & 20 & 20 & 10 \\
         \bottomrule
    \end{tabular}}
    \label{tab:datasets}
\end{table}

\subsection{Model specifications}\label{app:model_spec}

We implement a fully-connected network with residual connections, with 100 hidden nodes per layer. The networks contain 10 hidden layers for DUNs, or three hidden layers for MCDO. We use ReLU activations, and for DUNs batch normalisation is applied after every layer \citep{ioffe2015batch}. Optimisation is performed over $1,000$ iterations with full-batch gradient descent, with momentum of $0.9$ and a learning rate of $10^{-4}$. A weight decay value of $10^{-5}$ is also used. We do not implement early stopping, but the best model based on evaluation of the evidence lower bound on the validation set is used for reporting final results. For MCDO models a fixed dropout probability of $0.1$ is used and prediction is based on 10 MC samples. DUNs use a uniform categorical prior over depth. The hyperparameter settings largely follow those used in \citet{antoran2020depth} for their experiments with toy regression datasets.

All experiments are repeated 40 times with different weight initialisations and train-test splits (with the exception of the Protein dataset, for which experiments are repeated only 30 times due to the cost of evaluation on the larger test set). Unless otherwise specified, we report the mean and standard deviation of the relevant metric over the repeated experiment runs. Following \citet{farquhar_statistical_2020}, the proposal distribution $\alpha\left(i_{m} ; i_{1: m-1}, \mathcal{D}_{\text {pool }}\right)$ used is a stochastic relaxation of the Bayesian active learning by disagreement (BALD) acquisition function proposed by \citep{houlsby2011bayesian}. BALD scores are multiplied by a constant temperature $T$ before being passed through the \verb|softmax| function; candidate points are then sampled with probability proportional to the \verb|softmax| probabilities. We use $T=10$ as this is found to yield the best performance for most of the datasets considered (results for experiments to determine the optimal value of $T$ are given in \cref{app:temp}). 

\clearpage
\section{Additional experimental results}\label{app:add_results}

\subsection{Quantifying active learning bias}\label{app:res_bias_alb}

\Cref{fig:app_res_alb} presents the results of the active learning bias experiments, as in \cref{fig:res_alb}, for the remaining datasets. \Cref{fig:app_res_alb_mcdo} shows equivalent results for BNNs trained with MCDO.

\setcounter{figure}{0} 
\begin{figure}[h!] 
\centering
    \begin{subfigure}
        \centering
        \includegraphics[width=0.335\linewidth]{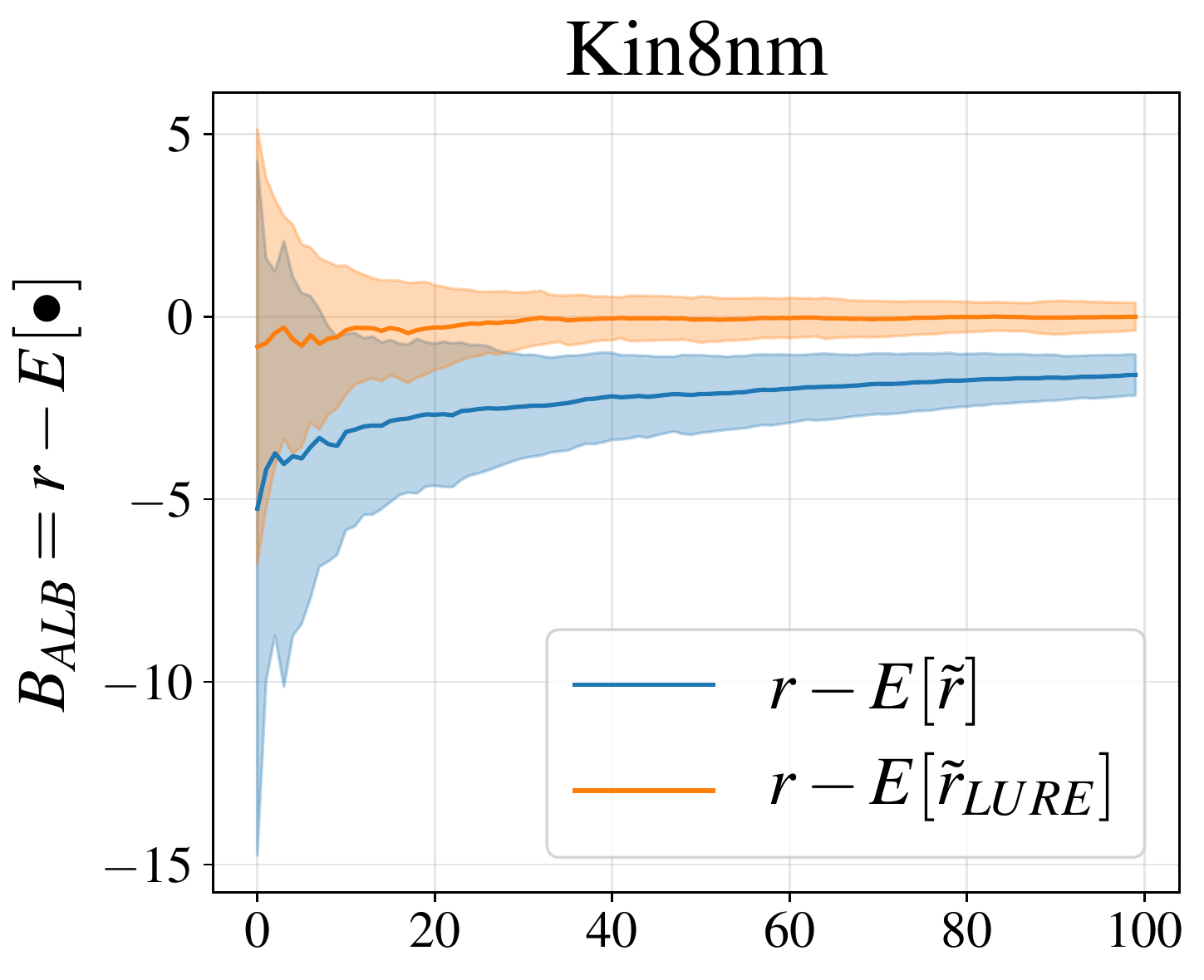}
    \end{subfigure} 
    \begin{subfigure}
        \centering
        \includegraphics[width=0.315\linewidth]{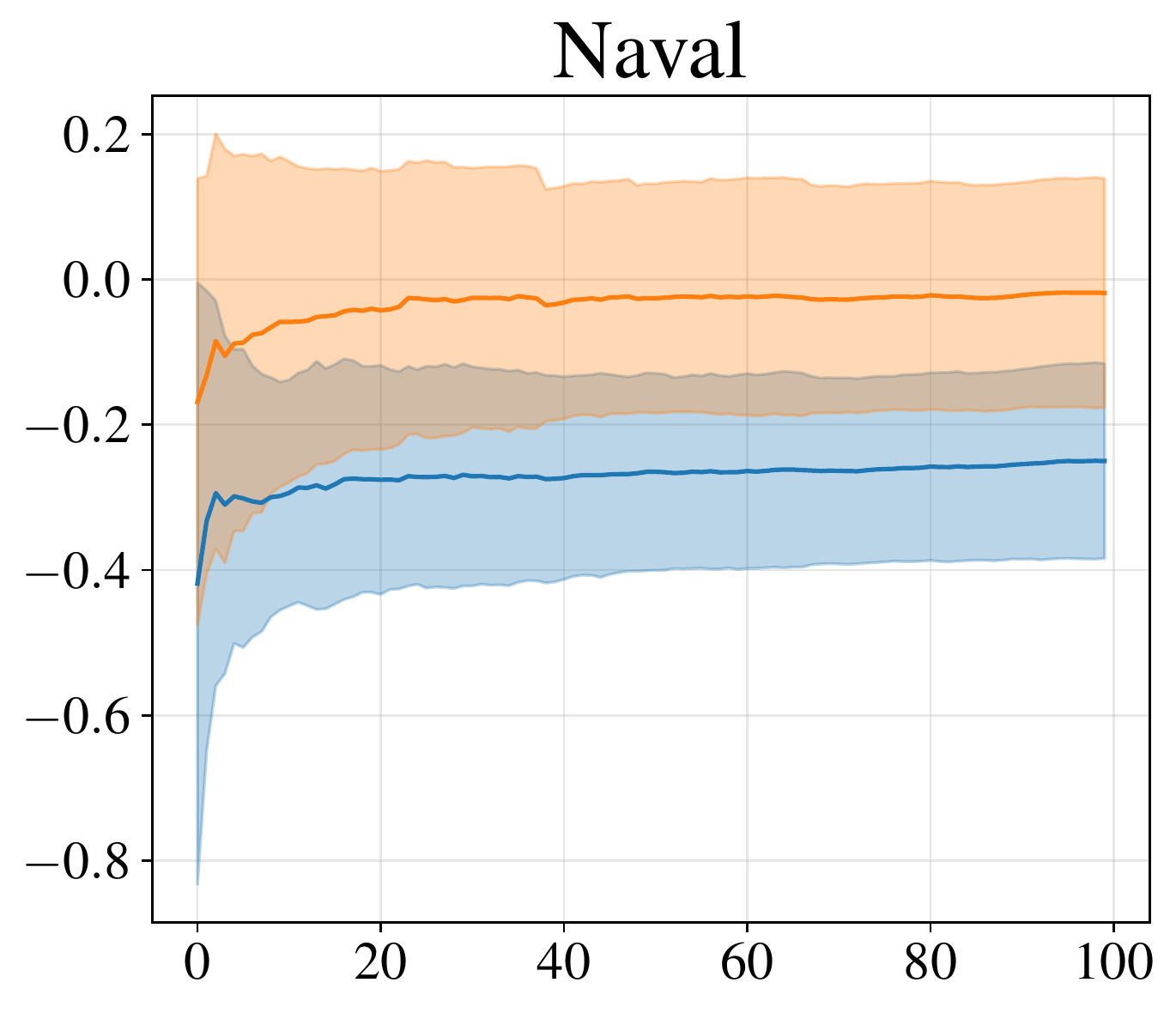}
    \end{subfigure} 
    \begin{subfigure}
        \centering
        \includegraphics[width=0.31\linewidth]{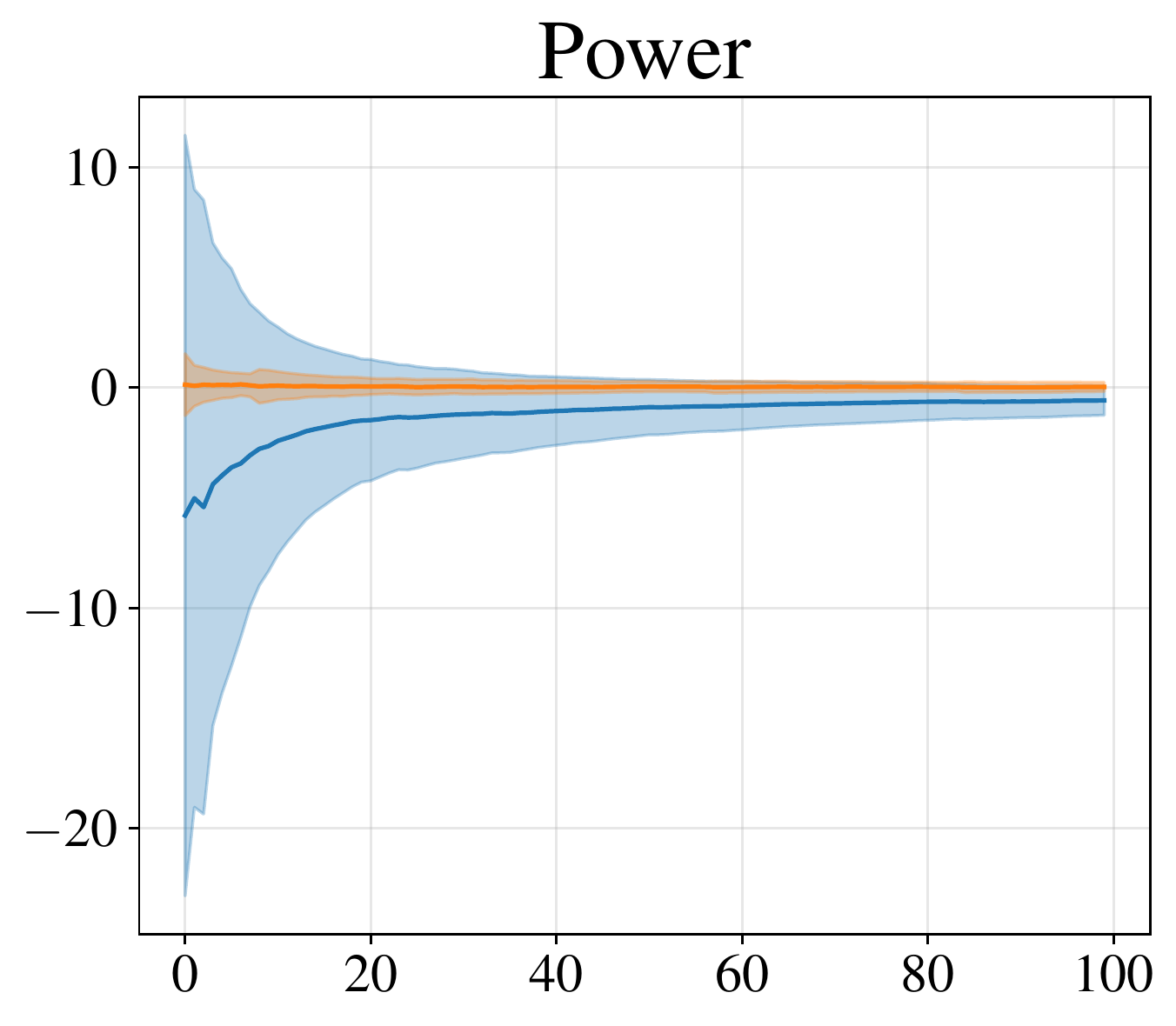}
    \end{subfigure} \\
    \begin{subfigure}
        \centering
        \includegraphics[width=0.345\linewidth]{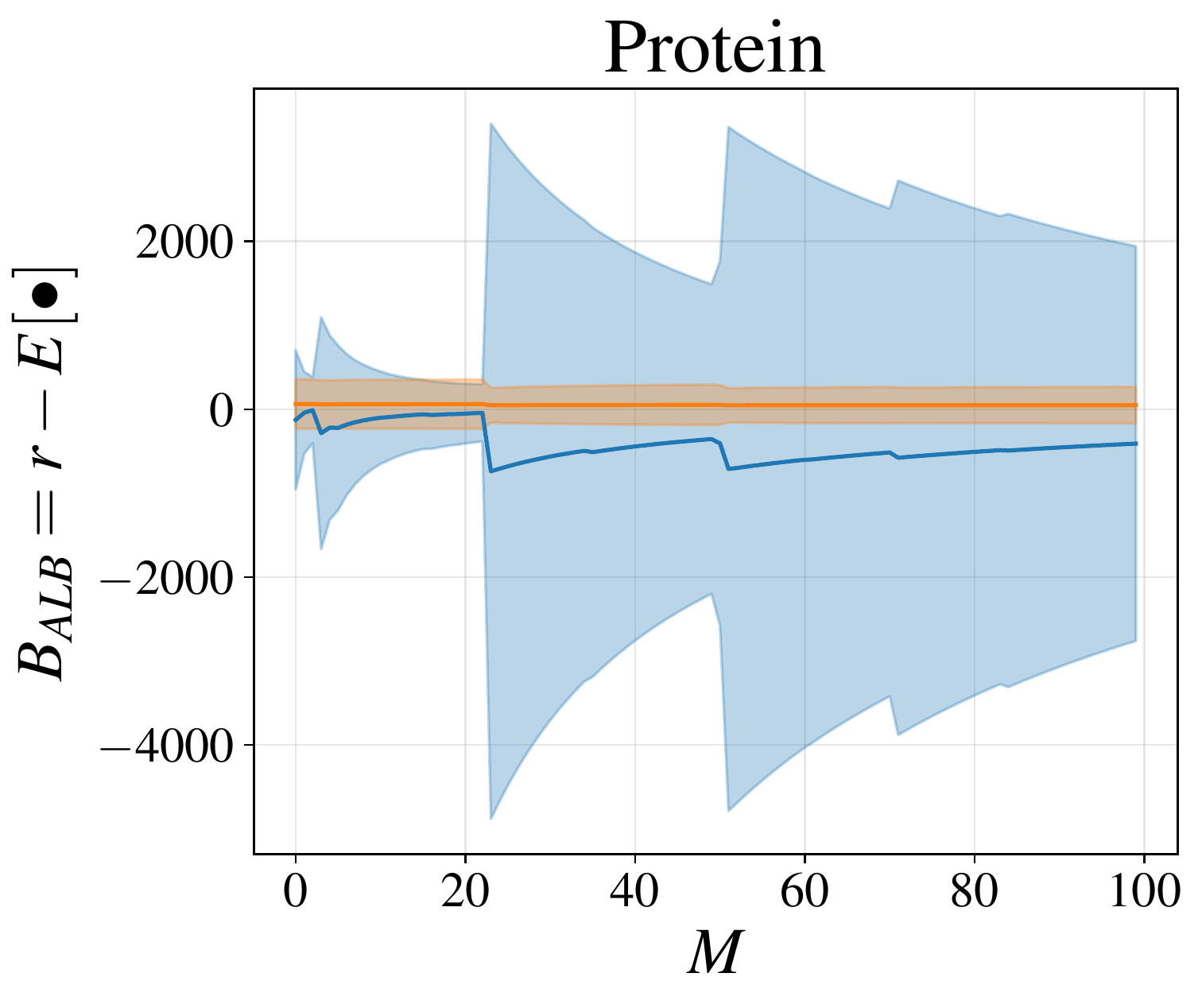}
    \end{subfigure} 
    \begin{subfigure}
        \centering
        \includegraphics[width=0.31\linewidth]{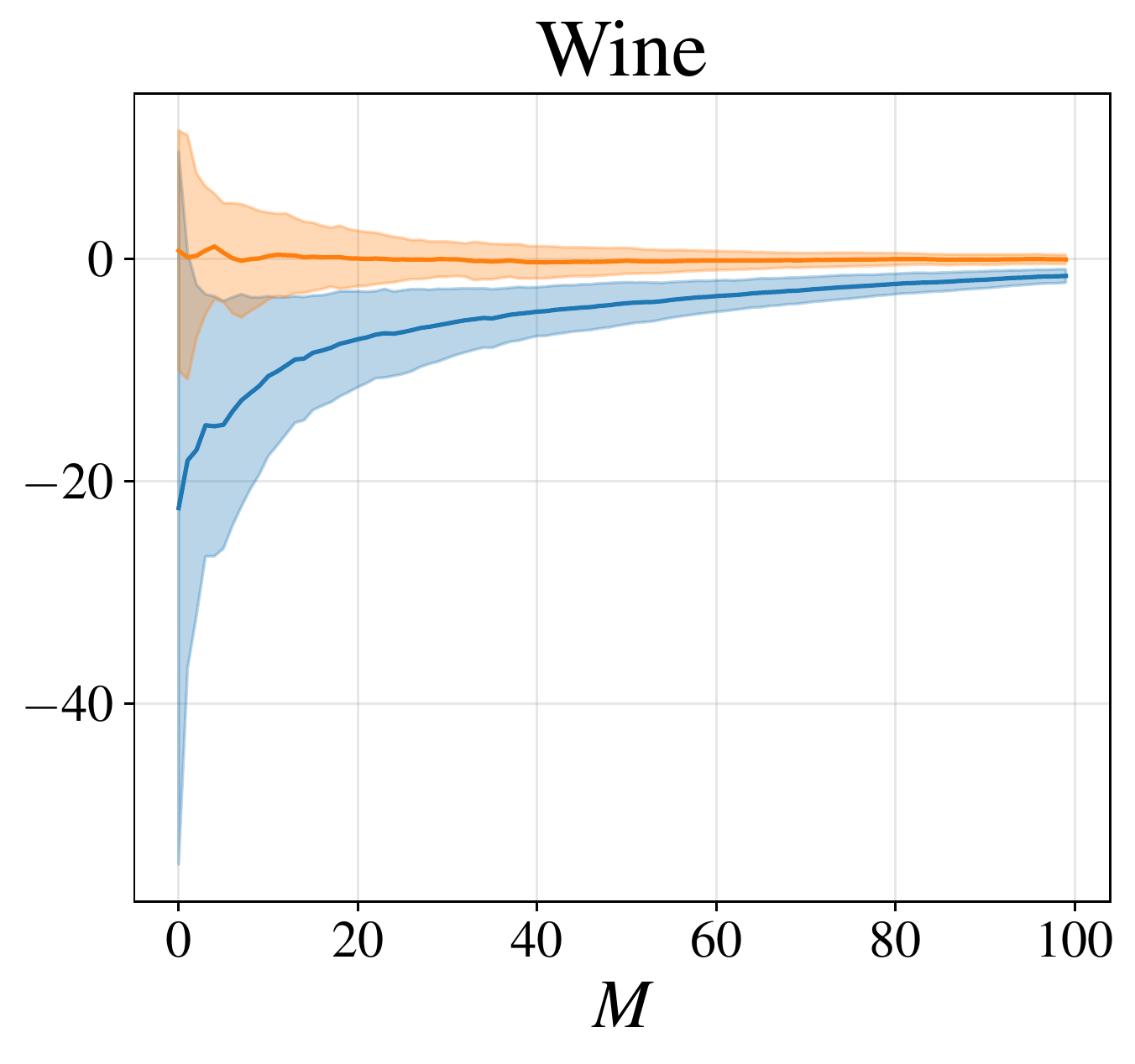}
    \end{subfigure} 
    \begin{subfigure}
        \centering
        \includegraphics[width=0.3075\linewidth]{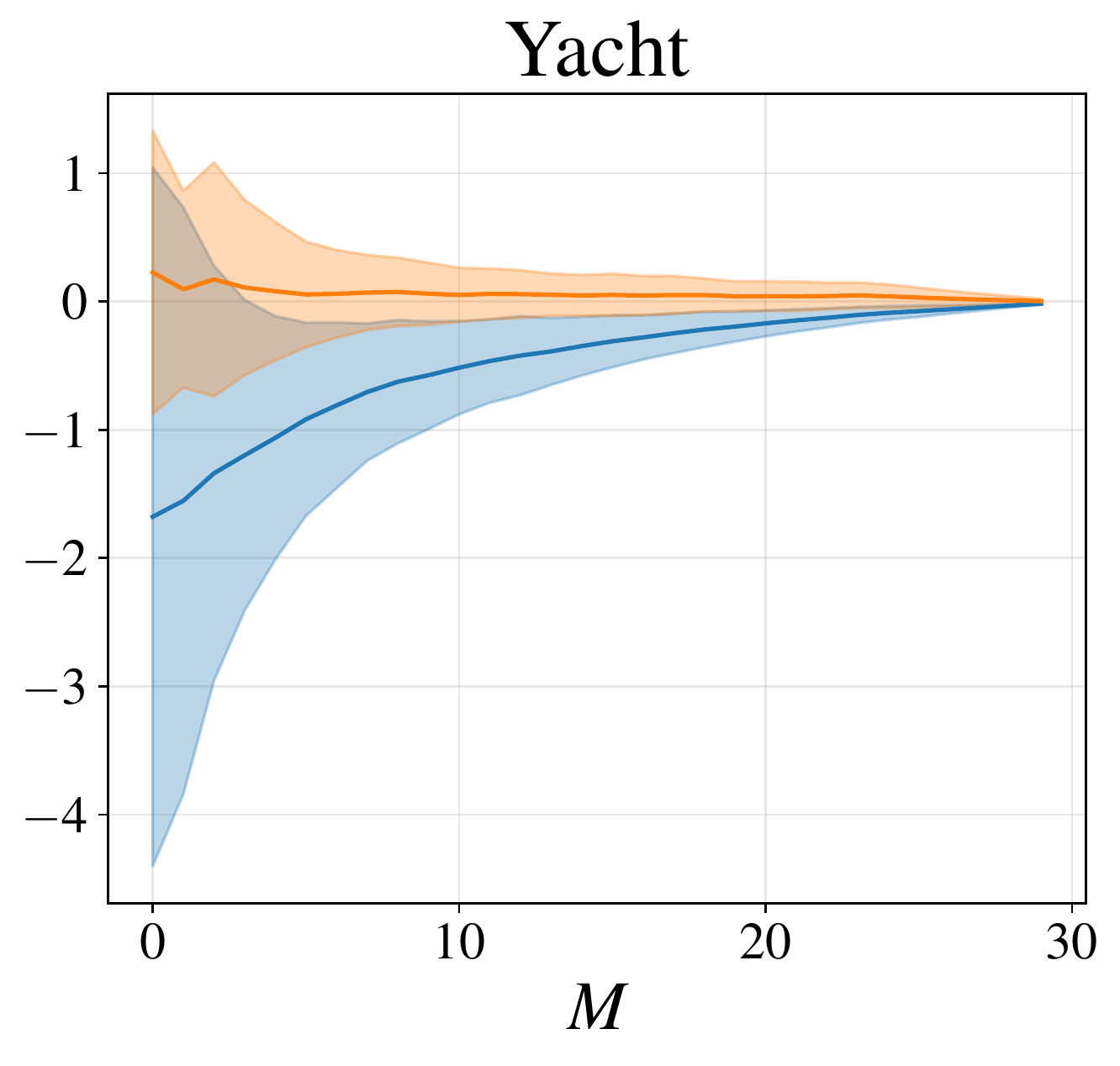}
\end{subfigure} \\
    \caption{Bias introduced by actively sampling points with DUNs, evaluated using $\tilde{R}$ (blue) and $\tilde{R}_{\text{LURE}}$ (orange).}
    \label{fig:app_res_alb}
\end{figure}

\begin{figure}[ht!] 
\centering
    \begin{subfigure}
        \centering
        \includegraphics[width=0.34\linewidth]{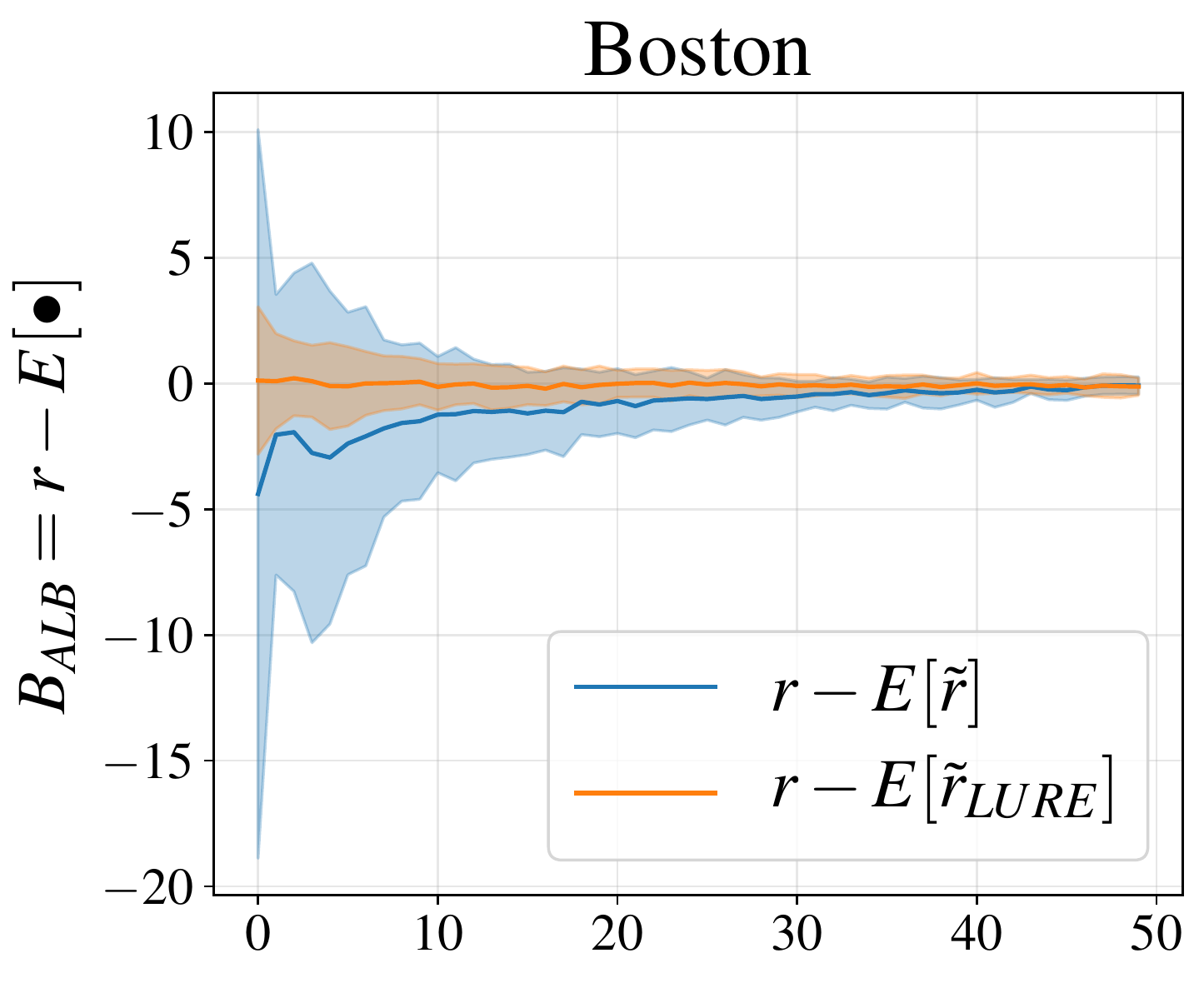}
    \end{subfigure}
    \begin{subfigure}
        \centering
        \includegraphics[width=0.31\linewidth]{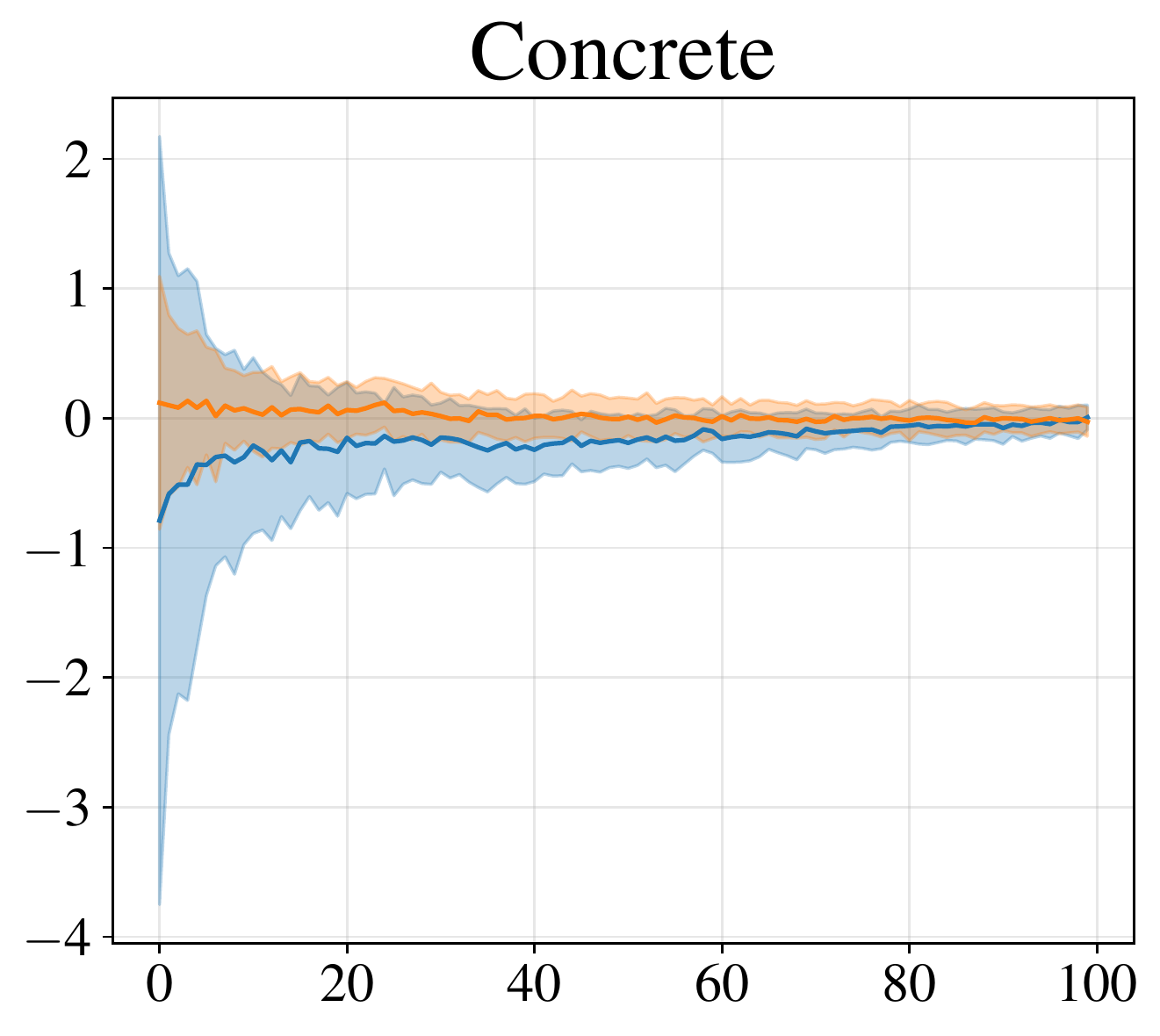}
    \end{subfigure} 
    \begin{subfigure}
        \centering
        \includegraphics[width=0.31\linewidth]{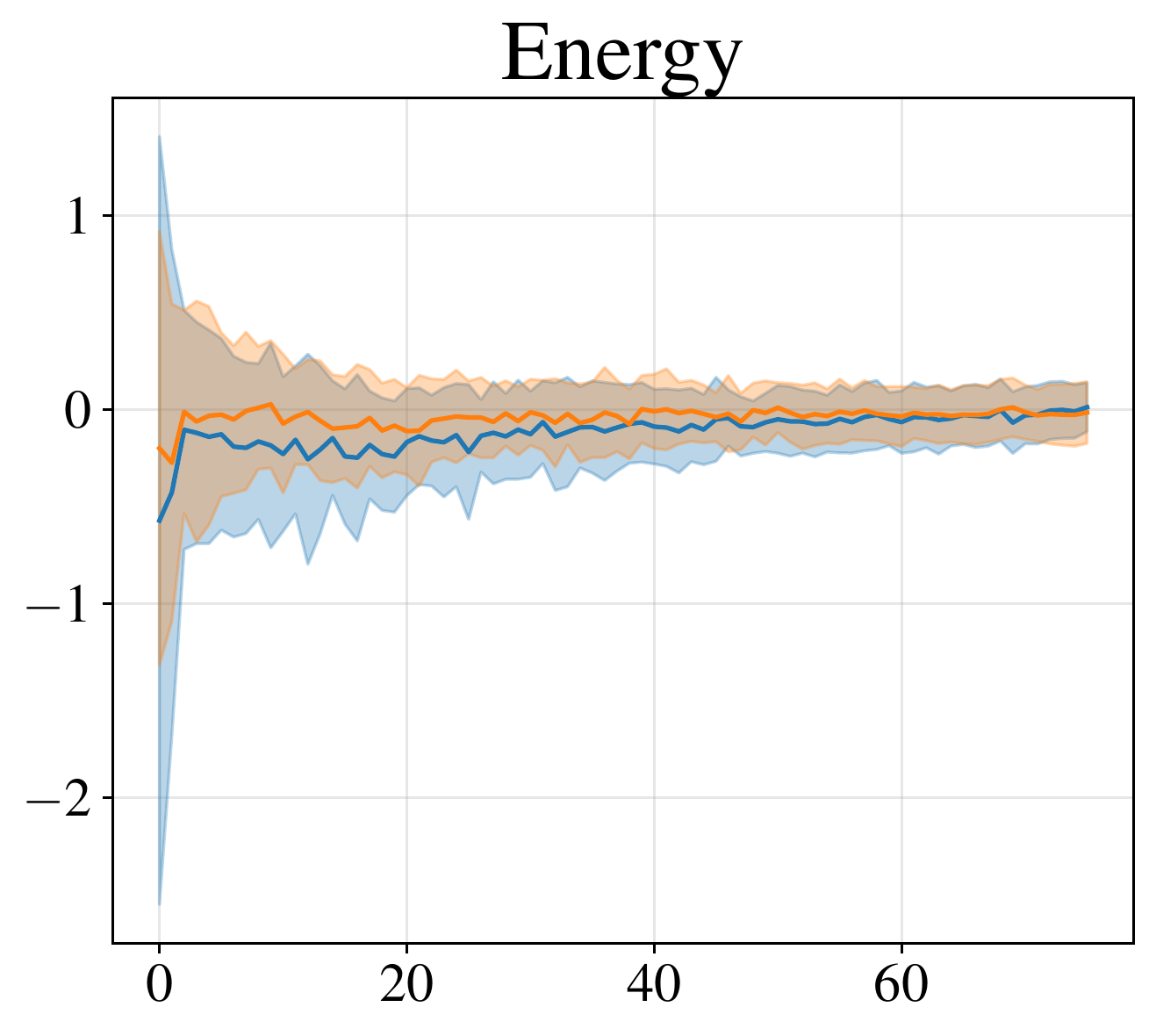} 
    \end{subfigure} \\
    \begin{subfigure}
        \centering
        \includegraphics[width=0.34\linewidth]{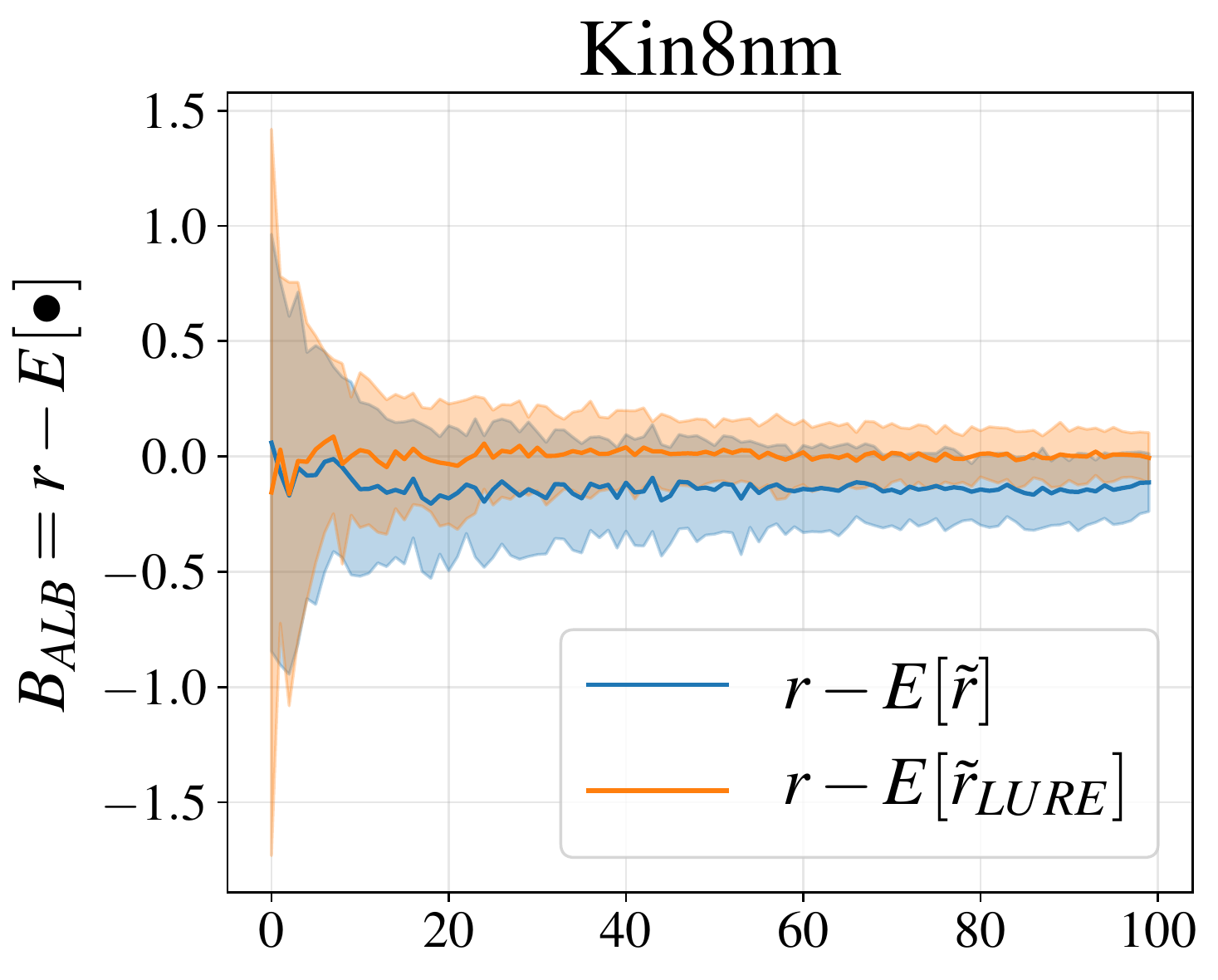}
    \end{subfigure} 
    \begin{subfigure}
        \centering
        \includegraphics[width=0.31\linewidth]{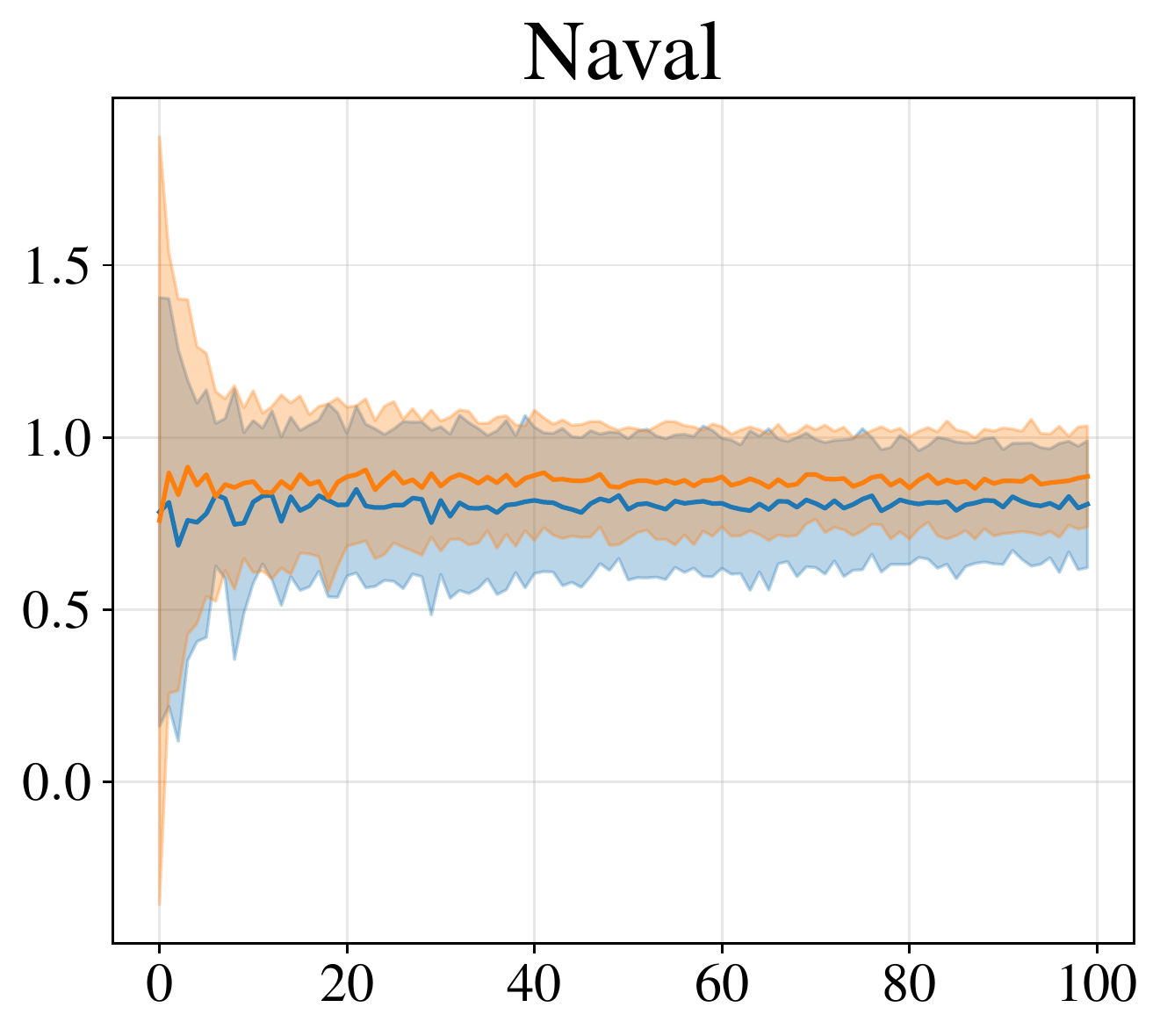}
    \end{subfigure} 
    \begin{subfigure}
        \centering
        \includegraphics[width=0.31\linewidth]{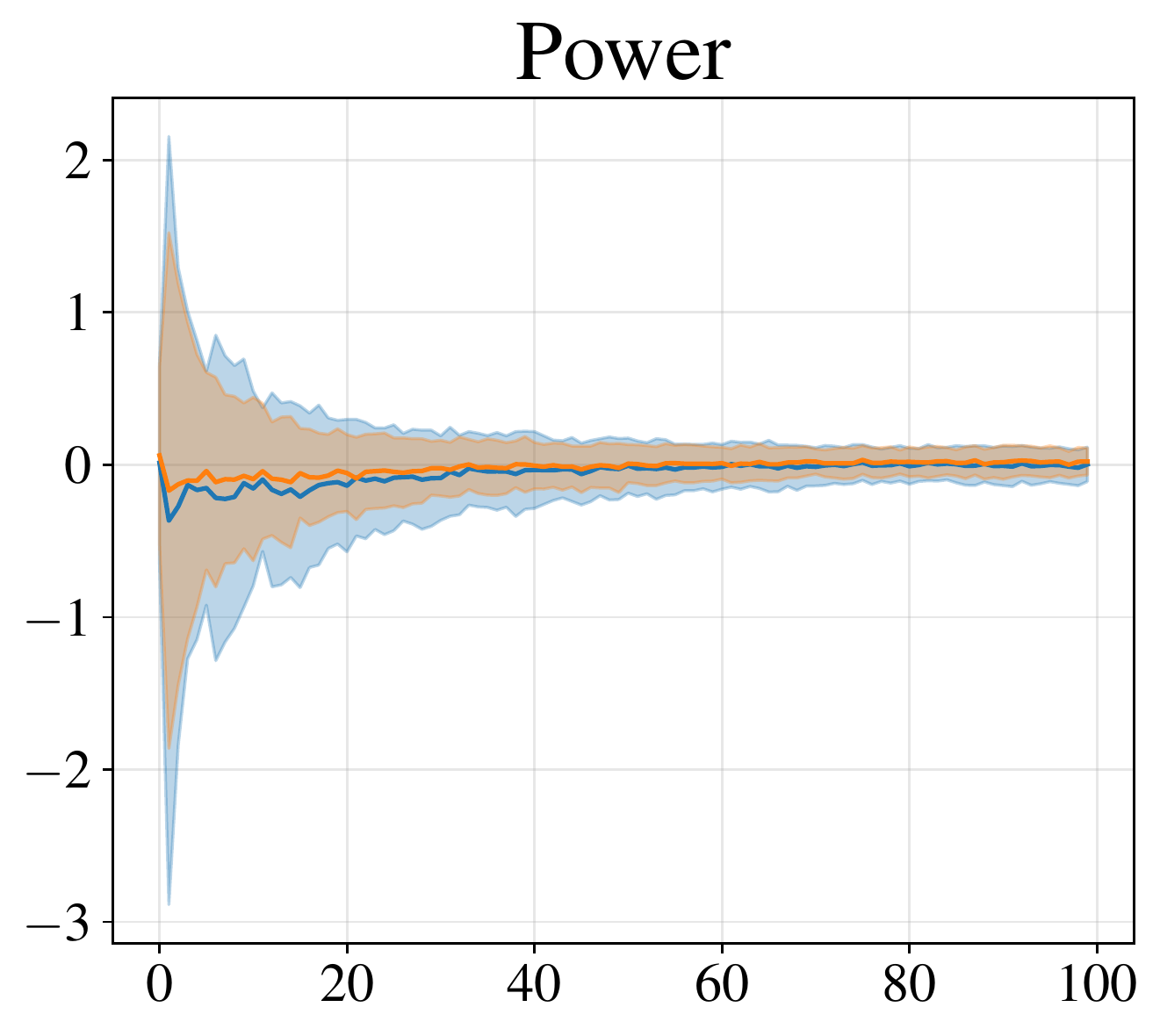}
    \end{subfigure} \\
    \begin{subfigure}
        \centering
        \includegraphics[width=0.34\linewidth]{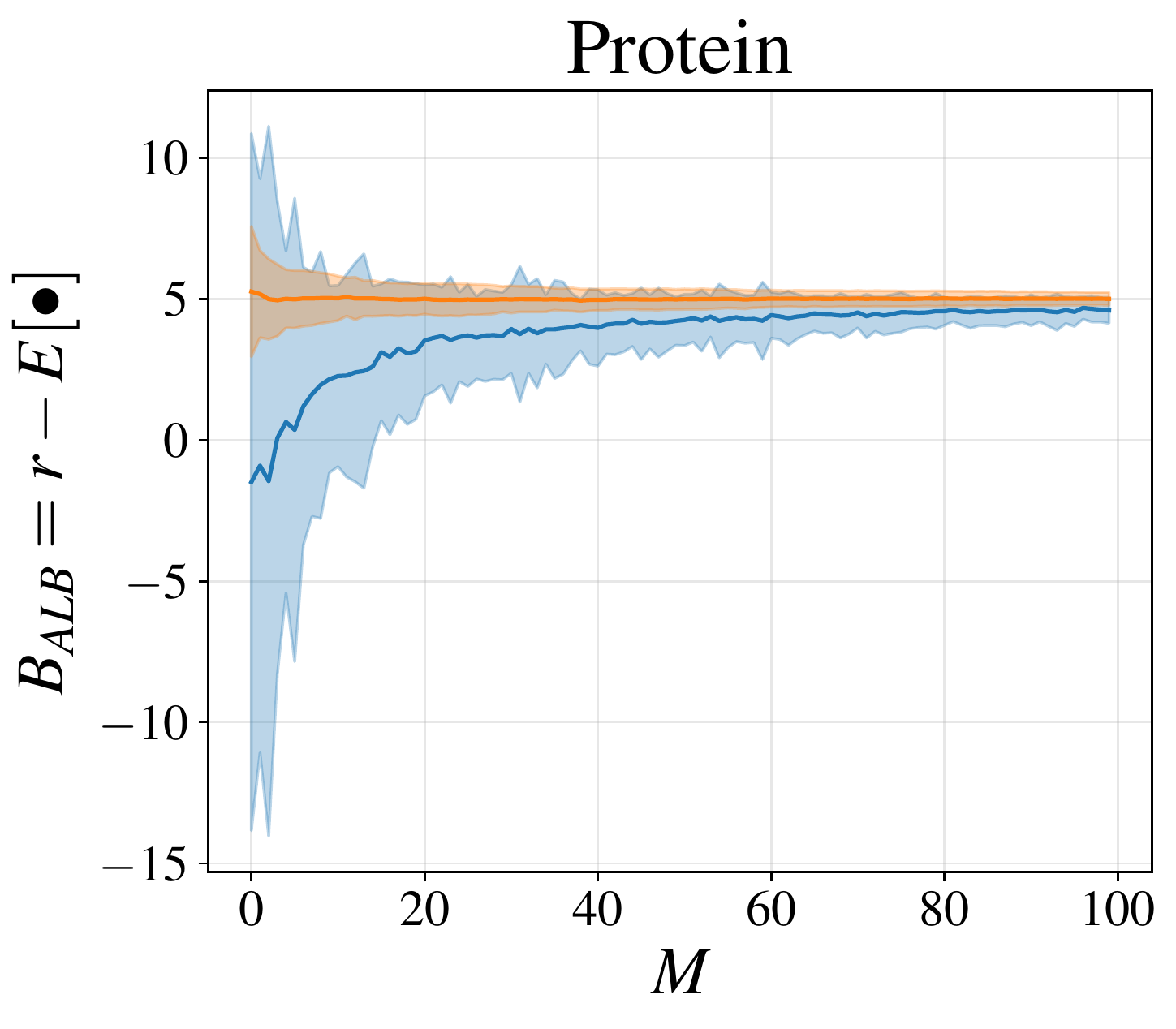}
    \end{subfigure} 
    \begin{subfigure}
        \centering
        \includegraphics[width=0.31\linewidth]{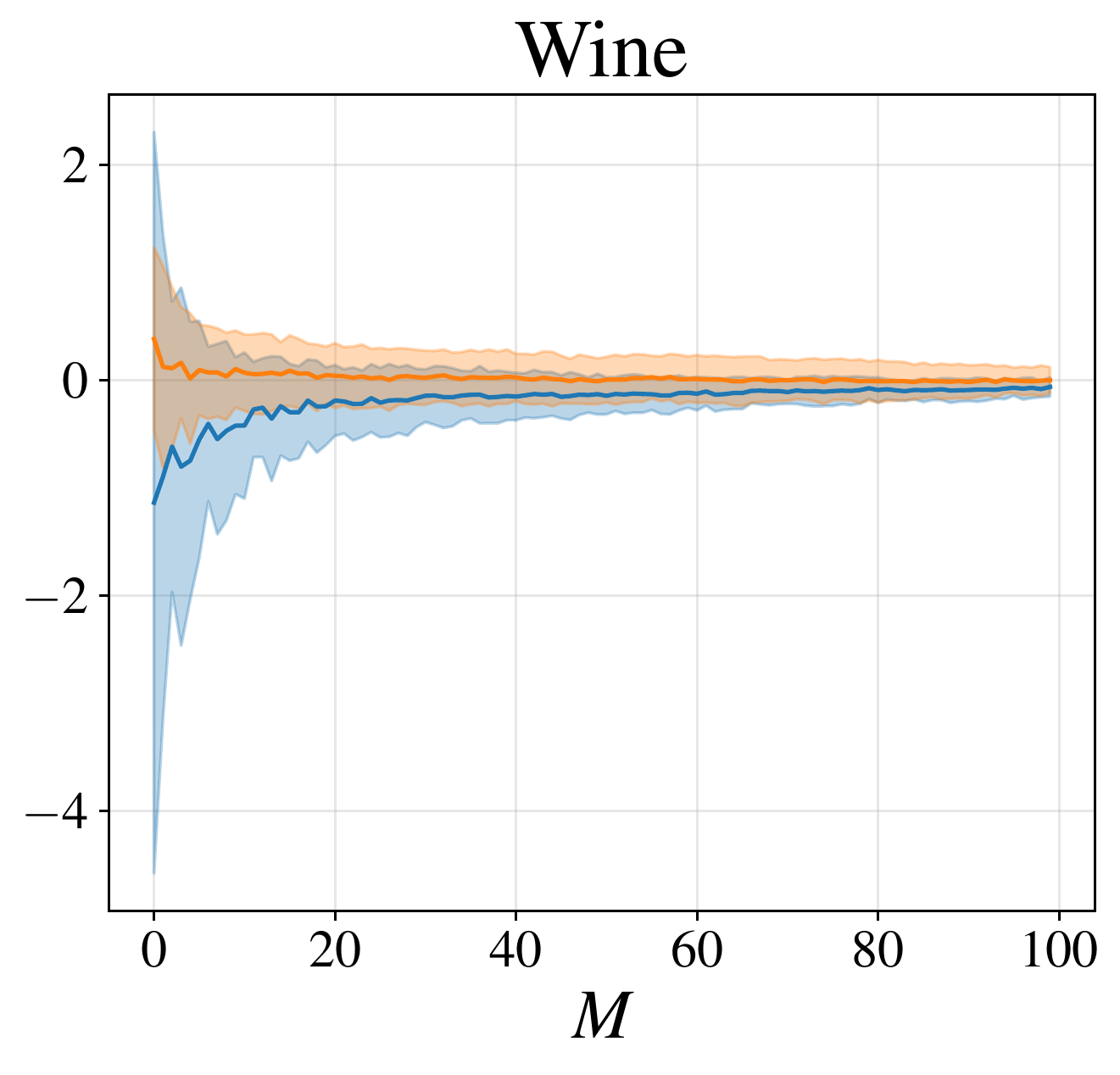}
    \end{subfigure} 
    \begin{subfigure}
        \centering
        \includegraphics[width=0.31\linewidth]{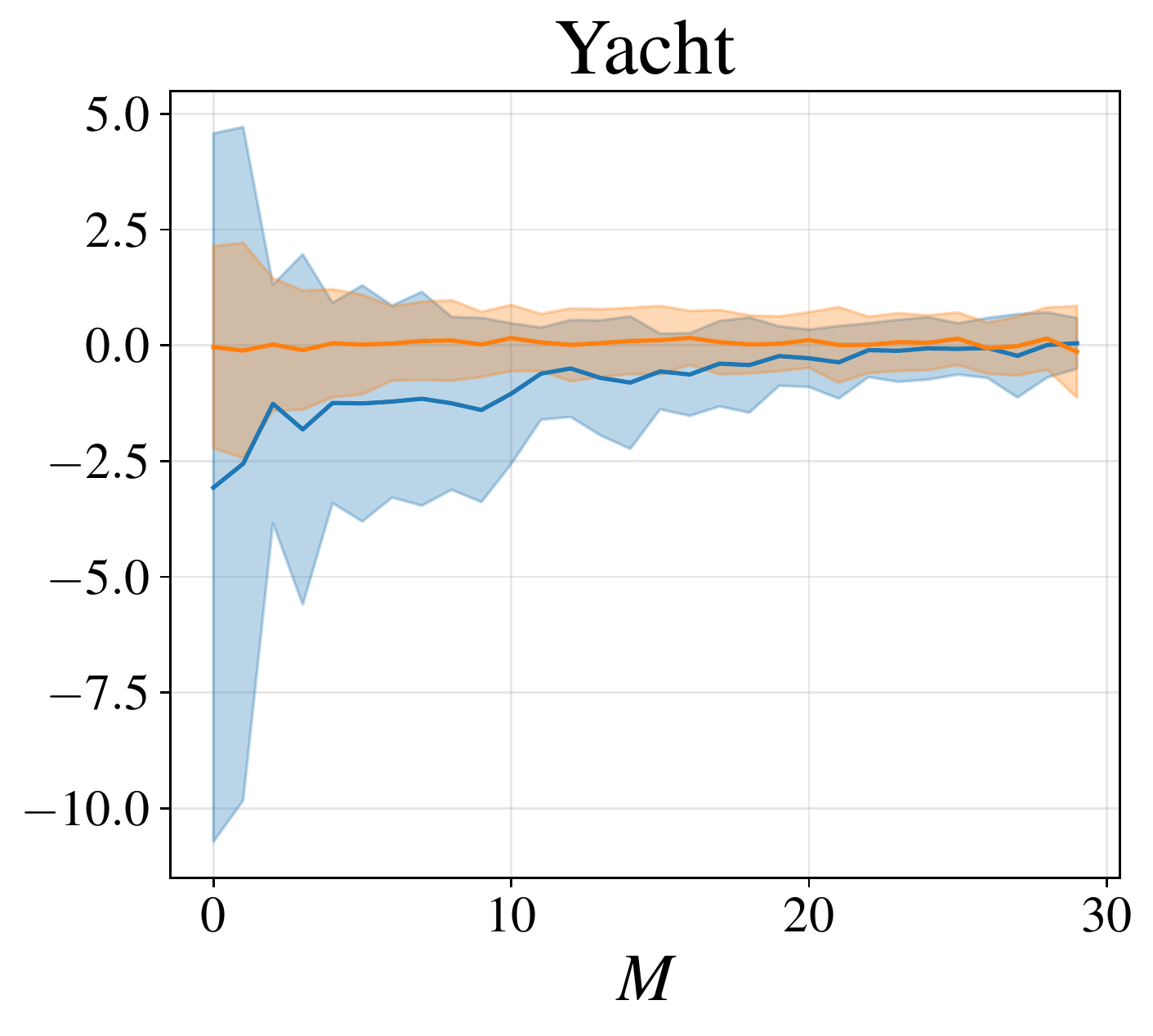}
\end{subfigure} \\
    \caption{Bias introduced by actively sampling points with MCDO, evaluated using $\tilde{R}$ (blue) and $\tilde{R}_{\text{LURE}}$ (orange).}
    \label{fig:app_res_alb_mcdo}
\end{figure}

\FloatBarrier
\clearpage
\subsection{Overfitting bias}\label{app:res_bias_ofb}

\begin{figure}[h] 
\centering
    \begin{subfigure}
        \centering
        \includegraphics[width=0.34\linewidth]{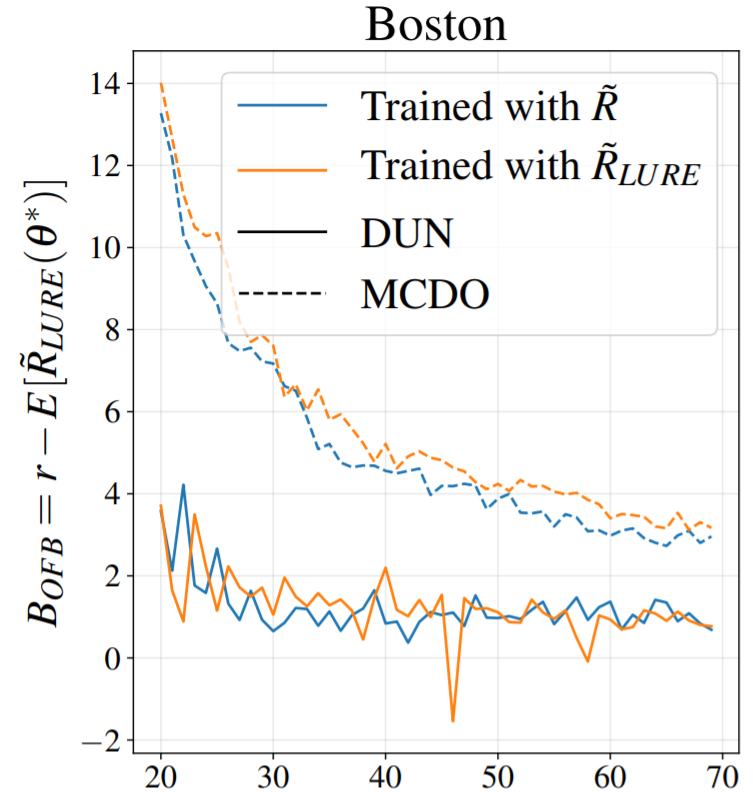}
    \end{subfigure}
    \begin{subfigure}
        \centering
        \includegraphics[width=0.3075\linewidth]{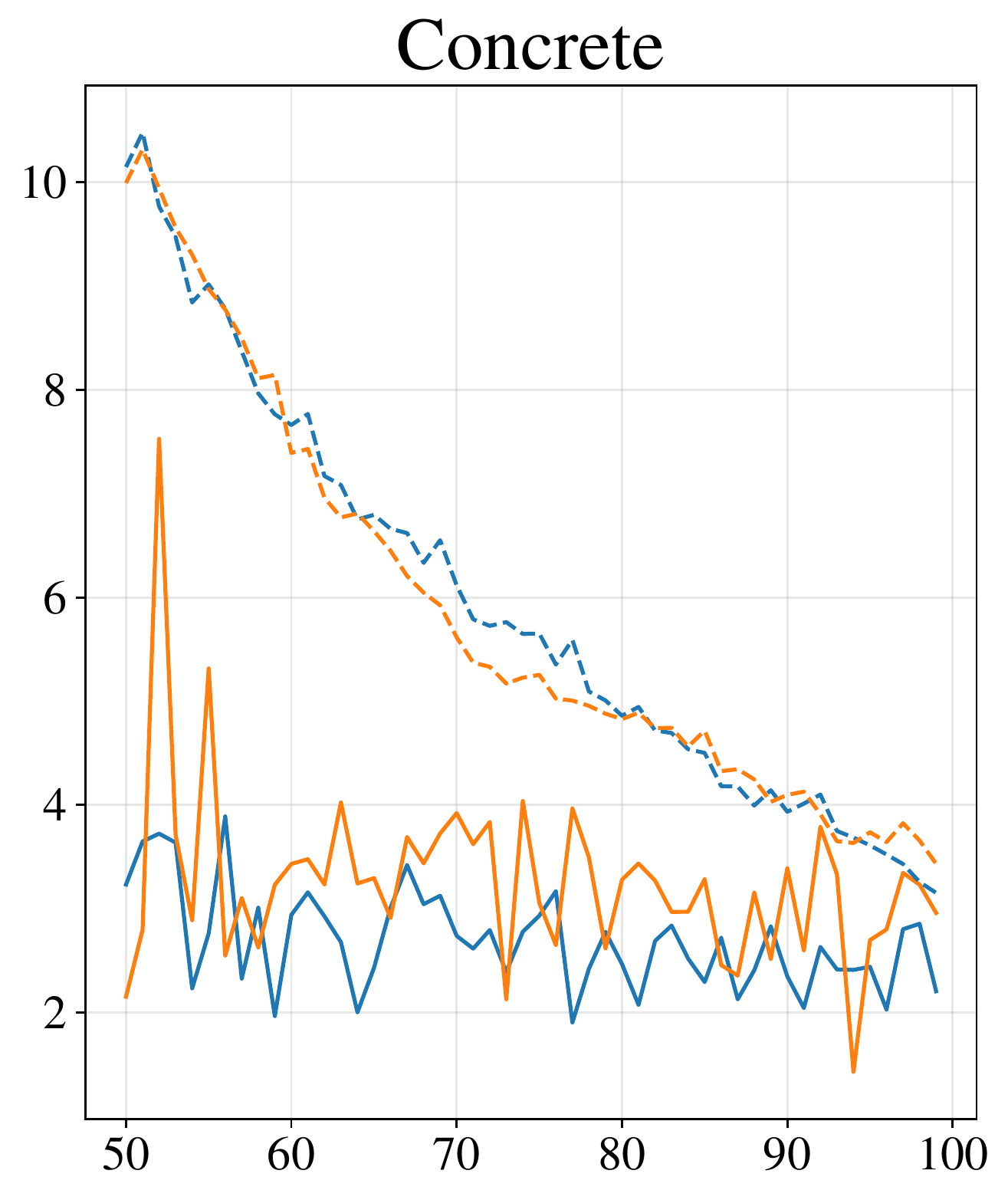}
    \end{subfigure} 
    \begin{subfigure}
        \centering
        \includegraphics[width=0.30\linewidth]{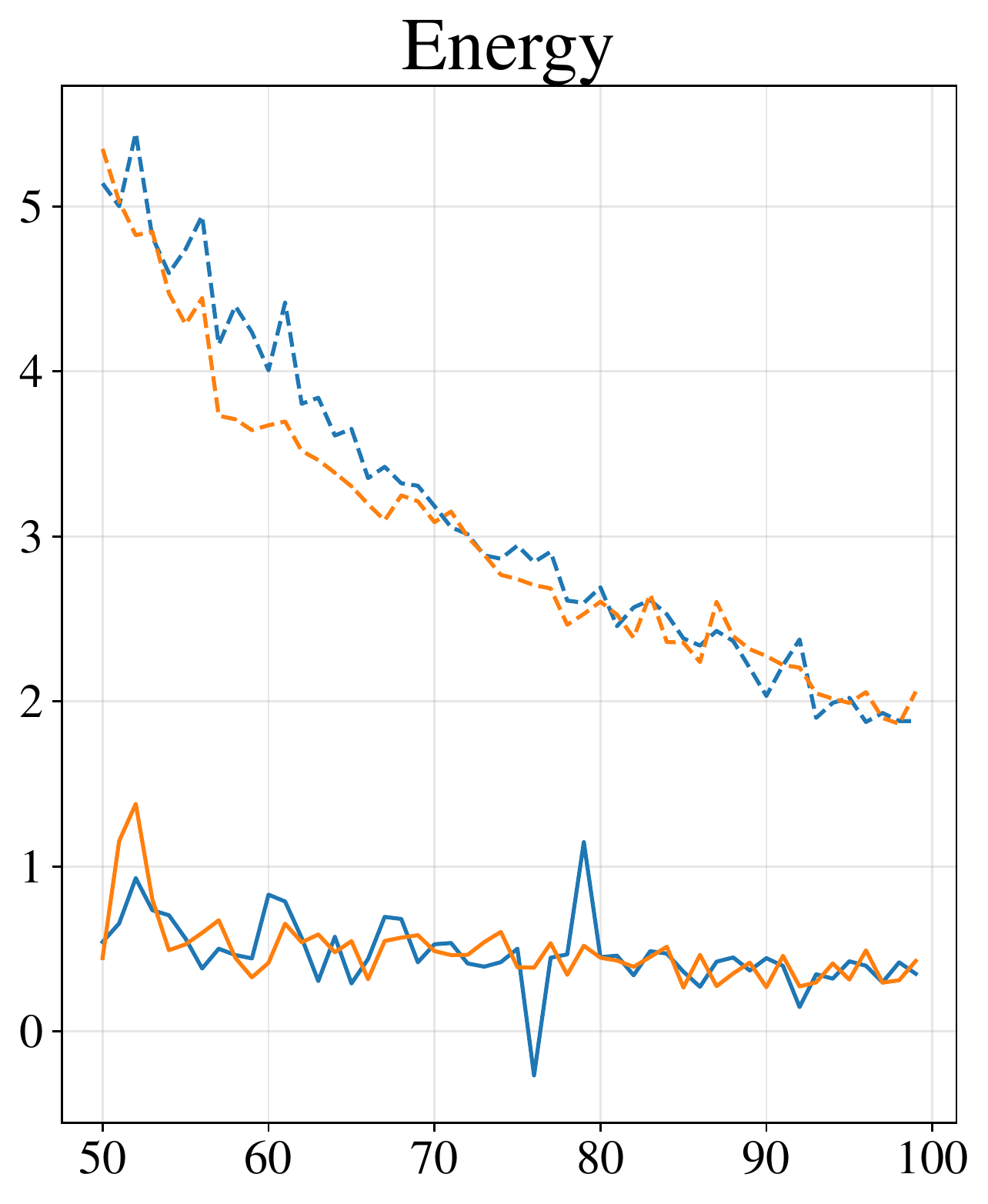} 
    \end{subfigure} \\
    \begin{subfigure}
        \centering
        \includegraphics[width=0.335\linewidth]{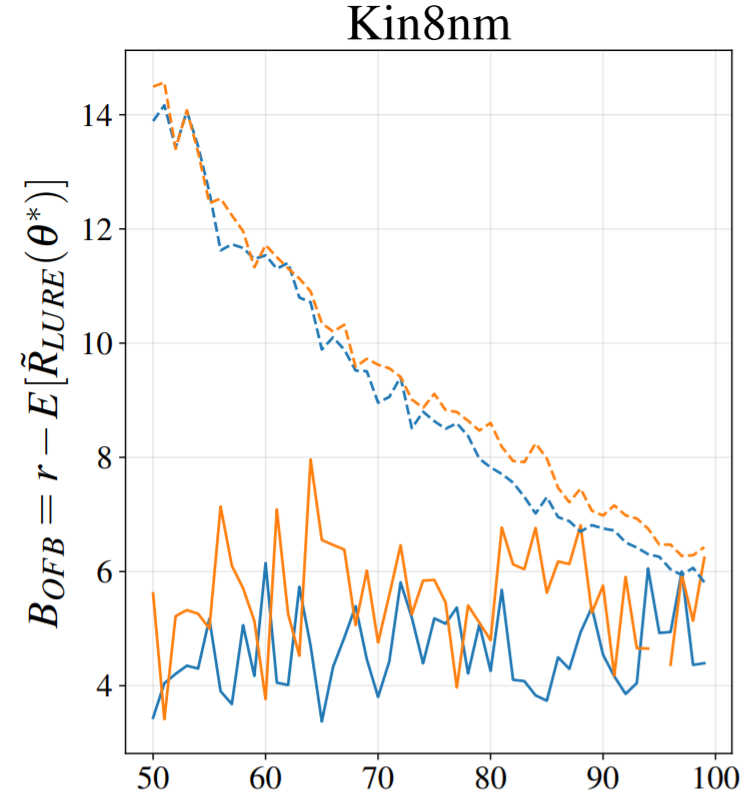}
    \end{subfigure} 
    \begin{subfigure}
        \centering
        \includegraphics[width=0.31\linewidth]{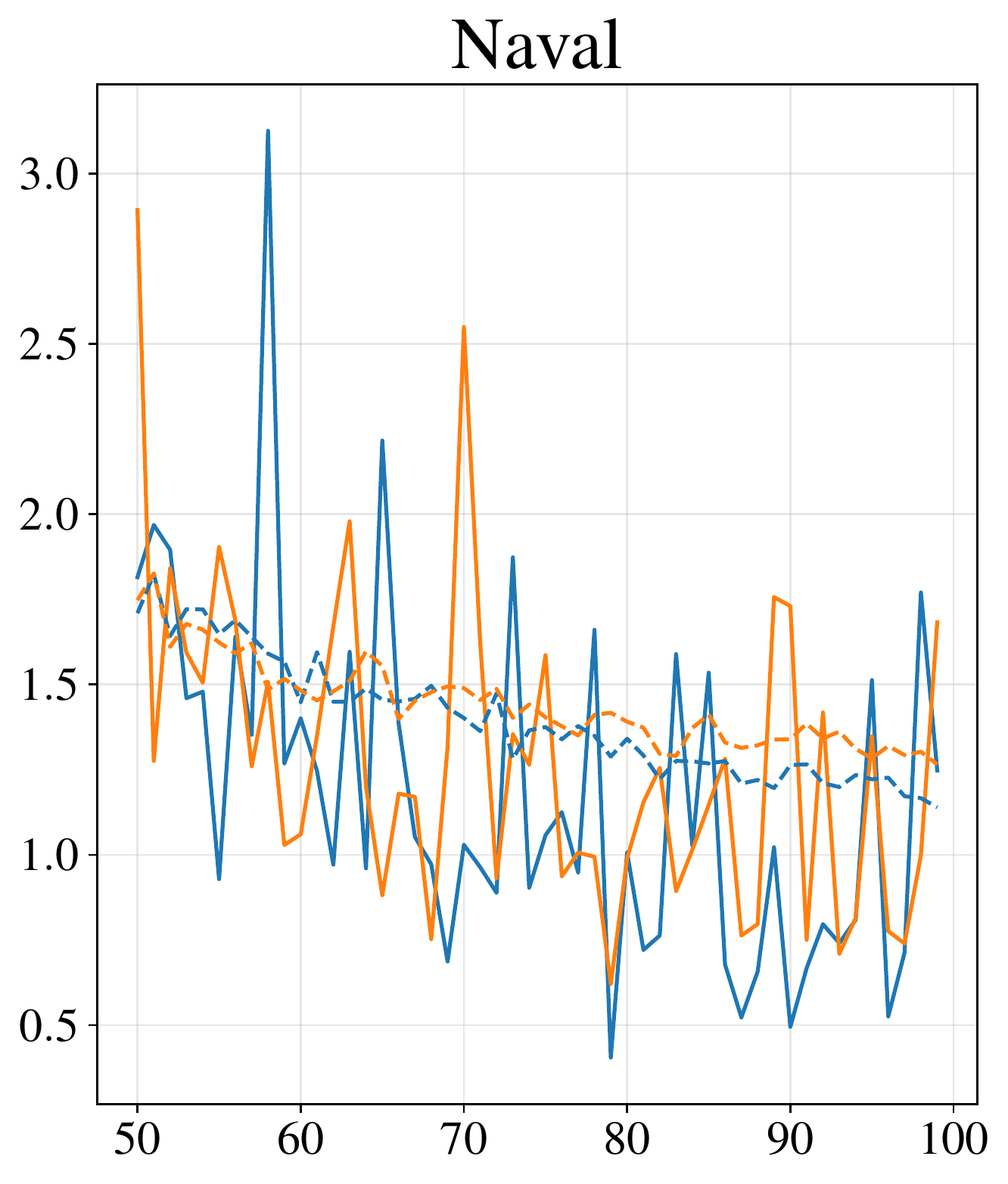}
    \end{subfigure} 
    \begin{subfigure}
        \centering
        \includegraphics[width=0.305\linewidth]{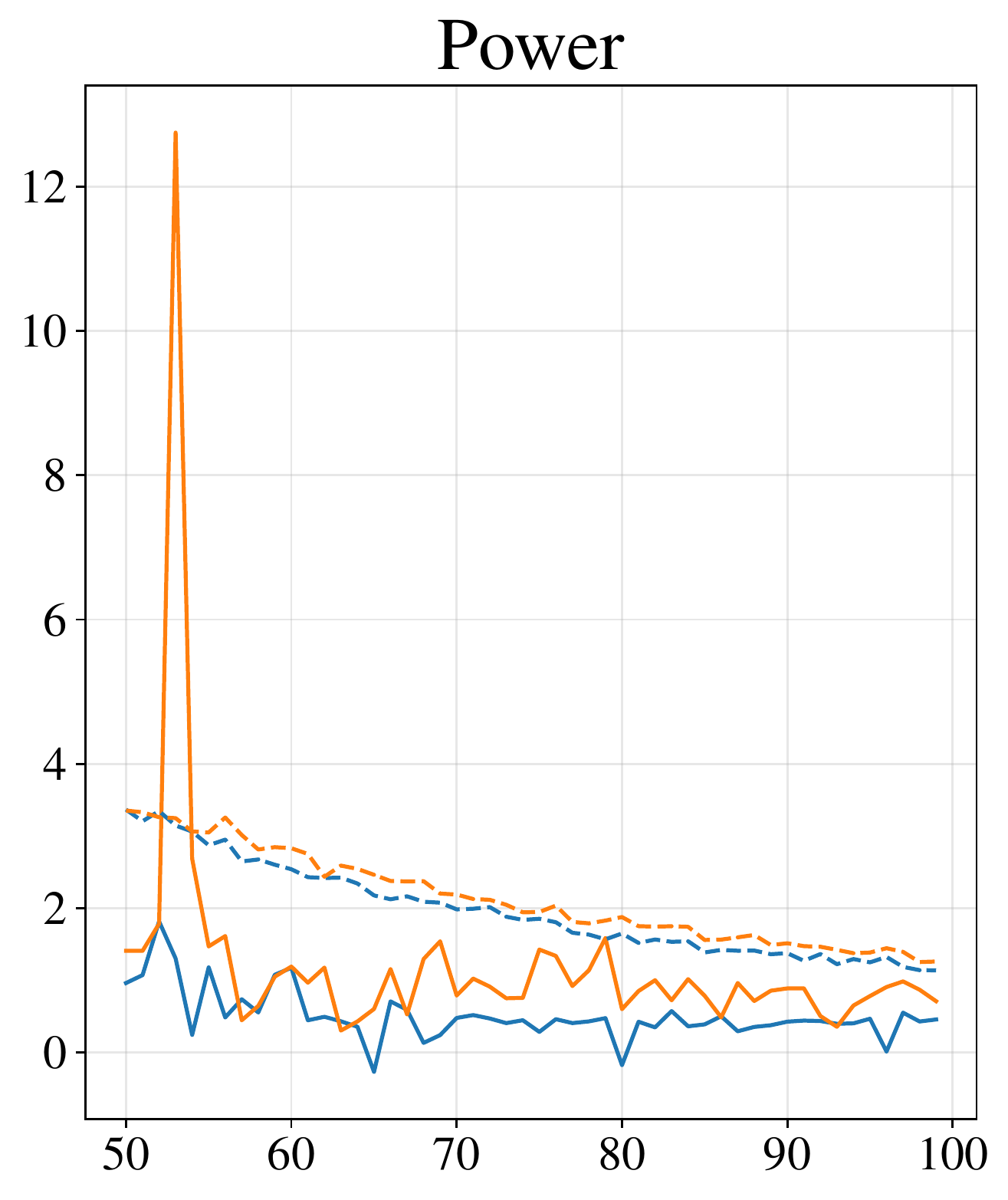}
    \end{subfigure} \\
    \begin{subfigure}
        \centering
        \includegraphics[width=0.343\linewidth]{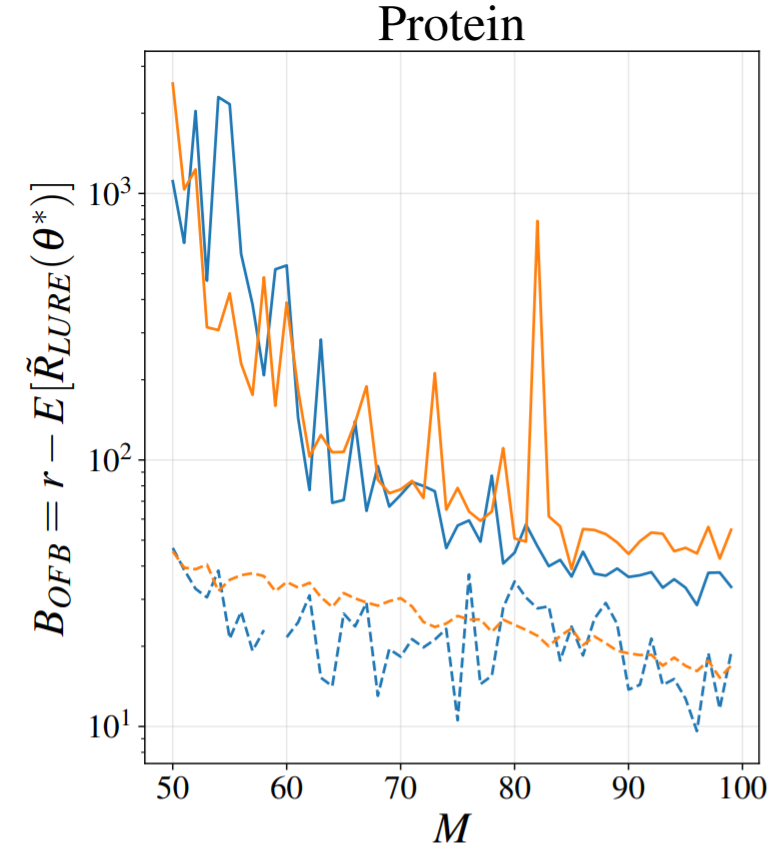}
    \end{subfigure} 
    \begin{subfigure}
        \centering
        \includegraphics[width=0.305\linewidth]{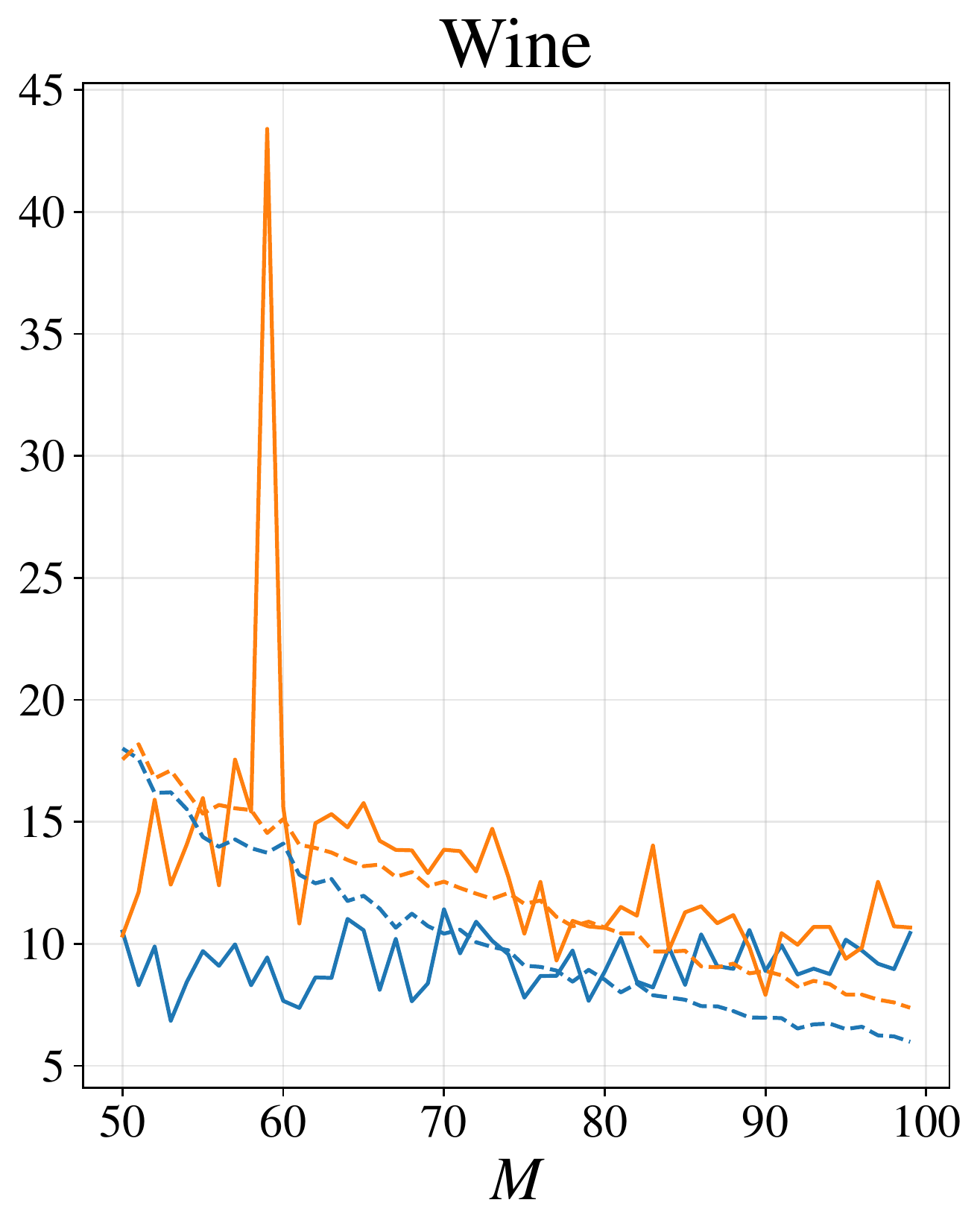}
    \end{subfigure} 
    \begin{subfigure}
        \centering
        \includegraphics[width=0.3125\linewidth]{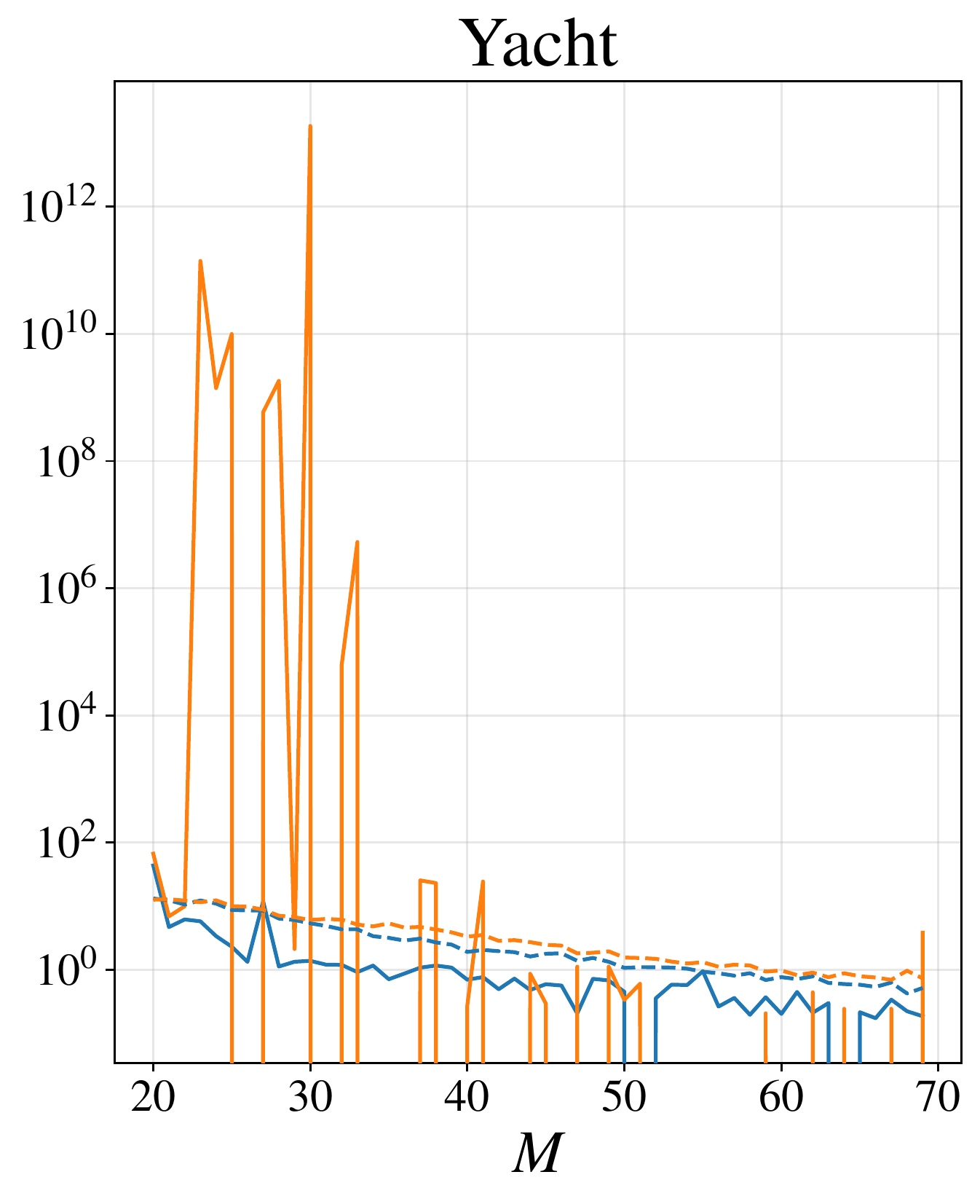}
\end{subfigure} \\
    \caption{Overfitting bias for DUNs (solid lines) and MCDO (dashed lines) trained with $\tilde{R}$ (blue) and $\tilde{R}_{\text{LURE}}$ (orange). NLL for Protein and Yacht is displayed in log scale.}
    \label{fig:app_res_ofb_DUNvMCDO}
\end{figure}

\Cref{fig:app_res_ofb_DUNvMCDO} compares the magnitude of overfitting bias for DUNs and BNNs trained with MCDO, as in \cref{fig:res_ofb_DUNvMCDO}. \Cref{fig:app_res_ofb_DUNvMCDO} shows additionally the overfitting bias when training with $\tilde{R}$.

\FloatBarrier
\newpage
\subsection{Training with unbiased risk estimators}\label{app:res_bias_fig3}

\begin{figure}[h!] 
\centering
    \begin{subfigure}
        \centering
        \includegraphics[width=0.33\linewidth]{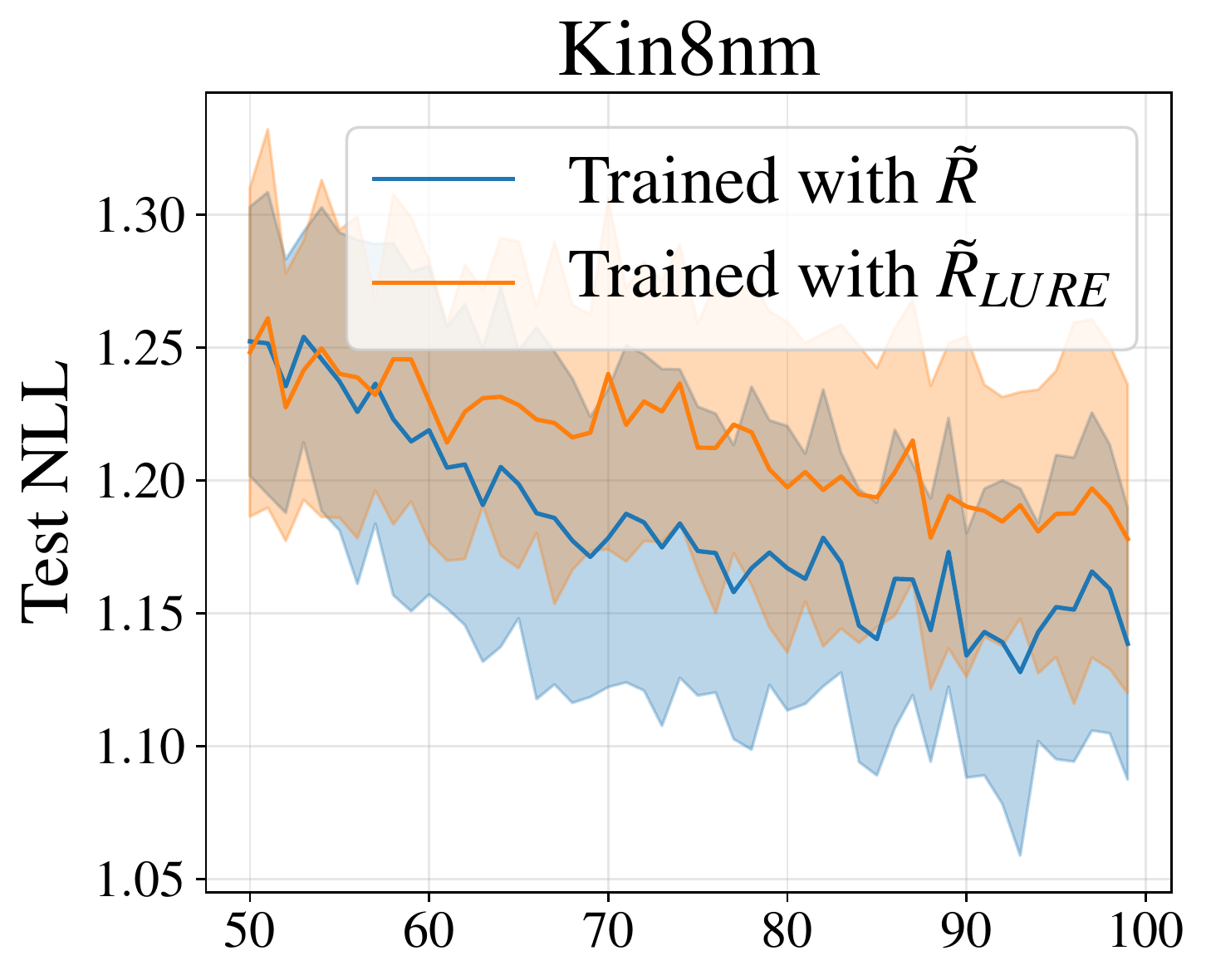}
    \end{subfigure} 
    \begin{subfigure}
        \centering
        \includegraphics[width=0.315\linewidth]{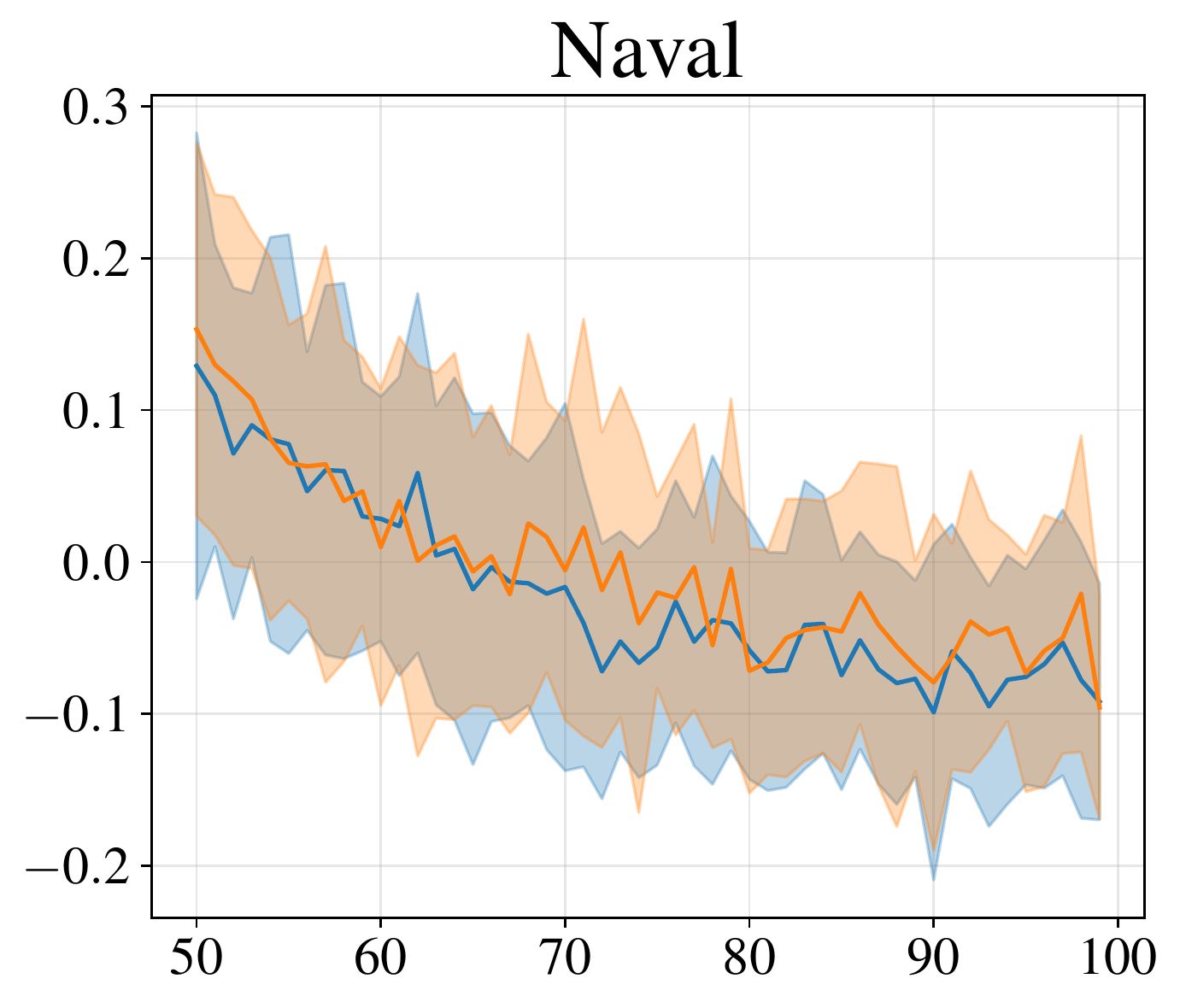}
    \end{subfigure} 
    \begin{subfigure}
        \centering
        \includegraphics[width=0.315\linewidth]{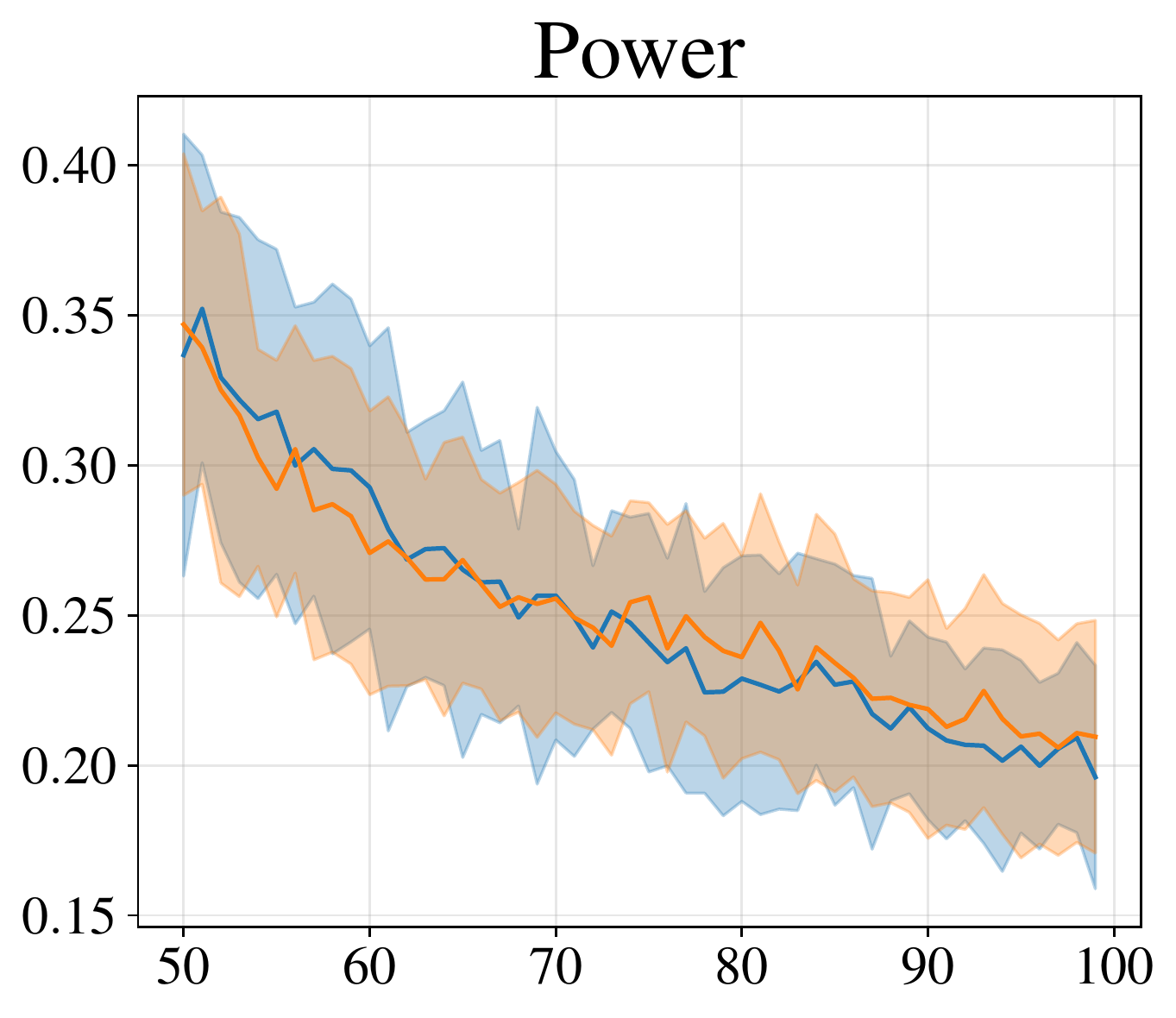}
    \end{subfigure} \\
    \begin{subfigure}
        \centering
        \includegraphics[width=0.33\linewidth]{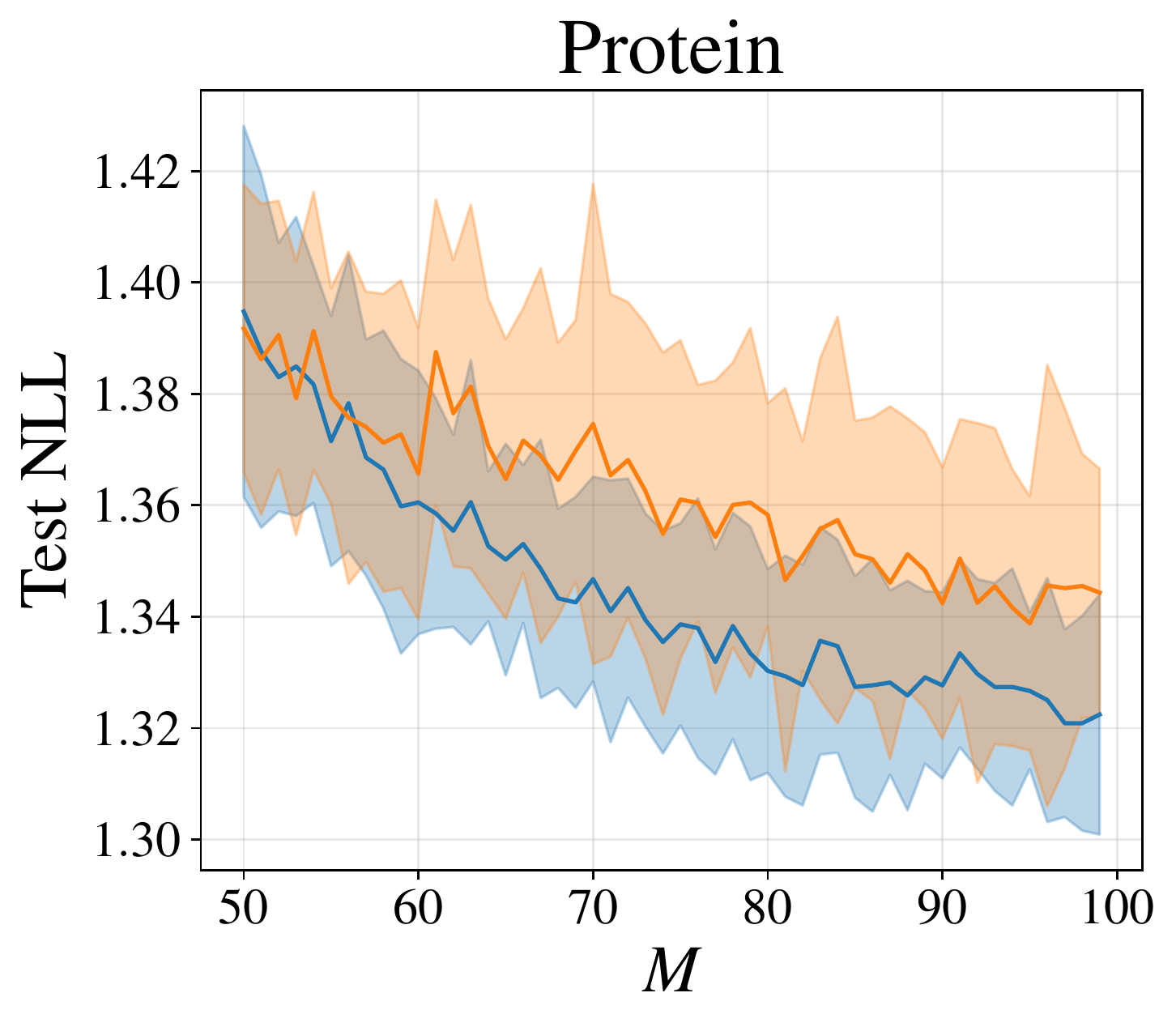}
    \end{subfigure} 
    \begin{subfigure}
        \centering
        \includegraphics[width=0.315\linewidth]{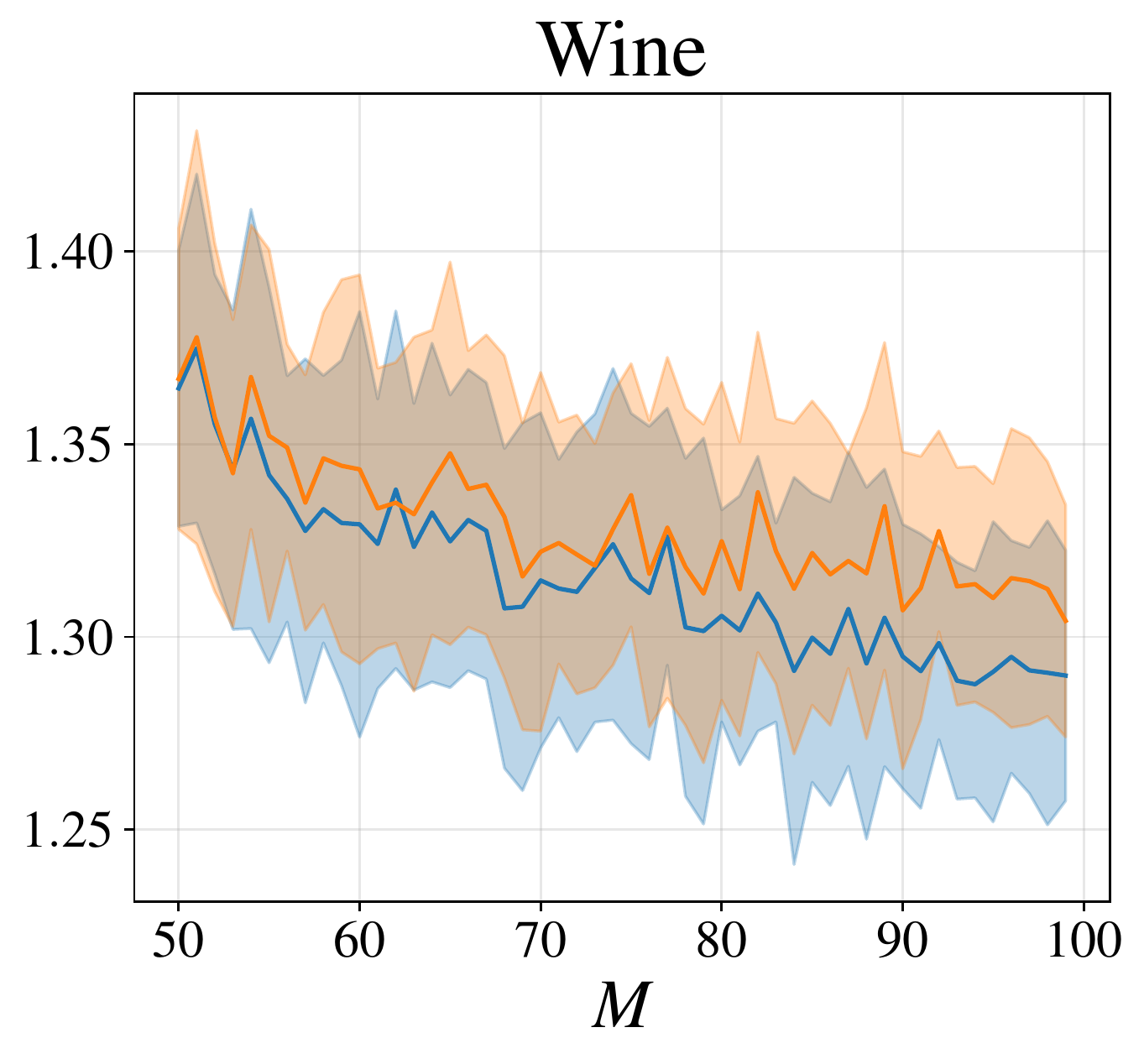}
    \end{subfigure} 
    \begin{subfigure}
        \centering
        \includegraphics[width=0.315\linewidth]{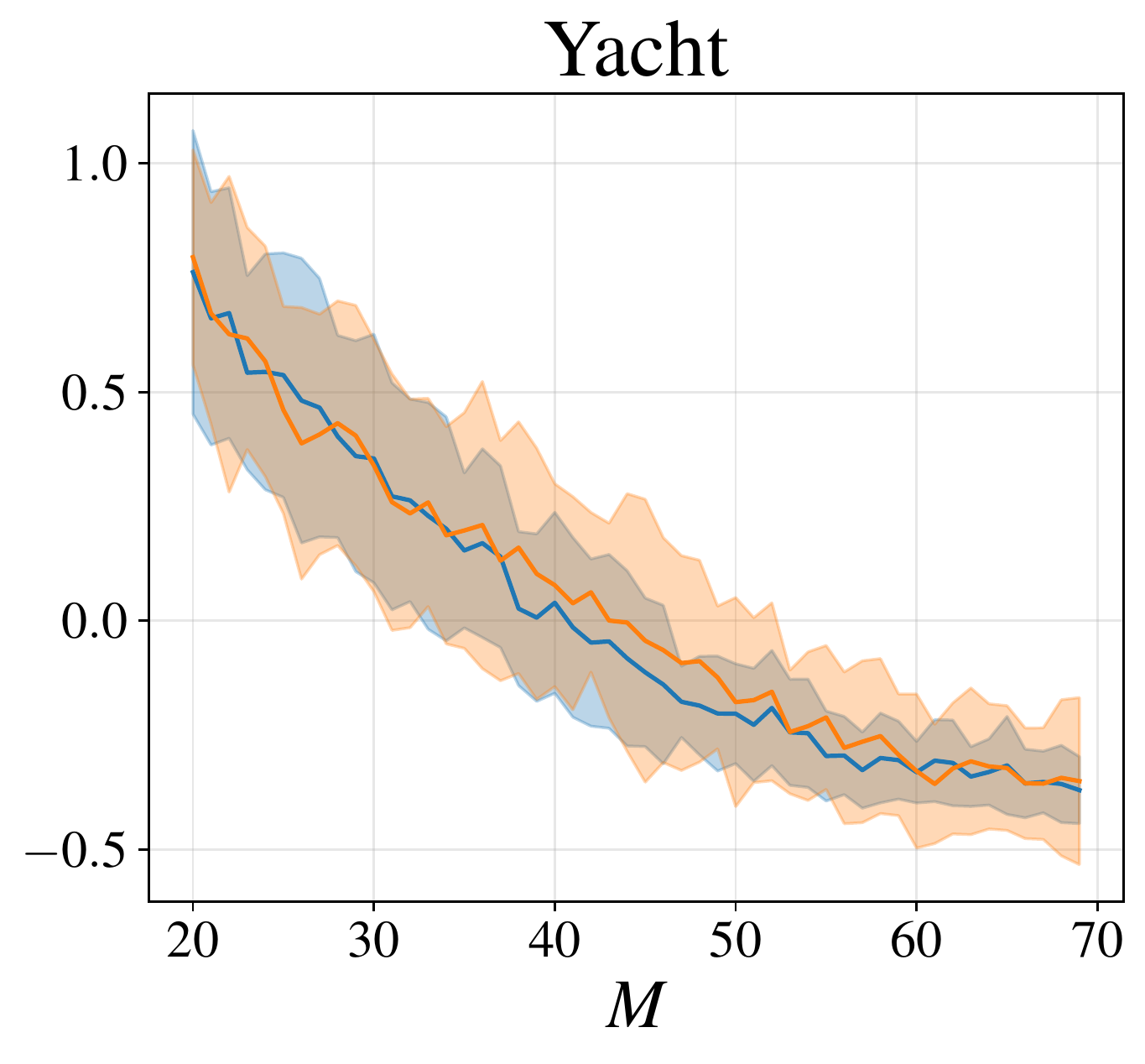}
    \end{subfigure} \\
    \caption{Test NLL for DUNs trained with $\tilde{R}$ (blue) and $\tilde{R}_{\text{LURE}}$ (orange). A larger value of the orange line indicates that training with $\tilde{R}_{\text{LURE}}$ harms performance, despite removing active learning bias.}
    \label{fig:app_res_fig3_uniform_nll}
\end{figure}

\Cref{fig:app_res_fig3_uniform_nll} presents the results of training DUNs with $\tilde{R}$ and $\tilde{R}_{\text{LURE}}$, as in \cref{fig:res_fig3_uniform_nll}, for the remaining datasets.

\FloatBarrier
\newpage
\subsection{Temperature of the proposal distribution}\label{app:temp}

\begin{figure}[h] 
\centering
    \begin{subfigure}
        \centering
        \includegraphics[width=0.33\linewidth]{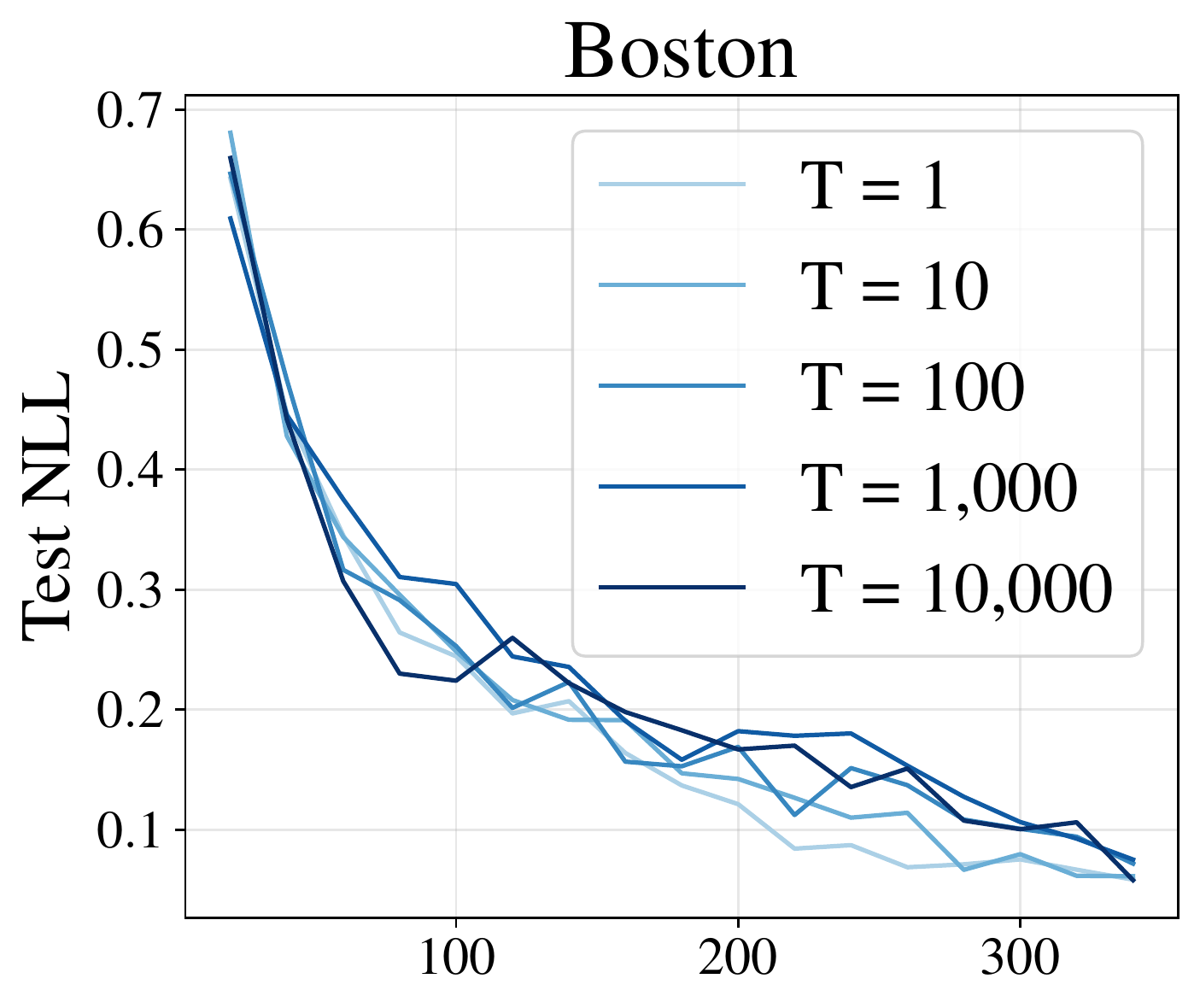}
    \end{subfigure}
    \begin{subfigure}
        \centering
        \includegraphics[width=0.31\linewidth]{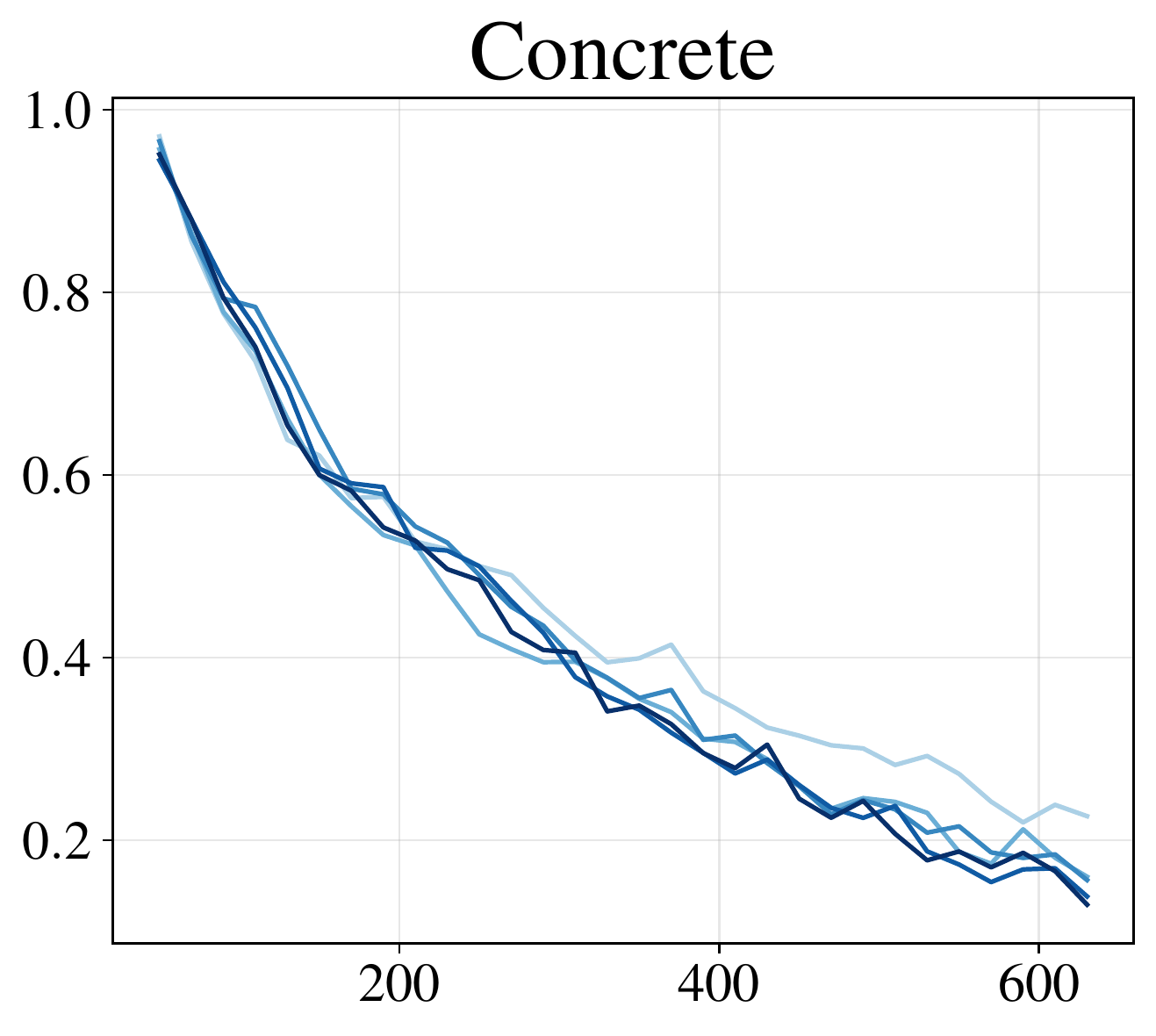}
    \end{subfigure} 
    \begin{subfigure}
        \centering
        \includegraphics[width=0.325\linewidth]{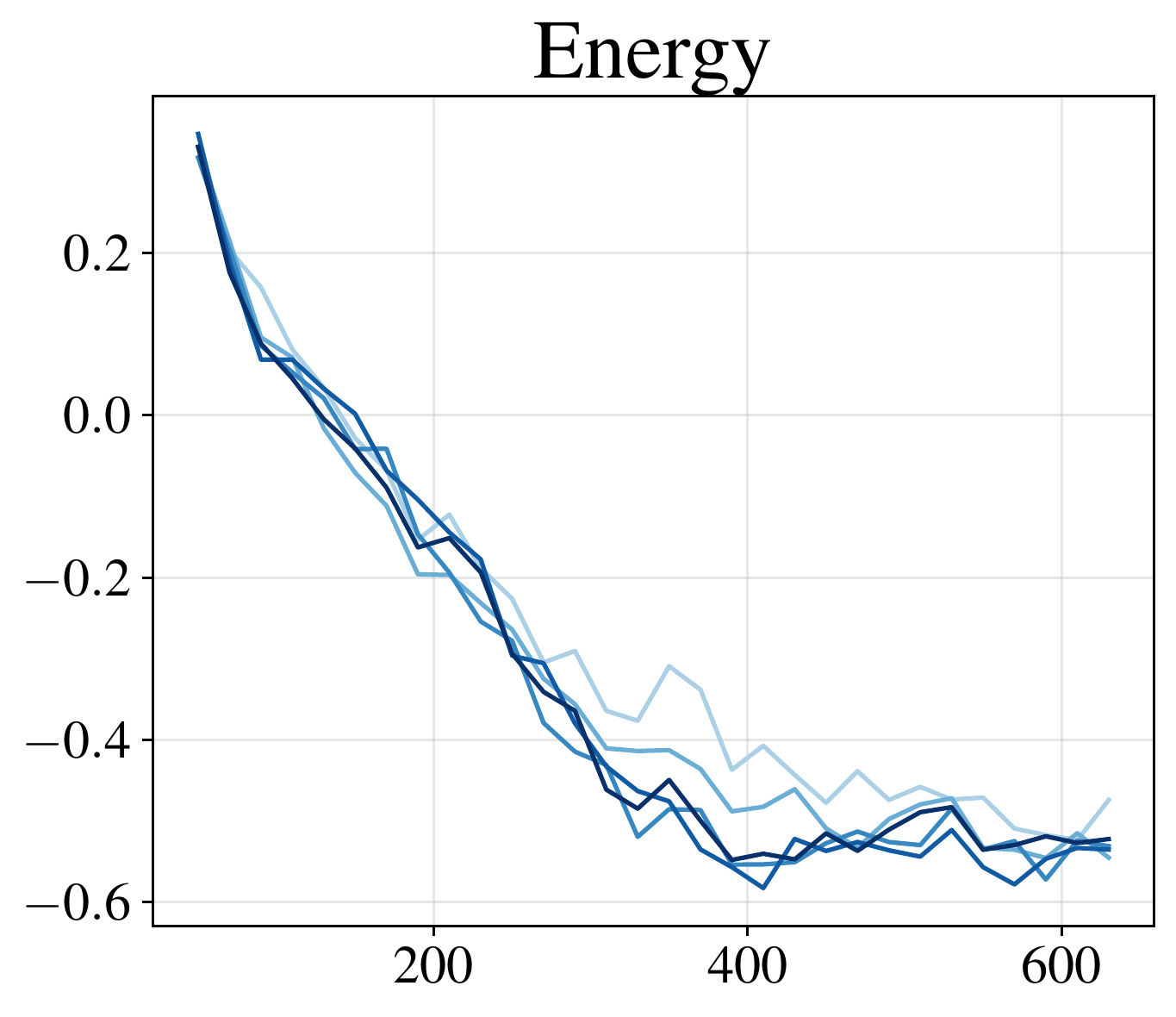} 
    \end{subfigure} \\
    \begin{subfigure}
        \centering
        \includegraphics[width=0.33\linewidth]{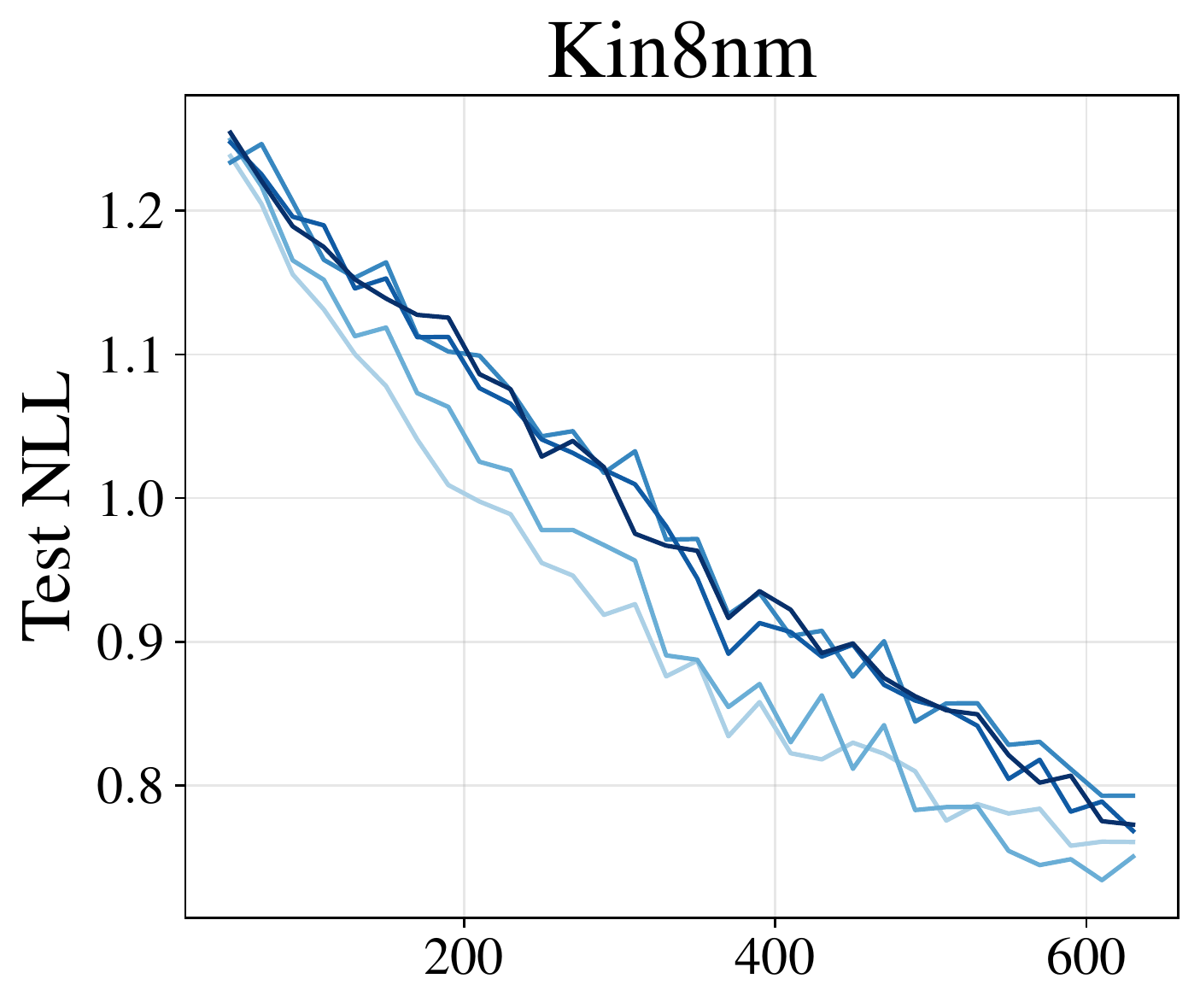}
    \end{subfigure} 
    \begin{subfigure}
        \centering
        \includegraphics[width=0.32\linewidth]{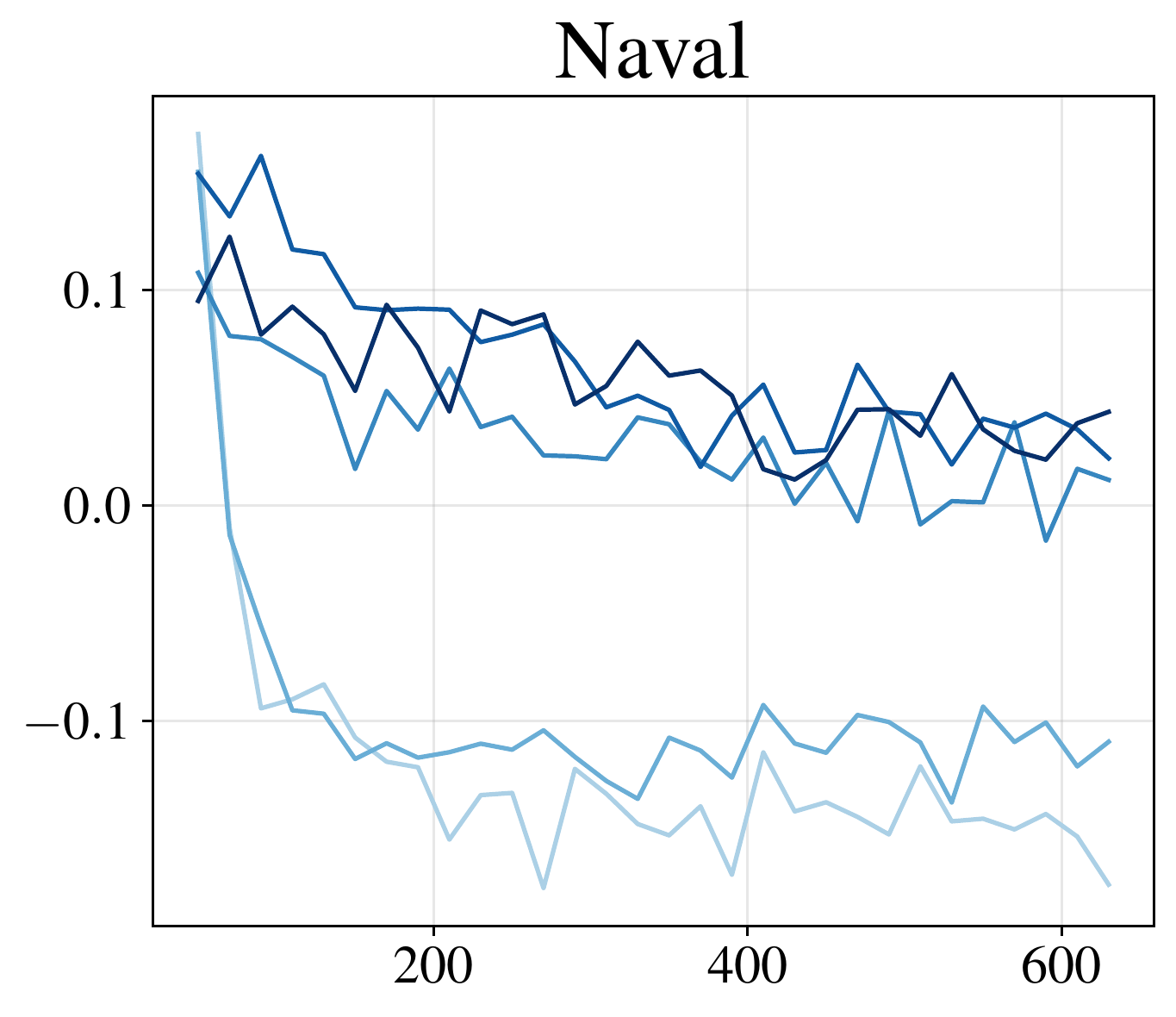}
    \end{subfigure} 
    \begin{subfigure}
        \centering
        \includegraphics[width=0.315\linewidth]{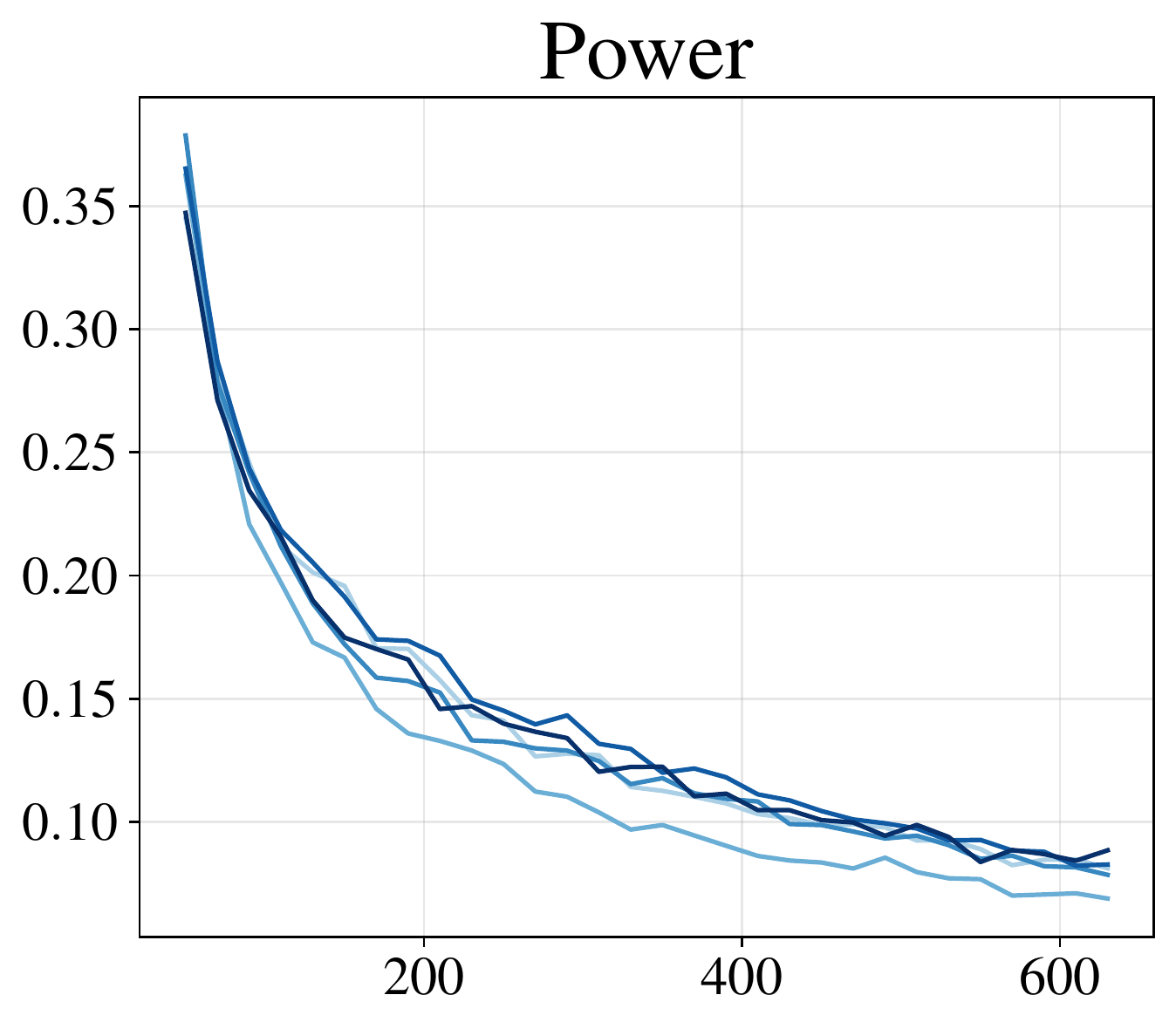}
    \end{subfigure} \\
    \begin{subfigure}
        \centering
        \includegraphics[width=0.33\linewidth]{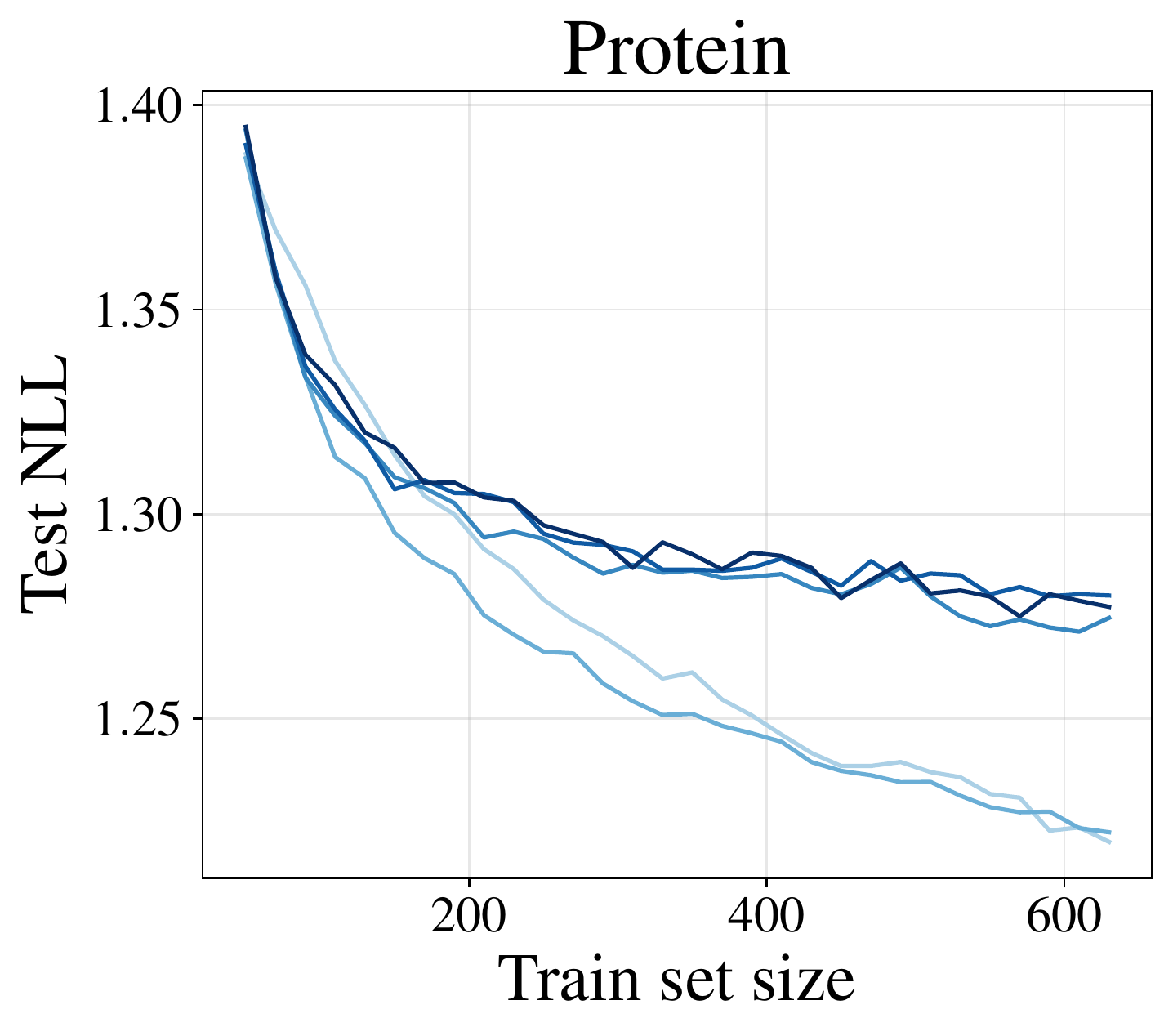}
    \end{subfigure} 
    \begin{subfigure}
        \centering
        \includegraphics[width=0.31\linewidth]{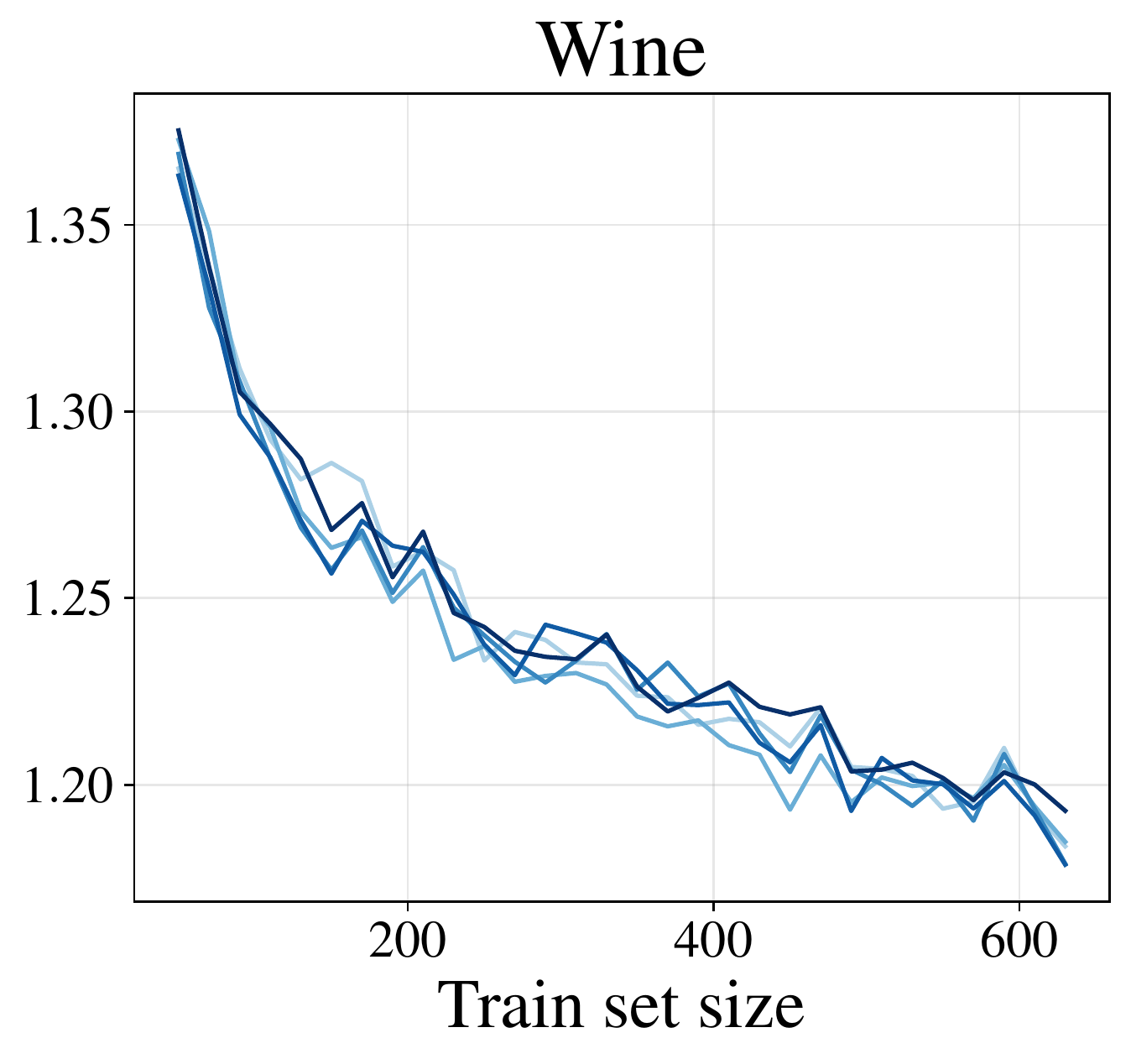}
    \end{subfigure} 
    \begin{subfigure}
        \centering
        \includegraphics[width=0.32\linewidth]{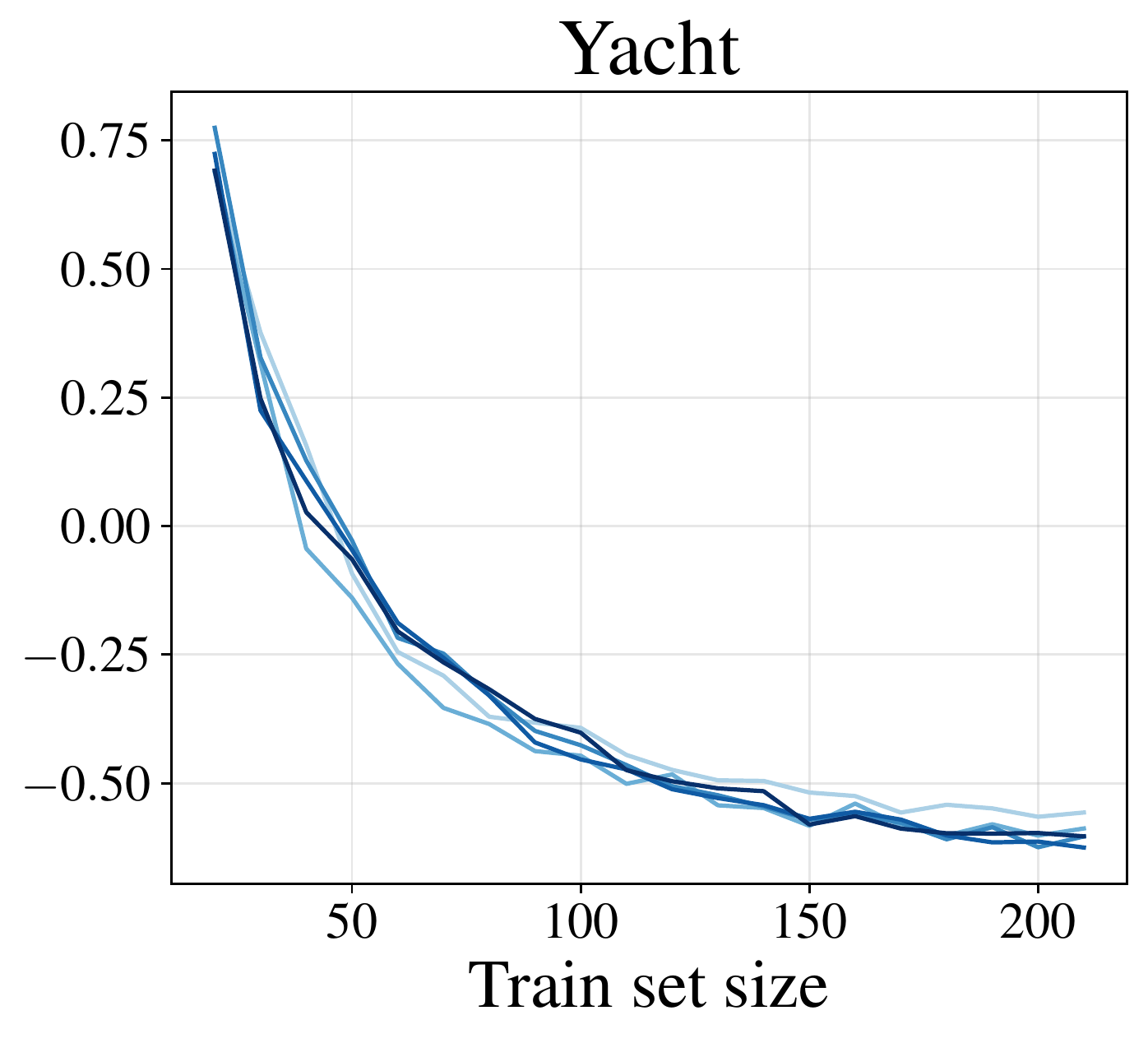}
\end{subfigure} \\
    \caption{Test NLL of DUNs using a stochastic relaxation of BALD. Different temperatures of the proposal distribution are compared. Means of 40 runs of the experiments are shown; standard deviations are not plotted for clarity.}
    \label{fig:res_stoch_bald_Ts_nll}
\end{figure}

\Cref{fig:res_stoch_bald_Ts_nll} shows the test NLL for DUNs using different temperatures $T$ for the proposal distribution.

\FloatBarrier
\newpage
\subsection{Depth posteriors for DUNs at different stages of active learning}

\begin{figure}[h!] 
\centering
    \begin{subfigure}
        \centering
        \includegraphics[width=0.34\linewidth]{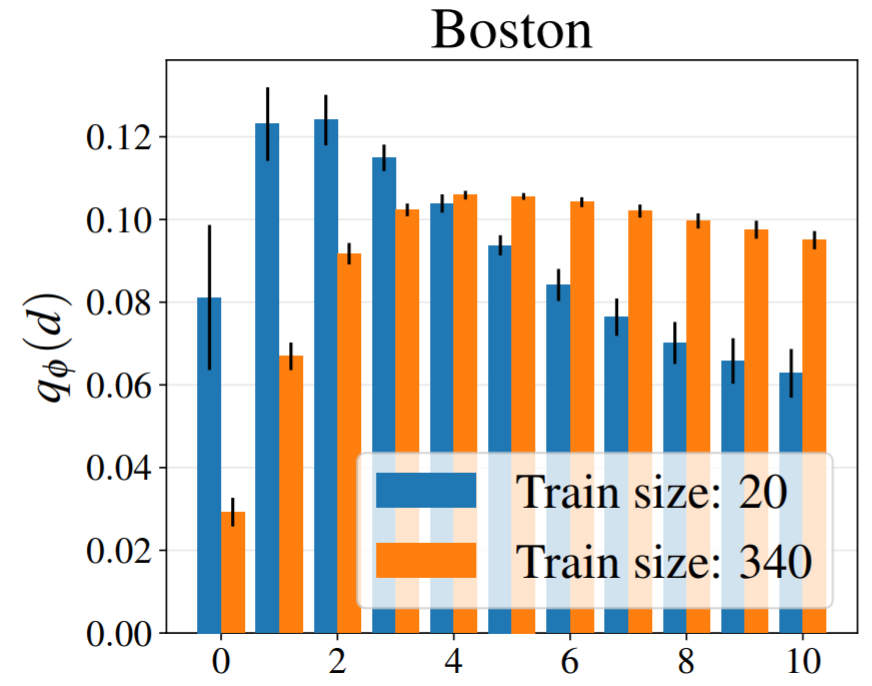}
    \end{subfigure}
    \begin{subfigure}
        \centering
        \includegraphics[width=0.31\linewidth]{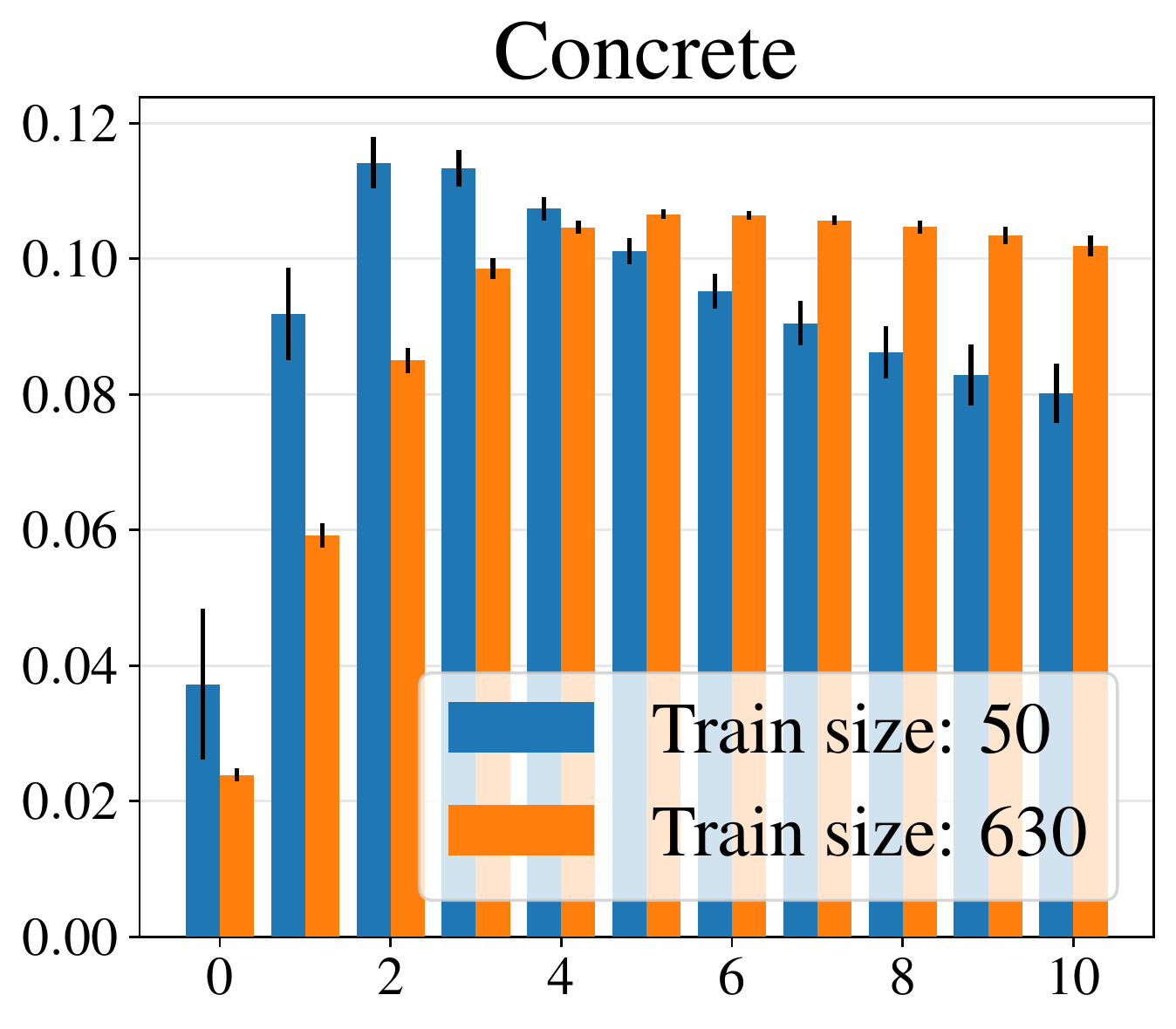}
    \end{subfigure} 
    \begin{subfigure}
        \centering
        \includegraphics[width=0.31\linewidth]{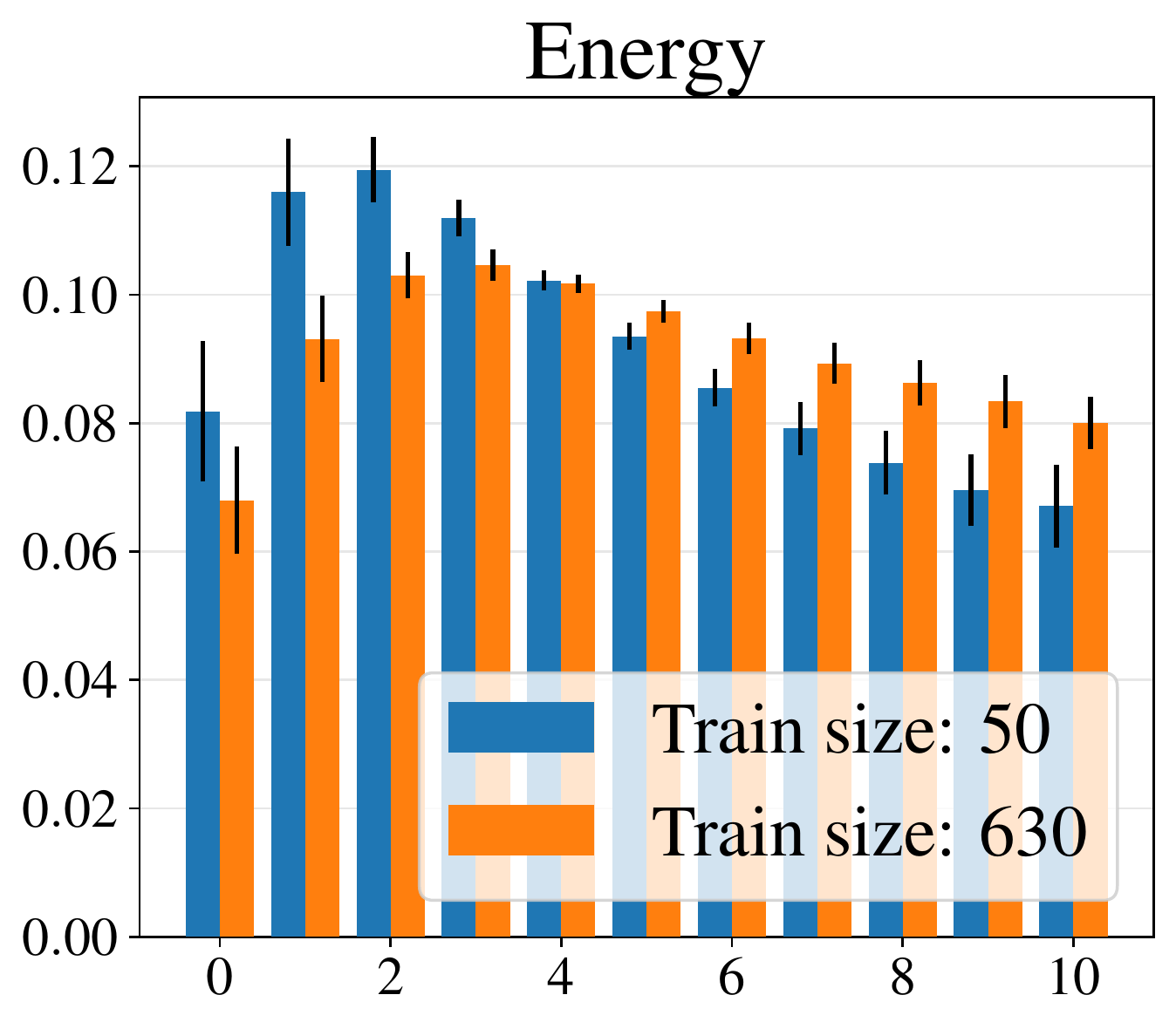} 
    \end{subfigure} \\
    \begin{subfigure}
        \centering
        \includegraphics[width=0.34\linewidth]{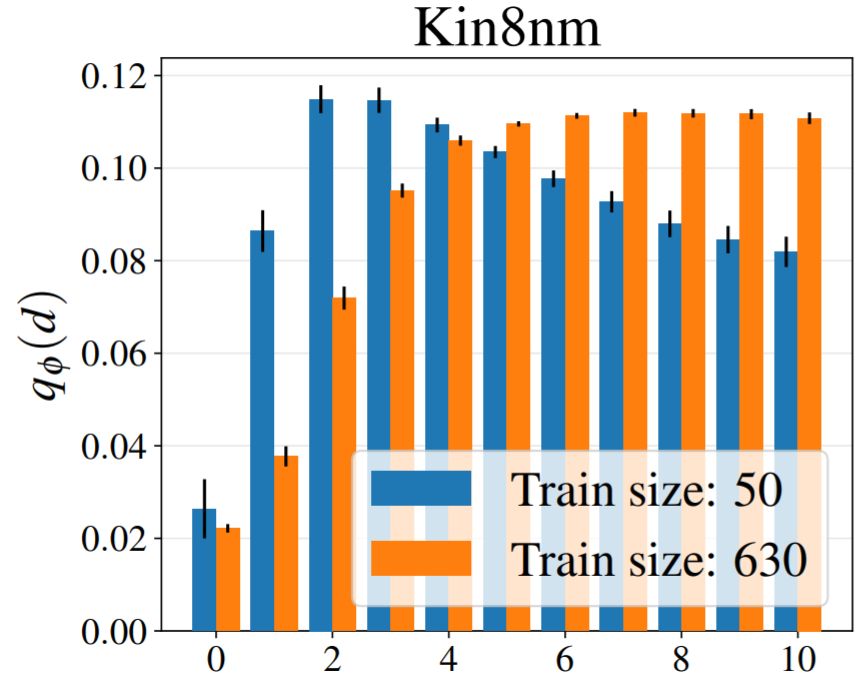}
    \end{subfigure} 
    \begin{subfigure}
        \centering
        \includegraphics[width=0.31\linewidth]{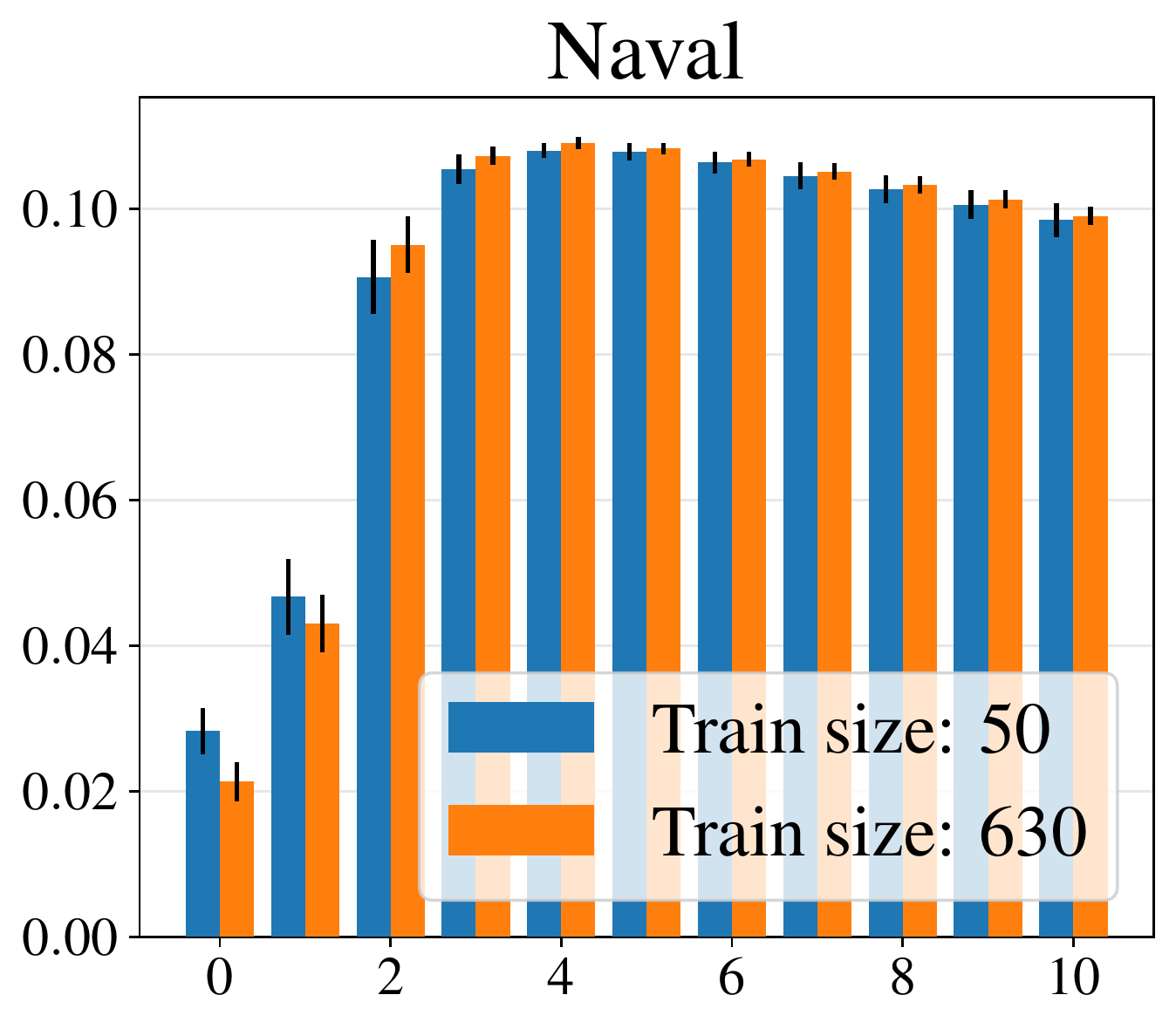}
    \end{subfigure} 
    \begin{subfigure}
        \centering
        \includegraphics[width=0.31\linewidth]{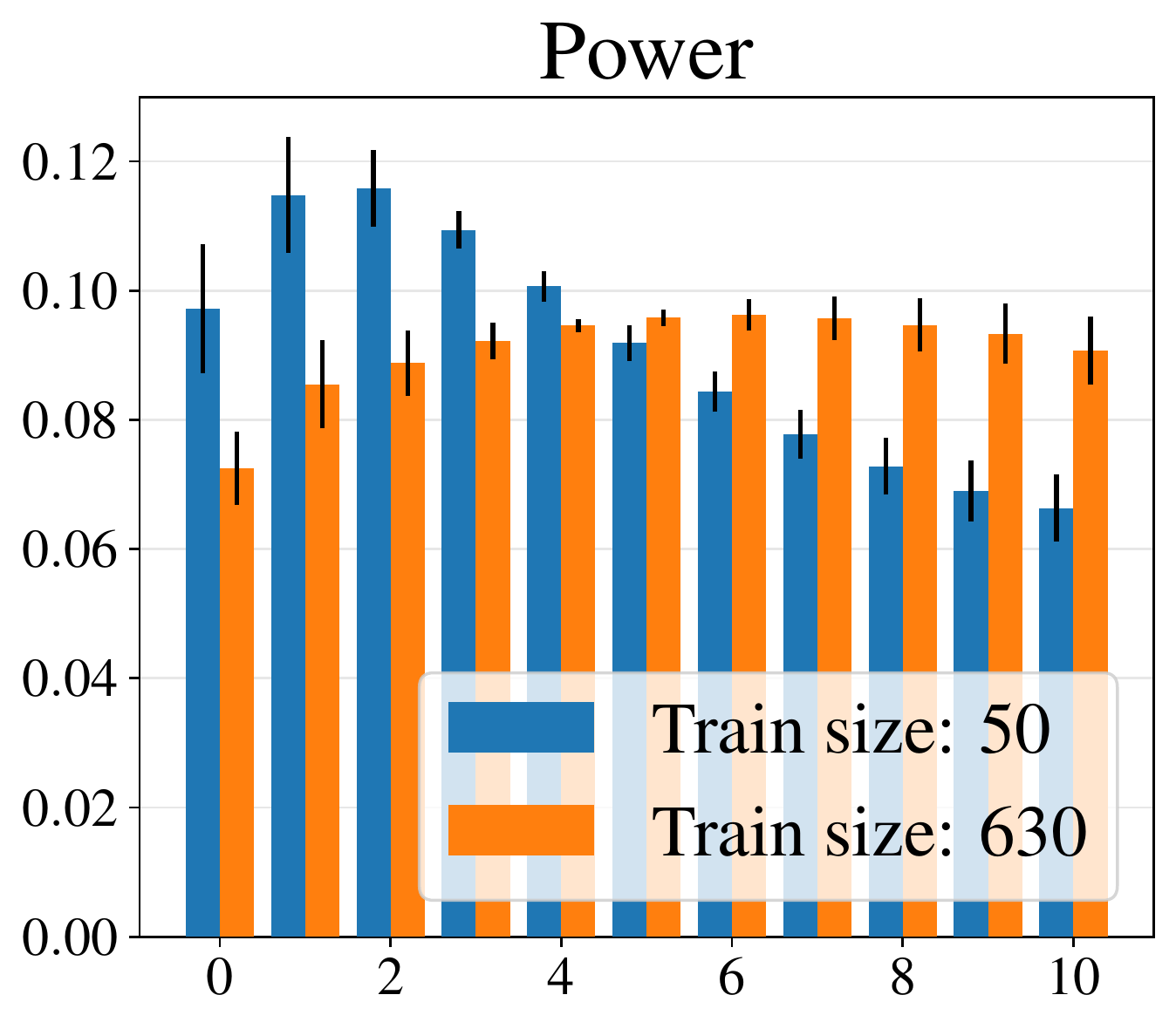}
    \end{subfigure} \\
    \begin{subfigure}
        \centering
        \includegraphics[width=0.34\linewidth]{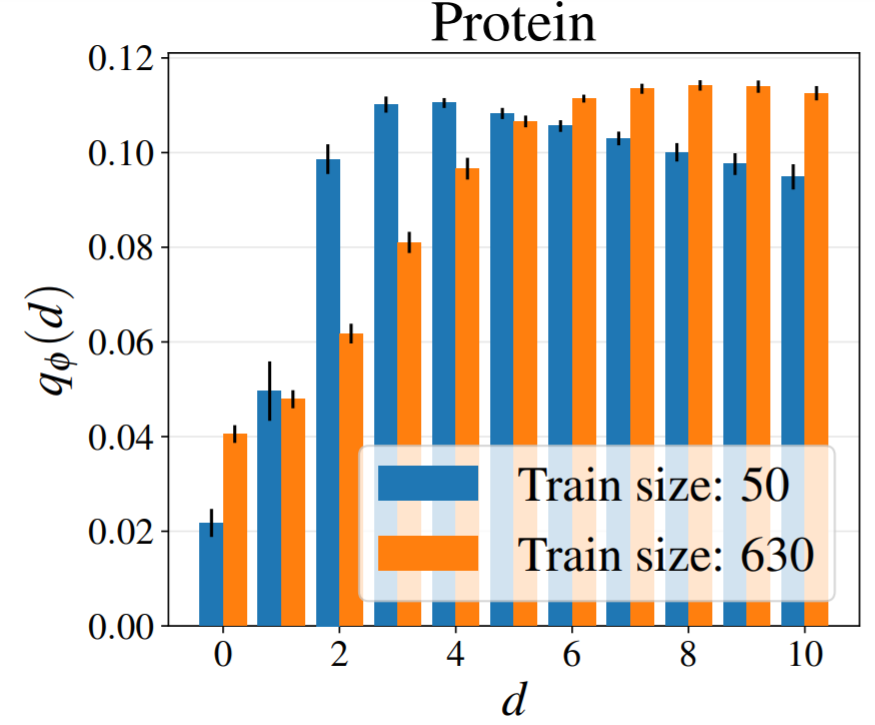}
    \end{subfigure} 
    \begin{subfigure}
        \centering
        \includegraphics[width=0.31\linewidth]{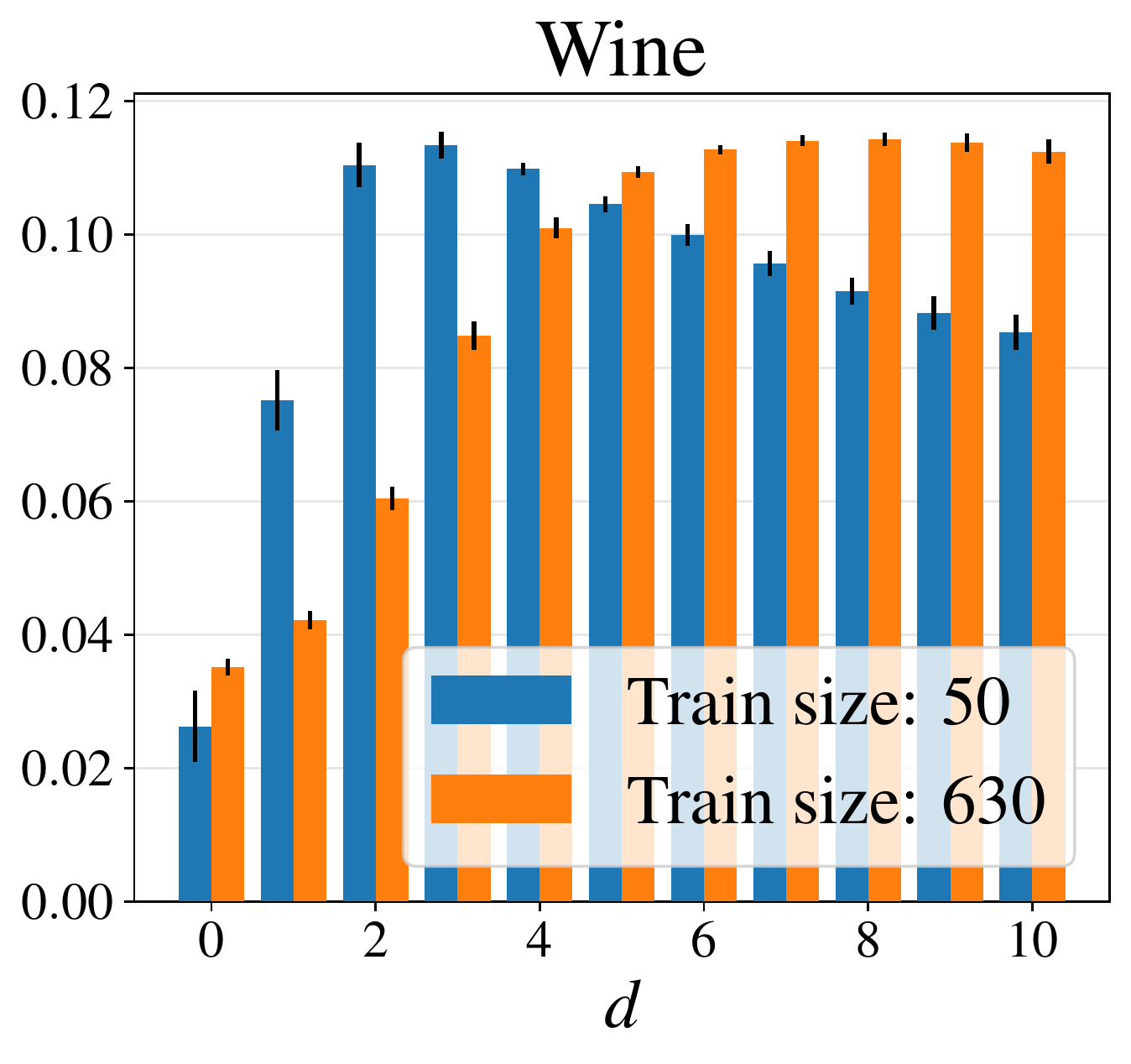}
    \end{subfigure} 
    \begin{subfigure}
        \centering
        \includegraphics[width=0.31\linewidth]{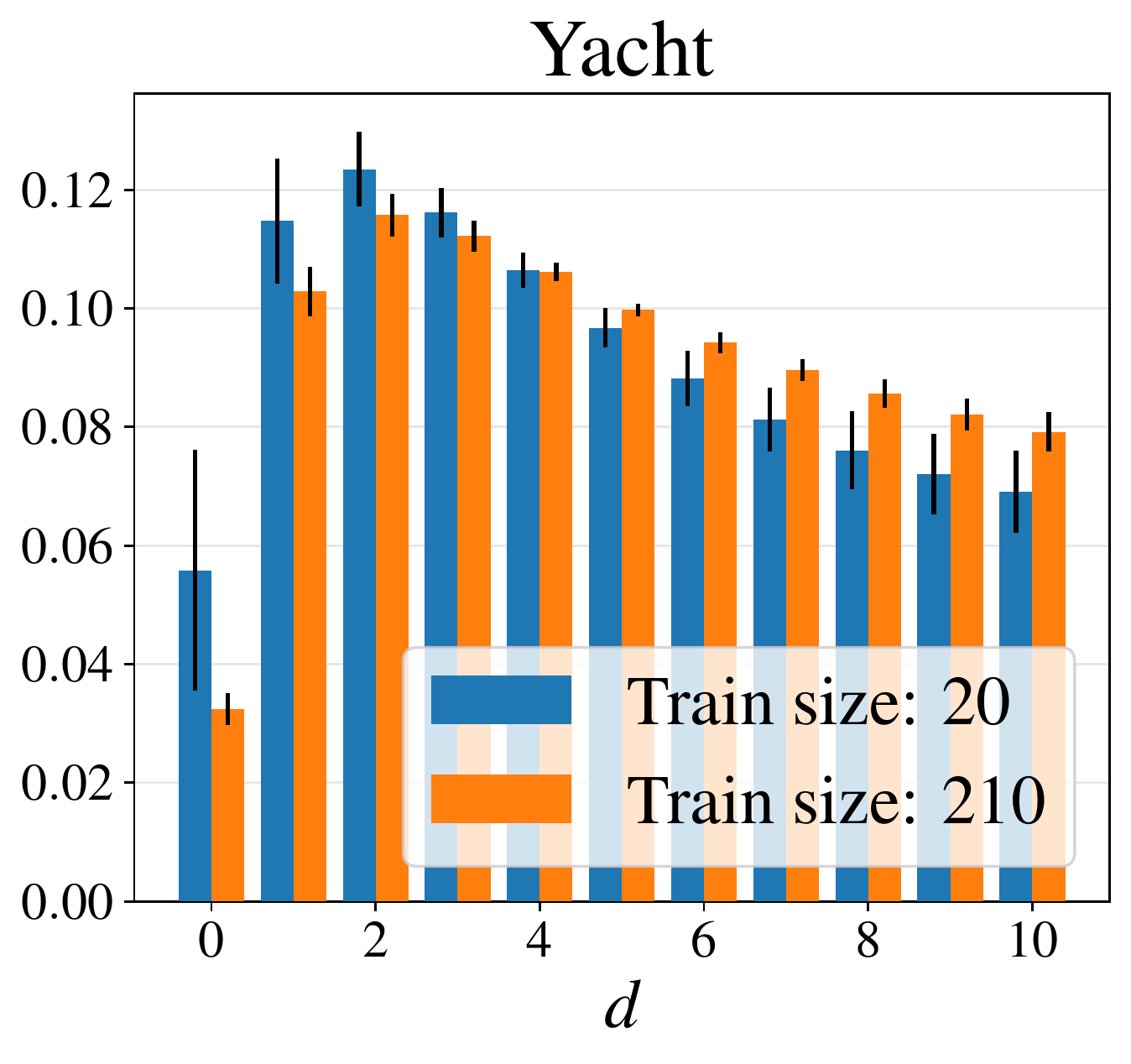}
    \end{subfigure} \\
    \caption{Posterior probabilities over depth for DUNs trained on UCI regression datasets, for the smallest (blue bars) and largest (orange bars) labelled datasets used in active learning.}
    \label{fig:res_reg_posts}
\end{figure}

\Cref{fig:res_reg_posts} shows the posterior distributions over depth for DUNs for the smallest and largest dataset sizes used during active learning. 

\end{document}